%% file: thesis.tex
\documentclass[a4paper,nobind]{ociamthesis} 

\usepackage{quoting} %
\quotingsetup{leftmargin=1em, rightmargin=1em, vskip=1ex}
\usepackage{pdfpages}
\usepackage{placeins}
\usepackage{array}
\usepackage{hhline}
\usepackage{xcolor,colortbl}
\usepackage{subcaption}
\definecolor{LightGreen}{rgb}{0.87,0.98,0.82}
\newcolumntype{a}{>{\columncolor{LightGreen}}>{\centering\arraybackslash}m}
\newcolumntype{b}{>{\centering\arraybackslash}m}

\newcommand{\tablecaption}{Summary of the attributes of  each chapter.}

\newenvironment{defenseinfo}[1][10em]
  {\noindent\begin{tabular}{@{}l@{~\makebox[#1]}}}
  {\end{tabular}}

\correctionstrue

\title{Risk-Sensitive and Robust Model-Based Reinforcement Learning and Planning}
\author{Marc Rigter}
\college{Pembroke College}

\degree{Doctor of Philosophy}
\degreedate{Michaelmas 2022}

\begin{document}

\setlength{\textbaselineskip}{22pt plus2pt}

\setlength{\frontmatterbaselineskip}{17pt plus1pt minus1pt}

\setlength{\baselineskip}{\textbaselineskip}


\setcounter{secnumdepth}{2}
\setcounter{tocdepth}{2}


\begin{romanpages}

\maketitle

\mbox{}
\vfill
\begin{defenseinfo}
	\textbf{Risk-Sensitive and Robust Model-Based Reinforcement Learning and} \\
	\textbf{Planning} \\
	Candidate: Marc Rigter \\
Supervisors: Professor Nick Hawes, Dr Bruno Lacerda \\
Examiners: Professor Jakob Foerster, Professor Aviv Tamar \\
Date of examination: 30th January, 2023 \\
\bigskip \\

\textbf{University of Oxford} \\
Goal-Oriented Long-Lived Systems Group (GOALS) \\
Department of Engineering Science \\
\bigskip
\end{defenseinfo}

\begin{acknowledgements}
 	\input{text/acknowledgements}
\end{acknowledgements}

\begin{abstract}
	\input{text/abstract}
\end{abstract}

\dominitoc 

\flushbottom

\tableofcontents




\end{romanpages}

\flushbottom
\include{text/ch1-intro}

\include{text/ch2-litreview}
\include{text/ch3-risk-mdps}
\include{text/ch4-regret-mdps}
\include{text/ch5-risk-bamdps}
\include{text/ch6-rambo}
\include{text/ch7-1r2r}
\include{text/ch8-lfd}
\include{text/ch9-conclusion}


\small
\bibliographystyle{unsrt}
\bibliography{references} 
\normalsize

\appendix
\include{text/appendices}

\end{document}

%% file: text/acknowledgements.tex

I'm deeply grateful to my supervisors Nick and Bruno.
Thank you for being fantastic mentors, and for always being generous with your time.
This thesis would not have been possible without your guidance, knowledge, enthusiasm, and support.

I greatly appreciate the insightful feedback of my examiners, Jakob Foerster and Aviv Tamar, which has helped to significantly improve the final version of this thesis.

Many thanks to Benjamin Morrell, KC Wong, Rob Reid, and David Willson, for mentoring me when I first became interested in research.
Thank you for having the patience and generosity to assist an enthusiastic undergraduate student.

I would like to extend my thanks to the members of the GOALS group that have been here throughout my DPhil: Charlie, Michael, Paul, Anna, Mohamed, Clarrisa, Matt, Raunak, Branton, Lara, and Alex.
Thank you also to Luigi.
I'm grateful to you all for your willingness to discuss ideas, and for making the environment we work in such a good one.

Thank you to Harrison Steel and Alex Kendall for taking the time to give me advice when I was applying to postgraduate studies when you had zero obligation to do so.
I will try my best to pass the help forward to future students.

I'm grateful to Paul Billing and Lara Cordoni, among the other teachers at JPC, for first helping me to get excited about science and mathematics.

Thank you to my friends outside of work in New Zealand, Australia, and Oxford.
 In particular, I would like to thank Bobby, Renee, Taylor, Sam K, Robin, Andrew, Rhys, Sam C, Tim, Rob, Kacy, Owen, Gonzalo, Kayla, Faizal, Flaminia, David, Jose, and Juliana.
I'm very lucky to know such awesome people.

Thank you Hannah for always being an amazing supporter, especially at the toughest points in this journey. 
I'm so glad that we have been able to go through our experience of Oxford together, and I'm very excited for the coming years.

I would like to express my gratitude to my grandmother, Lila, and to my sister, Katie.
Grandma, thank you for always keeping me well fed, and for being one of the most joyful people I know.
Katie, thank you for being a role model when we were younger, and for inspiring me to do more cool stuff outside of work.

Finally, I would like to thank my parents, Sandra and Tim. 
You have always been extremely supportive of everything I have tried to do, and I'm grateful for the lengths you went to in ensuring Katie and I had the best possible education.
I would not have made it very far without your support.

%% file: text/abstract.tex

Many sequential decision-making problems that are currently automated, such as those in manufacturing or recommender systems, operate in an environment where there is either little uncertainty, or zero risk of catastrophe.
As companies and researchers attempt to deploy autonomous systems in less constrained environments, it is increasingly important that we endow sequential decision-making algorithms with the ability to reason about uncertainty and risk. 

In this thesis, we will address both planning and reinforcement learning (RL) approaches to sequential decision-making.
In the planning setting, it is assumed that a model of the environment is provided, and a policy is optimised within that model. 
Reinforcement learning relies upon extensive random exploration, and therefore usually requires a simulator in which to perform training.
In many real-world domains, it is impossible to construct a perfectly accurate model or simulator.
Therefore, the performance of any policy is inevitably uncertain due to the incomplete knowledge about the environment.
Furthermore, in stochastic domains, the outcome of any given run is also uncertain due to the inherent randomness of the environment.
These two sources of uncertainty are usually classified as epistemic, and aleatoric uncertainty, respectively.
The over-arching goal of this thesis is to contribute to developing algorithms that mitigate both sources of uncertainty in sequential decision-making problems.

We make a number of contributions towards this goal, with a focus on model-based algorithms.
We begin by considering the simplest case where the Markov decision process (MDP) is fully known, and propose a method for optimising a risk-averse objective while optimising the expected value as a secondary objective.
For the remainder of the thesis, we no longer assume that the MDP is fully specified.
We consider several different representations of the uncertainty over the MDP, including a)  an uncertainty set of candidate MDPs, b) a prior distribution over MDPs, and c) a fixed dataset of interactions with the MDP.
In setting a), we propose a new approach to approximate the minimax regret objective, and find a single policy with low sub-optimality across all candidate MDPs.
In b), we propose to optimise for risk-aversion in Bayes-adaptive MDPs to avert risks due to both epistemic and aleatoric uncertainty under a single framework.
In c), the offline RL setting, we propose two algorithms to overcome the uncertainty that stems from only having access to a fixed dataset.
The first proposes a scalable algorithm to solve a robust MDP formulation of offline RL, and the second approach is based on risk-sensitive optimisation.
In the final contribution chapter, we consider an interactive formulation of learning from demonstration.
In this problem, it is necessary to reason about uncertainty over the performance of the current policy, to selectively choose when to request demonstrations.
Empirically, we demonstrate that the algorithms we propose can generate risk-sensitive or robust behaviour in a number of different domains.

%% file: text/ch1-intro.tex
\chapter{\label{ch:1-intro}Introduction}

In the coming years we will inevitably rely more on algorithms for decision-making, in applications ranging from robotics and self-driving cars~\cite{maddern20171}, to healthcare~\cite{tseng2017deep}, smart grids~\cite{charbonnier2022scalable}, and finance~\cite{spooner2018market}. 
Indeed, one study estimates that 47\% of current jobs are likely to be automated in the next two decades~\cite{frey2017future}.

Many processes that are currently automated, such as those in manufacturing or online advertising, operate in an environment where there is either little uncertainty, or zero risk of catastrophe.
As companies and researchers attempt to deploy autonomous systems in less constrained environments, it is increasingly important that we endow algorithms with the ability to reason about uncertainty and risk in a manner that reflects our priorities. 

As a motivating example, consider a healthcare setting where the goal is to recommend a treatment plan for a patient.
Due to variations between patients, it is impossible to predict the exact outcome of a treatment plan for any given patient.
An aggressive treatment plan may have the best outcome for most patients, yet result in adverse consequences for a minority. 
We may prefer to avoid these adverse consequences altogether by using a less aggressive treatment plan with lower risk, even if this results in slightly worse outcomes for the majority of patients.
An even better alternative could be to give each patient the opportunity to decide which treatment plan they would prefer, given their own outlook on the risks involved.

Another application in which it is necessary to reason about uncertainty and risk is robotics.
Robotic systems that interact with the human world, such as self-driving cars, must be robust to almost all situations that they might encounter, not just the most common ones.
As argued by~\cite{majumdar2020should}, the knowledge that a self-driving car typically avoids crashes is unlikely to be reassuring to passengers or regulators.
Instead, it would be more convincing to know that the car had been designed to act conservatively to avoid catastrophe even in unlikely situations, at the expense of making the system slower or less versatile.

The goal of this thesis is to contribute towards the development of algorithms that can reason about uncertainty and risk in a flexible manner.
We focus on~\textit{sequential} decision-making problems, where the decision made at one point in time influences the outcomes and possible courses of action in the future. 
Markov decision processes (MDPs) are a standard framework for sequential decision-making under uncertainty, and are the model that we will consider throughout this thesis.
The vast majority of research for MDPs develops algorithms that optimise performance in~\textit{expectation}. 
For most applications, reasoning about the expectation is sufficient.  
However, as we have discussed, in safety-critical settings we must also consider the~\textit{variability} in the outcomes.

In this thesis, we will predominantly focus on model-based approaches that model the dynamics of the environment, and use this model to inform the sequence of decisions that should be made.
In contrast to model-free approaches, model-based approaches are more suitable for guaranteeing safety~\cite{berkenkamp2017safe}, and can be used to improve interpretability~\cite{khan2009minimal}.
Furthermore, recent work has shown that it is possible for model-based methods to match the performance of model-free methods, even for complex tasks~\cite{hafner2020mastering}.

With the goal of designing algorithms that reason about uncertainty in mind, there are two key variables that we must consider: a) which source or sources of uncertainty would we like to address, and b) what type of objective we would like to optimise in the face of this uncertainty. 
The source of uncertainty can be classified as~\textit{epistemic} or~\textit{aleatoric} uncertainty, and the objectives that we may wish to optimise for can be classified as~\textit{risk-sensitive} or~\textit{robust} approaches. 
These concepts guide the development of most of the work in this thesis.
Furthermore, depending on the context, we may wish to develop an approach based on~\textit{planning} or~\textit{reinforcement learning} algorithms. 
This thesis considers both paradigms.
In the remainder of this section, we will provide an overview of each of these key concepts and how they relate to the research in this thesis.

\paragraph{Epistemic and aleatoric uncertainty} Sources of uncertainty are usually categorised into two types: epistemic uncertainty, which stems from a lack of knowledge about a parameter or phenomenon; and aleatoric uncertainty, which stems from the inherent variability in a random event.
Epistemic uncertainty can be reduced by gathering more data about the unknown phenomenon, whereas aleatoric uncertainty cannot.
In the context of control problems such as MDPs, epistemic and aleatoric uncertainty are also referred to as parametric and internal uncertainty respectively~\cite{mannor2007bias}. 

 In model-based algorithms for MDPs, epistemic uncertainty manifests in uncertainty over the transition and/or reward function of the MDP. 
 As more data is collected from interacting with the MDP, our knowledge about the MDP dynamics improves and the epistemic uncertainty is reduced.
 Aleatoric uncertainty manifests in the stochastic transition and/or reward function of the MDP. Even if we have perfect knowledge of the MDP, there is still variability over the possible outcomes for each episode due to this inherent stochasticity.
Throughout this thesis, we explore approaches that mitigate either epistemic or aleatoric uncertainty, or a combination of both sources of uncertainty.

\paragraph{Risk-sensitivity and robustness} Approaches to reasoning about and mitigating uncertainty can generally be placed into one of two different categories.
~\textit{Robust} approaches take a Knightian view of uncertainty~\cite{knight1921risk}, and do not require the relative likelihood of possible outcomes to be quantified.
In this type of approach, it is assumed that there is a set of plausible scenarios (the so-called ``uncertainty'' or ``ambiguity'' set).
The goal is usually to achieve the best possible performance in the worst-case instantiation of the uncertainty.
In other words, this approach ensures that the solution is ``robust'' to all possible scenarios.

In the context of MDPs, the uncertainty set is usually defined as a set of plausible MDPs. 
The most common objective is to find a policy that has the best expected value for the worst possible MDP within the set~\cite{iyengar2005robust, nilim2005robust}.
This approach is averse to the epistemic uncertainty over the MDP. 
However, due to the consideration of expected value in each MDP, it is neutral towards aleatoric uncertainty.
This is a natural formulation for problems where the policy is executed in the MDP many times, so that the aleatoric risk (i.e. the fluctuation in performance between each run) averages out~\cite{mannor2016robust}.

A shortcoming of the robust optimisation perspective is that it can be overly conservative: performance may be sacrificed in the vast majority of cases for the sake of improving performance in a highly unlikely worst-case situation. 
Furthermore, the robust approach only takes into consideration the epistemic uncertainty and therefore does not consider the risk of a poor outcome for any given run.

The alternative is to consider a~\textit{risk-sensitive} approach.
Here, the goal is to optimise a risk measure that maps the distribution over outcomes to a scalar value. 
Thus, unlike robust optimisation approaches, risk-sensitive approaches reason about the likelihood of each outcome.
Depending on the risk measure, the variability of the outcome, and the likelihood of catastrophic outcomes, are penalised to a varying degree.
Depending on the appetite for risk, an optimal risk-sensitive strategy may be willing to accept a small probability of an adverse outcome if it improves performance in most scenarios.

In this thesis, we consider robust approaches that optimise for the worst-case scenario to mitigate epistemic uncertainty in Chapters~\ref{ch:4-regret-mdps} and~\ref{ch:6-rambo}.
We also propose risk-sensitive approaches that optimise a risk measure to address aleatoric uncertainty (Chapter~\ref{ch:3-risk-mdps}) as well as epistemic uncertainty (Chapters~\ref{ch:5-risk-bamdps} and ~\ref{ch:7-1r2r}).
Our work on risk-sensitive algorithms will focus on generating~\emph{risk-averse} behaviour, i.e. avoiding risks.
This is in contrast to~\emph{risk-seeking} behaviour, which may be useful for exploration~\cite{dilokthanakul2018deep}.

\paragraph{Planning and reinforcement learning}
We consider algorithmic approaches related to both planning and reinforcement learning. 
In the planning setting, it is assumed that a model of the environment is known and we are interested in finding a suitable policy given that model.
In the reinforcement learning setting, a model of the environment is not provided a priori.
Instead, we derive a policy from samples of interaction with the environment.

Reinforcement learning methods can either be model-free, meaning that they do not require a model of the environment, or model-based, meaning that a model is learnt from environment interaction data, and this learnt model is utilised when optimising the policy.
The work on reinforcement learning in this thesis is predominantly model-based.

Reinforcement learning can further be categorised into online and offline variants.
In online reinforcement learning, further exploration can be used to gather more samples from the environment.
In offline, or batch reinforcement learning, the dataset  of interactions with the environment is fixed.

During the first half of this thesis, we focus on~\emph{tabular} planning methods. 
These methods are suitable for problems where there is a small discrete state and action space.
In the latter half of this thesis, we leverage deep reinforcement learning techniques that use neural network function approximation to enable greater scalability.
 
\section{Contributions}
This thesis contributes several approaches to solving sequential decision-making problems that go beyond optimising for the expected performance.
%
%
%
Each chapter considers a different combination of the concepts outlined in the previous section: whether to address aleatoric uncertainty, epistemic uncertainty, or both; whether to consider a risk-sensitive objective or a robust one; and whether the model is provided (planning) or learnt from interactions with the environment (model-based reinforcement learning).
In Chapters~\ref{ch:3-risk-mdps},~\ref{ch:4-regret-mdps},~\ref{ch:6-rambo}, and~\ref{ch:7-1r2r} we develop more performant algorithms for existing problem formulations. 
In Chapters~\ref{ch:5-risk-bamdps} and~\ref{ch:8-lfd} we propose new problem formulations in addition to algorithms for those problems.
Chapters~\ref{ch:3-risk-mdps}-\ref{ch:5-risk-bamdps} address tabular MDPs, while Chapters~\ref{ch:6-rambo}-\ref{ch:8-lfd} present more scalable approaches that utilise function approximation.

Here, we provide a summary of the contributions of each of the papers that comprise this thesis in the order that they are presented in Chapters~\ref{ch:3-risk-mdps}-\ref{ch:8-lfd}.
We begin by considering the simplest case where a tabular MDP is assumed to be known exactly.
We focus on the tradeoff between optimising for a risk-sensitive objective and the expected performance.
Specifically, we consider conditional value at risk (CVaR) which will be elaborated upon in the following chapter.
We make the following contributions:
\begin{itemize}
	\item Showing that there can be
	multiple policies that obtain the optimal CVaR in an MDP.
	\item An algorithm for optimising expected value in MDPs subject to the constraint that CVaR is optimal.
\end{itemize}

\noindent These contributions are presented in  Chapter~\ref{ch:3-risk-mdps}:~\textit{Planning for Risk-Aversion and Expected Value in MDPs}.
This work makes the strong assumption that the MDP is known exactly, i.e. that there is no epistemic uncertainty. 
Throughout the remainder of the thesis, we no longer make this restrictive assumption.

The \emph{uncertain MDP} setting considers uncertainty over the MDP represented in the form of an uncertainty set of plausible MDPs, any one of which could be the true MDP.
The~\textit{minimax regret} objective aims to find a single policy that has the lowest sub-optimality, in terms of expected value, over the set of possible MDPs.

In our work on the uncertain MDP setting, we make the following contributions:
\begin{itemize}
	\item A new approach to approximating the minimax regret objective.
	\item Proposing the use of options to capture dependencies between uncertainties to trade solution quality against the computation required.
\end{itemize}	

\noindent These contributions are presented in Chapter~\ref{ch:4-regret-mdps}: \textit{Minimax Regret Optimisation for Robust Planning in Uncertain Markov Decision Processes.}
In an earlier version of this work, we incorrectly claimed that our approach is exact under certain assumptions.
This error is discussed in Chapter~\ref{ch:4-regret-mdps}, and corrected in the associated corrigendum.

In the uncertain MDP setting in Chapter~\ref{ch:4-regret-mdps}, we assumed that we cannot learn more information about the environment through interaction.
In the next chapter, we consider the Bayesian formulation of model-based RL in which we maintain a belief distribution over the unknown MDP.
The belief is refined as experience is collected within the environment, thus enabling the policy to adapt to the true underlying environment.
This problem can be formulated as planning in a Bayes-adaptive MDP (BAMDP), a special class of Partially Observable MDP (POMDP), in which the state is augmented with the belief over the underlying MDP.

In this problem setting, we make the following contributions:

\begin{itemize}
	\item  Proposing to optimise the CVaR of the return in the BAMDP as a means to mitigate both epistemic and aleatoric uncertainty under a single framework.
	\item  An algorithm based on MCTS and Bayesian optimisation for this problem.
\end{itemize}

\noindent These contributions are presented in Chapter~\ref{ch:5-risk-bamdps}: \textit{Risk-Averse Bayes-Adaptive Reinforcement Learning.}

Chapter~\ref{ch:4-regret-mdps} and Chapter~\ref{ch:5-risk-bamdps}  assume that prior information about the MDP is specified either in the form of a set of plausible MDPs, or a suitable prior over MDPs, respectively.
In Chapters~\ref{ch:6-rambo} and~\ref{ch:7-1r2r} we address the offline deep reinforcement learning setting, in which the starting point is a dataset of transitions collected from the environment.
Unlike the standard RL setting, no further exploration of the environment is allowed. 
We consider model-based offline RL, where we first learn a model of the environment dynamics from the dataset, and optimise a policy within that model.
The naive application of this approach results in the policy learning to exploit areas of the model not covered by the dataset, where the model is inaccurate.
This problem is referred to as distributional shift.
In Chapter~\ref{ch:6-rambo},~\textit{RAMBO: Robust Adversarial Model-Based Offline RL}, we propose a novel approach to address this issue, and make the following contributions:
\begin{itemize}
	\item Introducing a novel and theoretically-grounded model-based offline RL algorithm which enforces
	conservatism and prevents distributional shift by training an adversarial dynamics model to generate pessimistic synthetic data for out-of-distribution actions.
	\item Adapting Robust Adversarial RL (RARL) to model-based offline RL by proposing a new
	formulation of RARL, where instead of defining and training an adversary policy, we directly train
	the model adversarially.
\end{itemize}

In Chapter~\ref{ch:7-1r2r},~\textit{One Risk to Rule Them All: A Risk-Sensitive Perspective on Model-Based Offline RL}, we propose a risk-sensitive approach to offline RL.
In this chapter, we utilise risk-sensitive optimisation to avoid the issue of distributional shift. 
This work makes the following contributions:

\begin{itemize}
	\item Proposing risk-sensitive optimisation as a method to address the issue of distributional shift in
	offline RL.
	\item Introducing the first model-based algorithm for risk-sensitive offline RL, which jointly addresses
	epistemic and aleatoric uncertainty.
\end{itemize}

The main motivation for the offline RL setting addressed in Chapters~\ref{ch:6-rambo} and~\ref{ch:7-1r2r} is avoiding the extensive online exploration required by online RL.
In the last contribution chapter, we take an alternative approach to reducing the amount exploration required, and consider learning from demonstration.
In our interactive setting, demonstrations can be requested from an expert to help train a policy that is further fine-tuned using RL.

In this work, we make the following contributions:
\begin{itemize}
	\item A new problem formulation where there is a cost associated with the human effort for providing demonstrations, and resetting the system from failure. The goal is to minimise the total human cost incurred over many episodes.
	\item An algorithm for deciding when to request demonstrations which is based on posing the problem as a contextual multi-armed bandit.
\end{itemize}

\noindent This is presented in Chapter~\ref{ch:8-lfd}: \emph{A Framework for Learning from Demonstration with Minimal Human Effort}.

\subsection{Contribution Summary}
Table~\ref{tab:summary} summarises the attributes of each chapter in this thesis. 
Each chapter is summarised according to the following attributes:
\begin{itemize}
	\item \textbf{Prior Knowledge} At one end of the spectrum, Chapter~\ref{ch:3-risk-mdps} assumes that we have perfect knowledge of the environment, while at the other end, Chapters~\ref{ch:6-rambo} and~\ref{ch:7-1r2r} assume that we only have access to transition data from the environment which may be noisy or of low quality.
	The other chapters make different assumptions: that the uncertainty over the MDP is represented by an uncertainty set or prior distribution, or that we have access to demonstrations from an expert.
	\item \textbf{Considering Aleatoric Uncertainty} Approaches that consider aleatoric uncertainty take into account the variability in the performance of the policy due to inherent stochasticity in the environment.
	 Approaches which optimise for the expected value disregard aleatoric uncertainty.
	 Chapters~\ref{ch:3-risk-mdps},~\ref{ch:5-risk-bamdps}, and~\ref{ch:7-1r2r} take aleatoric uncertainty into consideration by optimising a risk measure.
	\item \textbf{Considering Epistemic Uncertainty} Approaches that consider epistemic uncertainty take into consideration the presence of incomplete knowledge about the environment due to modelling errors, or a lack of data. 
	With the exception of Chapter~\ref{ch:3-risk-mdps}, all the chapters of this thesis consider epistemic uncertainty in some manner.
	\item \textbf{Risk or Robustness} Risk-sensitive approaches optimise a risk measure, while robust approaches optimise for the worst-case scenario over an uncertainty set.
	With the exception of Chapter~\ref{ch:8-lfd}, all the chapters of this thesis can be classified into one of these two categories.
	\item \textbf{Paradigm} This attribute describes whether the work addresses tabular MDPs, or proposes an approach that utilises function approximation to improve scalability to large and/or continuous problems.
	The function approximators that we consider in this work are predominantly neural networks.
	The first half of this thesis considers tabular MDPs, while the latter half proposes approaches that utilise function approximation.
\end{itemize}

\begin{table}[]
	\renewcommand{\arraystretch}{1.3}
	\caption{\tablecaption \label{tab:summary}
}
	\hyphenpenalty=100000
	\footnotesize
	\resizebox{\columnwidth}{!}{%
	\begin{tabular}{l|b{1.7cm}|b{1.7cm}|b{1.7cm}|b{1.7cm}|b{1.7cm}|b{1.9cm}|}
		\hhline{~|-|-|-|-|-|-|}
		&  \textbf{Chapter~\ref{ch:3-risk-mdps}} &  \textbf{Chapter~\ref{ch:4-regret-mdps}}  & \textbf{Chapter~\ref{ch:5-risk-bamdps}} & \textbf{Chapter~\ref{ch:6-rambo}} & \textbf{Chapter~\ref{ch:7-1r2r}} & \textbf{Chapter~\ref{ch:8-lfd}} \\ \hline
		\multicolumn{1}{|m{1.9cm}|}{\textbf{Prior knowledge}} & MDP fully known & MDP uncertainty set & Prior over MDPs & Fixed dataset & Fixed dataset & Expert demonstrator \\ \hline
		\multicolumn{1}{|m{1.9cm}|}{\textbf{Considers aleatoric uncertainty}} & Yes & No & Yes &  No & Yes & No \\ \hline
		\multicolumn{1}{|m{1.9cm}|}{\textbf{Considers epistemic uncertainty}} & No & Yes & Yes & Yes & Yes & Yes \\ \hline
		\multicolumn{1}{|m{1.9cm}|}{\textbf{Risk or robustness}} & Risk &  Robustness  & Risk & Robustness & Risk & N/A \\ \hline
		\multicolumn{1}{|m{1.9cm}|}{\textbf{Paradigm}} & Tabular &  Tabular  & Tabular & Function Approx. & Function Approx. & Function Approx. \\ \hline
	\end{tabular}
	}%
	\vspace{2mm}
	Chapter~\ref{ch:3-risk-mdps} - Planning for Risk-Aversion and Expected Value in MDPs. \\
	Chapter~\ref{ch:4-regret-mdps} - Minimax Regret Optimisation for Robust Planning in Uncertain
	Markov Decision Processes. \\
	Chapter~\ref{ch:5-risk-bamdps} - Risk-Averse Bayes-Adaptive Reinforcement Learning. \\
	Chapter~\ref{ch:6-rambo} - RAMBO-RL: Robust Adversarial Model-Based Offline Reinforcement Learning. \\
	Chapter~\ref{ch:7-1r2r} - One Risk to Rule Them All: A Risk-Sensitive Perspective on Model-Based Offline RL. \\
	Chapter~\ref{ch:8-lfd} - A Framework for Learning from Demonstration with Minimal
	Human Effort. \\
\end{table}

In each chapter of the thesis, we repeat Table~\ref{tab:summary} to discuss how the ideas proposed in that chapter relate back to each of the attributes described here.

\FloatBarrier

\section{Thesis Structure}
The following chapter is a literature review providing context for this thesis.
As this is an integrated thesis, each of the subsequent chapters corresponds to a paper.
In each of the paper chapters, we first give an overview of the contributions of the paper.
Following the presentation of each paper, we discuss the limitations of the work and suggest directions for future work to address these limitations.
The full appendices of each paper are included at the end of this document. However, in the contribution chapters we have included portions of the appendices that provide additional insight into our work.
Citations in the main body of the thesis refer to references included in the bibliography at the end of this thesis, following Chapter~\ref{ch:9-conclusion}.
Citations within each paper refer to the references in the bibliography at the end of that paper.

In chronological order, the publications which comprise this integrated thesis are the following~\cite{rigter2020framework, rigter2021minimax, rigter2021risk, rigter2022planning, rigter2022rambo, rigter2022one}:

\begin{itemize}
	\item \underline{Marc Rigter}, Bruno Lacerda, and Nick Hawes (2020). A framework for learning from demonstration with minimal human effort. \textit{IEEE Robotics and Automation Letters} (RAL).
	
	\item \underline{Marc Rigter}, Bruno Lacerda, and Nick Hawes (2021). Minimax regret optimisation for robust planning in uncertain Markov decision processes. \textit{AAAI Conference on Artificial Intelligence} (AAAI).
	
	\item \underline{Marc Rigter}, Bruno Lacerda and Nick Hawes (2021). Risk-averse Bayes-adaptive reinforcement learning. \textit{Advances in Neural Information Processing Systems} (NeurIPS).
	
	\item \underline{Marc Rigter}, Paul Duckworth, Bruno Lacerda and Nick Hawes (2022). Planning for risk-aversion and expected value in MDPs. \textit{International Conference on Automated Planning and Scheduling} (ICAPS).
	
	\item \underline{Marc Rigter}, Bruno Lacerda and Nick Hawes (2022). RAMBO-RL: Robust adversarial model-based offline reinforcement learning. \textit{Advances in Neural Information Processing Systems} (NeurIPS).
	
	\item \underline{Marc Rigter}, Bruno Lacerda and Nick Hawes (2022). One Risk to Rule Them All: A Risk-Sensitive Perspective on Model-Based Offline Reinforcement Learning. \textit{arXiv preprint arXiv:2212.00124}.
\end{itemize}

Note that to improve the flow of the thesis, the papers are not presented in chronological order.
Other publications that I have contributed to over the course of my DPhil that are not included in this thesis are~\cite{ishida2019robot, rigter2022optimal, costen2022shared, gautier2022risk, costen2022planning}:

\begin{itemize}
	\item Shu Ishida, \underline{Marc Rigter}, and Nick Hawes. Robot path planning for multiple target regions (2019).~\textit{European Conference on Mobile Robots} (ECMR). 
	
	\item \underline{Marc Rigter}, Danial Dervovic, Parisa Hassanzadeh, Jason Long, Parisa Zehtabi, Daniele Magazzeni (2022). Optimal admission control for multiclass queues with time-varying arrival rates.~\textit{AAAI Conference on Artificial Intelligence} (AAAI).
	
	\item Clarissa Costen, \underline{Marc Rigter}, Bruno Lacerda and Nick Hawes (2022). Shared autonomy systems with stochastic operator models.~\textit{International Joint Conference on Artificial Intelligence} (IJCAI).
	
	\item Anna Gautier, \underline{Marc Rigter}, Bruno Lacerda, Nick Hawes, and Michael
	Wooldridge. (2022). Risk-Constrained Planning for Multi-Agent
	Systems with Shared Resources.~\textit{International Conference on Autonomous Agents and Multiagent Systems}.
	
	\item Clarissa Costen, \underline{Marc Rigter}, Bruno Lacerda and Nick Hawes (2023). Planning with hidden parameter polynomial MDPs.~\textit{AAAI Conference on Artificial Intelligence} (AAAI).
\end{itemize}

%% file: text/ch2-litreview.tex
\chapter{\label{ch:2-litreview}Background}

\section{Markov Decision Processes}
Markov decision processes (MDPs)~\cite{puterman1994markov, sutton2018reinforcement} are a standard framework for sequential decision-making under uncertainty.
An MDP is defined by the tuple $M = (S, A, T, R, \mu_0, \gamma)$.
$S$ defines the set of possible states of the environment, $A$ defines the set of actions available to the agent, and $\mu_0$ defines the initial state distribution.
At each step, the agent observes the current state, $s \in S$, and on that basis chooses an action, $a \in A$.
After applying the action, the agent receives a reward according to the reward function, $R: S \times A \rightarrow \mathbb{R}$ (or equivalently a cost). Subsequently, the environment transitions stochastically to a new state according to the transition function, $T : S \times A \rightarrow \Delta S$, where $\Delta S$ indicates a distribution over states.

A deterministic Markovian policy is a mapping from each state to an action, $\pi : S \rightarrow A$. 
The standard objective in MDPs is to find the policy, $\pi^*$, that maximises the expected cumulative discounted reward: 

\begin{equation}
	\label{eq:mdp_optimality}
	\pi^* = \textnormal{arg}\max_\pi \mathbb{E}^\pi \Big[\sum_{t=0}^\infty \gamma^t R(s_t, a_t) \Big].
\end{equation}

\subsection{Solution Methods}
Value-based methods for MDPs are based on the principle of dynamic programming~\cite{bellman1966dynamic}, which decomposes the overarching optimisation problem into a sequence of sub-problems.
These methods solve Equation~\ref{eq:mdp_optimality} by computing the optimal~\textit{value function} at each state:

$$
V^*(s) = \max_\pi \Big[\sum_{t=0}^\infty \gamma^t R(s_t, \pi(s_t))\ |\ s_0 = s \Big],
$$

\noindent which can be found by recursively applying the Bellman equation:

$$
V^*(s) = \max_a \big[ R(s, a) + \gamma \sum_{s' \in S} T(s' | s, a) \cdot V^*(s') \big].
$$

Exact planning approaches~\cite{kolobov2012planning, puterman1994markov} compute the value function exactly. 
In cases where the transition dynamics are unknown or state space is large, approximate methods such as sample-based planning~\cite{browne2012survey} or reinforcement learning (RL)~\cite{sutton2018reinforcement} are more suitable. 
These latter approaches usually approximate the value function by sampling interactions with the environment, although some RL methods forgo learning a value function in favour of directly optimising the policy~\cite{sutton1999policy}.

\subsection{Overview of Related Work}

The majority of algorithms for MDPs optimise the expected value objective given in Equation~\ref{eq:mdp_optimality}, and ignore the variability in the cost or reward received between each episode.
As discussed in Chapter~\ref{ch:1-intro}, for some applications, ignoring variability in the outcomes is inappropriate.
Some degree of sensitivity towards adverse outcomes can be introduced into the expected reward framework by shaping the reward function by adding large negative rewards.
However, ``hacking'' the reward function in this manner magnifies the already difficult problem of reward function design~\cite{knox2021reward}.
If the reward function is not designed carefully, this can lead to irrational behaviour~\cite{amodei2016concrete}.
Furthermore, shaping the reward function can make it difficult to understand what the algorithm is ultimately optimising for~\cite{smith2021exponential}, especially in domains where the original reward function has a clear interpretation.
Therefore, in Section~\ref{sec:lit_risk} we will review risk-sensitive approaches that take into account variability in performance via the optimisation of risk measures which map the distribution over rewards to a scalar value.

An additional limitation of planning-based approaches is that they assume access to an exact model of the MDP.
However, in practical applications the MDP cannot be estimated exactly, and even small errors in the model can result in significant errors in the solution~\cite{mannor2007bias}.
In Section~\ref{sec:lit_robustness}, we review approaches that mitigate errors in the MDP model via worst-case approaches inspired by robust optimisation~\cite{iyengar2005robust, nilim2005robust}.
Furthermore, in Section~\ref{sec:lit_bamdps} we will consider approaches which produce a policy that adapts to the true underlying MDP.

Analogous to the requirement for a model for planning, RL methods usually require an accurate simulator~\cite{hwangbo2018per}.
This is because RL algorithms require extensive exploration to find a solution, which can be costly or dangerous in the real environment~\cite{garcia2015comprehensive}.
Therefore, performing online RL in the real environment is infeasible in many domains.
An alternative is to instead consider the~\textit{offline} RL setting where a policy is trained using a fixed dataset with no online exploration.
By optimising the reward signal, offline RL algorithms can improve upon the behaviour that generated the dataset.
Offline RL is applicable to domains such as robotics~\cite{kumar2021workflow} and healthcare~\cite{tang2021model}, where it is possible to leverage large existing historical datasets. 
The main limitation of offline RL is that if there is no high-quality data in the dataset, it is not possible to learn a performant policy due to the absence of further exploration.
We provide a more detailed review of offline RL in Section~\ref{sec:offline_rl}.

Another means to overcome the extensive exploration required by online RL is to use methods based on learning from demonstration~\cite{argall2009survey} (also known as imitation learning).
Unlike the offline RL setting, learning from demonstration assumes that the interaction data is collected using an expert policy.
The policy derived from the demonstrations can be used as the final policy, or alternatively, if a reward function is available it can be further fine-tuned using RL~\cite{osa2018algorithmic}. 
We provide a brief overview of learning from demonstration in Section~\ref{sec:lfd}.

\section{Risk in MDPs}
\label{sec:lit_risk}

In this subsection, we review previous work on risk in MDPs. 
We give an overview of risk measures, before discussing algorithms that optimise for risk in MDPs based on both planning or reinforcement learning.
Throughout this subsection we will discuss risk in the context of mitigating aleatoric uncertainty due to the inherent stochasticity of the environment. 
We will address epistemic uncertainty in future subsections.

For this subsection, we will assume that the objective is to minimise cost, as this is the standard convention when analysing risk. 
However, reward maximisation can equivalently be considered.

\paragraph{Risk Measures}
Early work on risk in MDPs~\cite{howard1972risk} considers the exponential utility framework~\cite{browne1995optimal}, where the cost is transformed according to a convex utility function to achieve risk-aversion.
The expected value of this transformed cost is optimised.
This generates risk-averse behaviour by magnifying the influence of large costs.
However, as previously noted in this chapter, it is difficult to ``shape'' the cost function appropriately to achieve the desired behaviour~\cite{majumdar2020should}.

A~\textit{risk measure} maps a probability distribution over possible costs to a scalar value~\cite{emmer2015best}.
Such measures take into consideration the variability of the cost.
Depending on the risk measure, a varying degree of emphasis can be placed on the worst possible outcomes.   

Denote by $\Omega$ the set of all possible outcomes in a probability space. 
Consider a function $Z : \Omega \rightarrow \mathbb{R}$ that assigns a cost of $Z(\omega)$ to any stochastic outcome $\omega \in \Omega$. 
$Z$ is then a random variable representing the cost incurred depending on the outcome.
A risk measure, $\rho$, is a mapping from any cost random variable, $Z$, to a real number.

~\textit{Coherent} risk measures are proposed by~\cite{artzner1999coherent}, and~\cite{majumdar2020should} argues that coherent risk measures should be employed when designing autonomous systems. 
In short, coherent risk measures adhere to several axioms associated with rational decision-making.
The axioms that any coherent risk measure must adhere to are the following~\cite{artzner1999coherent}:
\begin{itemize}
	\item \textit{Monotonicity}.  Suppose that $Z_1$ and $Z_2$ are two cost random variables, and that $Z_1(\omega) \leq Z_2(\omega)$ for all $\omega \in \Omega$. Then $\rho(Z_1) \leq \rho(Z_2)$. 
	
	Interpretation: if under all possible scenarios the cost for $Z_1$ is less than $Z_2$, then $Z_1$ should be considered less risky.
	\item \textit{Translation invariance}. Consider some cost random variable, $Z$, and fixed cost $c \in \mathbb{R}$. Then $\rho(Z - c) = \rho(Z) - c$. 
	
	Interpretation:  a guaranteed reduction in the cost by a fixed amount reduces the risk by the same amount.
	\item \textit{Positive homogeneity}. Let $Z$ be some cost random variable, and $\beta \geq 0$ a scalar. Then $\rho(\beta Z) = \beta \rho(Z)$. 
	
	Interpretation: if all costs are scaled by some amount, then the risk is scaled by that same amount.
	
	\item \textit{Subadditivity}. Consider two cost random variables, $Z_1$ and $Z_2$. Then $\rho(Z_1 + Z_2) \leq \rho(Z_1) + \rho(Z_2)$.
	
	Interpretation: diversification by jointly taking  two different strategies should be considered at most as risky as the individual strategies considered separately.
\end{itemize}

Some works on risk in MDPs consider the mean-variance criterion~\cite{sobel1982variance, tamar2012policy} where the mean of the total cost is penalised by the variance.
However, this is not a coherent risk measure as it is translation invariant but does not satisfy the other required axioms~\cite{majumdar2020should}.
As an illustration of why this may lead to irrational behaviour, consider a reduction in the cost associated with all possible outcomes.
Because the mean-variance criterion does not adhere to the monotonicity property, this is not guaranteed to improve the mean-variance objective.
This is because even if the cost is reduced for all outcomes, the variance may increase, and outweigh the improvement in the mean.

Another risk measure that is not coherent is the~\textit{value at risk} (VaR).
VaR is frequently used in financial decision-making~\cite{sollis2009value}. 
For some probability level, the VaR is equal to the corresponding quantile of the cost distribution.
VaR is not coherent as it does not satisfy the subadditivity property.
In other words, the VaR of jointly taking two strategies can be higher than the sum of the VaR of the individual strategies.
This problem is caused by the fact that the VaR is a quantile and does not take into account the distribution over costs in the tail~\cite{danielsson2005subadditivity}.
Because the distribution of the tail outcomes is not taken into consideration, VaR may under-represent the risk of the worst outcomes for fat-tailed distributions~\cite{favre2002mean}.

\begin{figure}[htb]
	\centering
	\includegraphics[width=9cm]{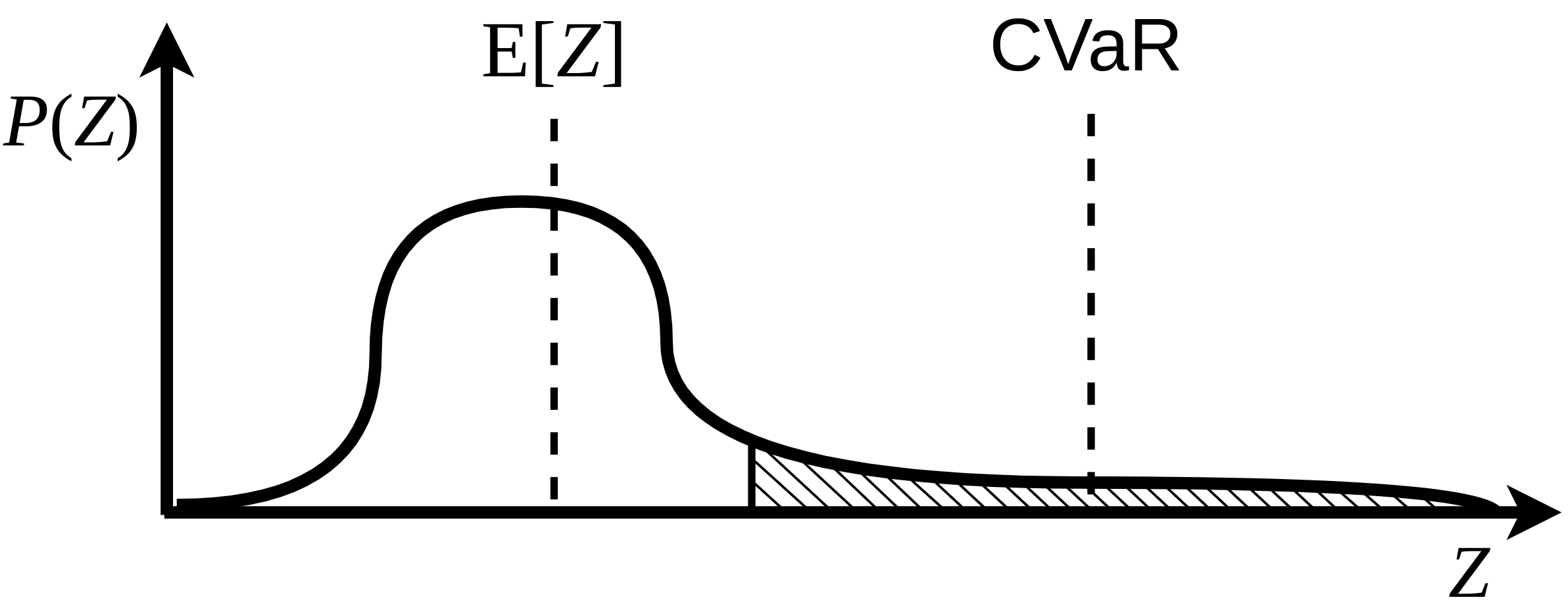}
	\caption{Illustration of the conditional value at risk at confidence level $\alpha$ for cost random variable $Z$. The shaded region indicates the $\alpha$-portion of the tail of the distribution. \label{fig:cvar}}
\end{figure}

Risk measures which are coherent include the conditional value at risk (CVaR)~\cite{rockafellar2000optimization}, the entropic value at risk~\cite{ahmadi2012entropic}, the mean-semideviation~\cite{tamar2015policy}, and the Wang risk measure~\cite{wang2000class}.
Throughout this thesis, we focus predominantly on CVaR as it is intuitive to understand and is the risk measure most commonly used in recent research on both planning under uncertainty and reinforcement learning~\cite{chow2014algorithms, chow2015risk, tamar2014optimizing, bauerle2011markov}.
It is also the risk measure recommended by central banks for assessing financial risk ~\cite{basel2013fundamental}.
For a continuous cost random variable, the CVaR at confidence level $\alpha$ is the mean of the $\alpha$-portion of the right tail of the cost distribution~\cite{tamar2014optimizing}, as illustrated in Figure~\ref{fig:cvar}.

\paragraph{Static and Dynamic Risk}
There are two natural approaches to considering risk in sequential problems.
The first is the~\textit{static} risk perspective, where we treat the total~\textit{cumulative} cost to be the random variable of interest~\cite{chow2015risk}. 
Thus, risk is evaluated by applying a risk measure to the distribution over the cumulative cost.
The main drawback of the static perspective for sequential problems is that the evaluation of static risk is not~\textit{time-consistent}~\cite{boda2006time}. 
This means that if the risk of an action is evaluated to be riskier at one point time, then it is not necessarily evaluated to be more risky at all other points in time.
This leads to the optimal policy for static risk objectives being non-Markovian~\cite{bauerle2011markov, chow2015risk}.

The alternative approach is to consider the~\textit{dynamic} perspective of risk. 
In this approach, the risk measure is applied recursively at each point in time~\cite{ruszczynski2010risk}.
Thus, risk is evaluated by applying a risk measure to the distribution over outcomes at~\emph{every} time step.
A formal definition of static and dynamic risk can be found in the preliminaries section of Chapter~\ref{ch:7-1r2r}.

\begin{figure}[htb]
	\centering
	\includegraphics[width=7cm]{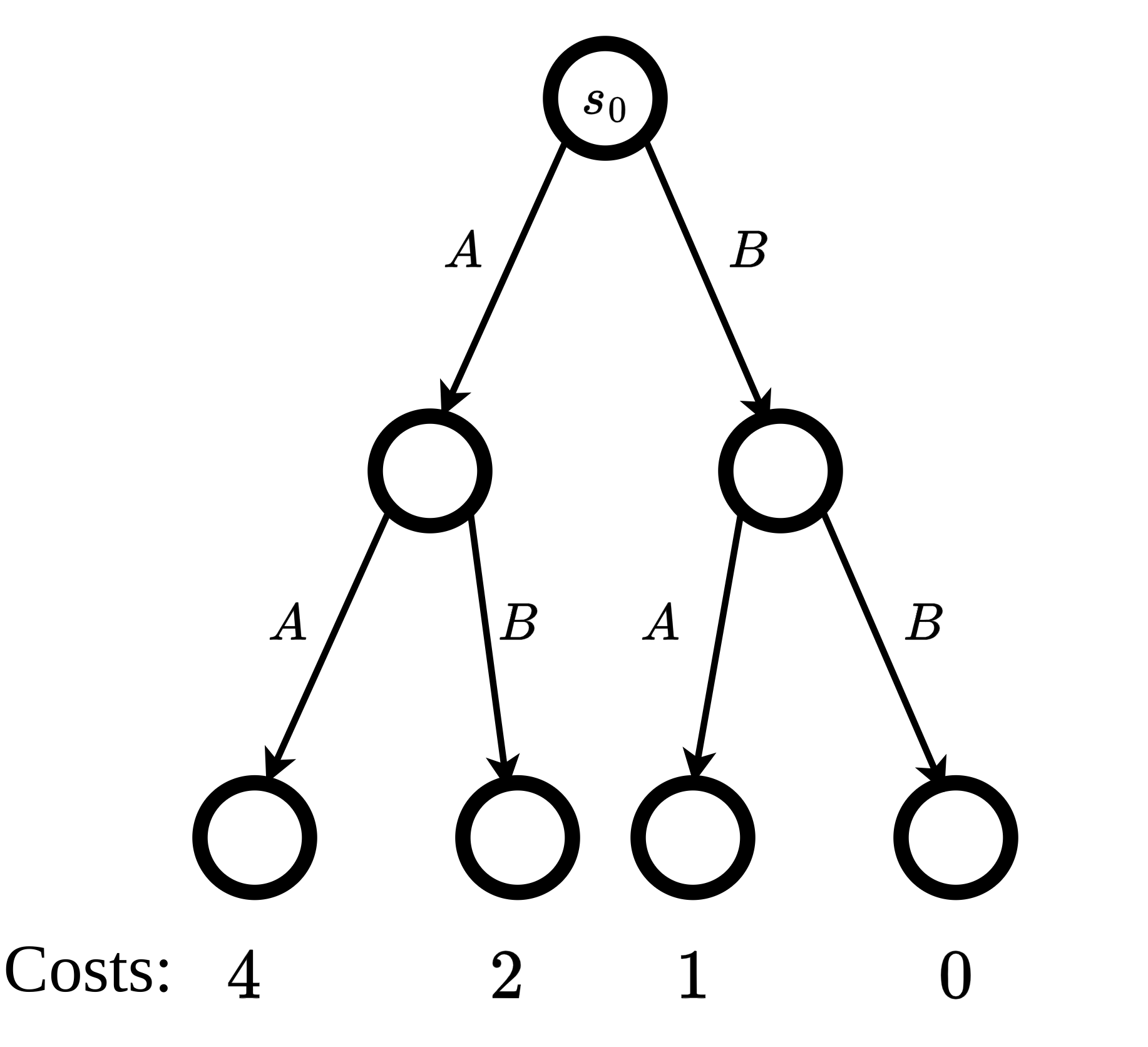}
	\caption{Illustration of a Markov chain (i.e. an MDP with no actions). At each state, transition $A$ or transition $B$ occurs with equal probability of 0.5. At the final state, the annotated cost is received. \label{fig:risk_comparison}}
\end{figure}

To illustrate the difference between static and dynamic risk, consider the conditional value at risk for the example illustrated in Figure~\ref{fig:risk_comparison}.
The static conditional value at risk is the expected value of the worst $\alpha$-portion of runs. 
In the example, reaching any of the four final states is equally likely. 
We can compute static conditional value at risk for $\alpha = 0.5$ by computing the average cost on the worst 50\% of runs.
The worst 50\% of runs are those which obtain a cost of either 2 or 4, and therefore the static conditional value at risk in the example is 3.
In contrast, the dynamic conditional value at risk for $\alpha = 0.5$ is the expected value assuming that the worst 50\% of transitions occur at each time step.
This is the expected value assuming that transition $A$ occurs at both time steps, and therefore the dynamic conditional value at risk for $\alpha=0.5$ is 4.

The advantage of the dynamic risk perspective is that it is time-consistent and therefore the optimal policy is always Markovian~\cite{ruszczynski2010risk}.
%
%
A drawback of dynamic risk is that because it is evaluated recursively, it compounds the risk of poor outcomes at every step.
Therefore, it can be much more conservative than the static risk approach and it is also more difficult to interpret~\cite{gagne2021two}.

\paragraph{Algorithms}
Algorithms for optimising dynamic risk include exact dynamic programming approaches~\cite{ruszczynski2010risk}, as well as approaches based on policy gradients~\cite{tamar2015policy}.
For the static CVaR setting, algorithms based on value iteration over an augmented state space have been proposed~\cite{bauerle2011markov, chow2015risk}.
Other works use policy gradient methods to find a locally optimal solution for CVaR~\cite{borkar2010risk, chow2014algorithms, tamar2014optimizing, tamar2015policy, prashanth2014policy, chow2017risk, tang2019worst}, or general coherent risk metrics~\cite{tamar2016sequential}.
A provably sample-efficient RL algorithm for optimising CVaR based on optimism in the face of uncertainty is presented in~\cite{keramati2020being}.

An alternative approach is taken by methods based on~\textit{distributional} reinforcement learning~\cite{bellemare2017distributional, morimura2010nonparametric}. Rather than estimating the expected value, these methods instead learn the distribution over the return.
With access to a prediction of the full return distribution associated with each action, the policy can then be optimised for a static risk objective~\cite{dabney2018implicit, ma2020dsac, urpi2021risk}.

\paragraph{Relevance to this thesis} In this thesis, we propose and address new formulations of risk-sensitive optimisation in both fully-known MDPs, and MDPs which are partially known.
The fully specified MDP case is considered in Chapter~\ref{ch:3-risk-mdps}, and the partially known MDP case is addressed in Chapters~\ref{ch:5-risk-bamdps} and~\ref{ch:7-1r2r}.

\section{Robustness in MDPs}
\label{sec:lit_robustness}

The previous section discussed risk-sensitivity towards the inherent stochasticity of the MDP.
This addresses scenarios where either the MDP is known (for planning), or unlimited experience can be gathered in the true MDP (for reinforcement learning).
However, in many cases we do not have access to the true MDP and have to optimise a policy using an imperfect approximation of the MDP.
Therefore, we may wish to mitigate epistemic (also called parametric) uncertainty due to inexact information about the MDP.

One approach to doing this is to find a robust solution that performs well in all possible instantiations of the uncertainty~\cite{ben2000robust}.
~\textit{Uncertain} MDPs~\cite{wolff2012robust} assume that the true transition function for the MDP is known to lie within an uncertainty set.
The~\textit{robust} solution to uncertain MDPs~\cite{nilim2005robust, iyengar2005robust, wolff2012robust}, also known as the Robust MDP approach, optimises for the best expected cost in the worst-case instantiation of the MDP within the uncertainty set.
This can be optimised efficiently with robust dynamic programming provided that: a) the uncertainty set is convex, and b) the uncertainty set is~\textit{rectangular}, meaning that the uncertainties  are independent between each state-action pair~\cite{iyengar2005robust, nilim2005robust, wiesemann2013robust} (see the preliminaries section of Chapter~\ref{ch:4-regret-mdps} for a more formal definition of rectangular uncertainties).
The first assumption enables the worst-case transition function to be found efficiently at each state-action pair, and the latter assumption means that the problem admits a dynamic programming decomposition.
Note that because the expected value is considered within each plausible MDP, this approach does not address aleatoric uncertainty.

This exact dynamic programming approach for Robust MDPs is applicable to problems with small state spaces. 
To scale up Robust MDPs to larger problems, algorithms which utilise reinforcement learning with function approximation have also been proposed~\cite{tamar2014scaling, mankowitz2020robust}.

Optimising for the worst-case expected value often results in overly conservative policies~\cite{delage2010percentile}.
This problem is exacerbated by the rectangularity assumption which allows the transition function for all state-action pairs to be realised as their worst-case values simultaneously. 
To avoid the assumption that all transition probabilities can take on their worst-case values, several alternative modelling approaches have been proposed. 
The ``lightning does not strike twice'' model~\cite{mannor2015lightning, mannor2016robust} assumes that the MDP parameter values deviate from nominal values a maximum number of times.
An alternative approach is to represent the uncertainty over the MDP using a discrete set of plausible MDPs~\cite{adulyasak2015solving, ahmed2013regret, ahmed2017sampling, chen2012tractable, cubuktepe2020scenario, steimle2018multi}. 
We refer to this approach as~\textit{sample-based} uncertain MDPs. 
This enables the uncertainty set to represent correlations between the uncertainties at different state-action pairs.
However, the sample-based approach drastically increases the computational complexity of the problem as it prohibits dynamic programming~\cite{ahmed2017sampling, steimle2018multi}.

Another option is to consider the~\textit{minimax regret} optimisation objective as an alternative to the worst-case objective discussed thus far~\cite{xu2009parametric}.
In this context, minimax regret refers to minimising the maximum expected sub-optimality of a single policy across any of the possible instantiations of the true MDP.
This is less conservative than the worst-case robust objective, as to achieve low maximum regret the policy must be near-optimal across all possible instantiations of the MDP.
In contrast, for the standard worst-case robust objective, the policy must only perform well in the worst-case MDP.

Regret is used to measure sample-efficiency in the context of RL theory (e.g.~\cite{jaksch2010near, cohen2020near, tarbouriech2020no}).
In the RL setting, the goal is usually to minimise the total cumulative regret, which is the total sub-optimality incurred throughout training over many episodes.
In contrast, the uncertain MDP setting assumes that the uncertainty is fixed: there is no opportunity to learn more about the MDP over many episodes.
The sub-optimality of the final policy is evaluated in expectation over a single episode, rather than accumulated throughout training.
Therefore, the uncertain MDP planning perspective is appropriate for problems where online exploration is not feasible, and the policy must be fixed offline.

Minimax regret in uncertain MDPs where only the cost function is uncertain is addressed in~\cite{regan2009regret, regan2010robust, regan2011robust, xu2009parametric}. 
For the case where the transition function is also uncertain,~\cite{ahmed2013regret} proposes to use the sample-based representation of the uncertain MDP.
The authors propose an exact solution based on solving a mixed integer linear program, however this approach does not scale.
The authors also propose an approximation based on dynamic programming, although for many problems the approximation is of low-quality~\cite{ahmed2017sampling}.

Thus far, all the approaches to optimising uncertain MDPs that we have discussed have been model-based.
Model-free approaches to the robust uncertain MDP problem have also been proposed~\cite{roy2017reinforcement, wang2021online, kuang2022learning}.
These approaches eliminate the need to specify the set of possible MDPs and instead assume that samples can be drawn from a misspecified
MDP which is known to be similar to the true MDP.

\paragraph{Robust Adversarial Reinforcement Learning (RARL)}
The uncertain MDP approach of specifying a set of plausible MDPs is impractical for large-scale problems. 
Robust Adversarial RL (RARL) proposes an alternative approach to optimising policies that are robust to worst-case uncertainties.
RARL is posed as a two-player zero-sum game where the agent plays against an adversary which perturbs the environment~\cite{pinto17robust}.
RARL alternates between optimising the policy of the agent to increase the expected reward, and optimising the policy of the adversary to decrease the expected reward.
In this manner, the adversary is trained to systematically find worst-case disturbances which decrease the performance of the agent policy.
By training in the presence of these worst-case disturbances, the agent learns to find a policy that is robust to the presence of worst-case disturbances.

Formulations of model-free RARL differ in how they define the action space of the adversary.
Options include allowing the adversary to apply perturbation forces to the simulator~\cite{pinto2017robust}, add noise to the agent's actions~\cite{tessler2019action}, or periodically take over control~\cite{pan2019risk}.

\paragraph{Relevance to this thesis}
In Chapter~\ref{ch:4-regret-mdps} we consider the minimax regret criterion for uncertain MDPs. 
We propose a new approximation for this problem, and a means to relax the assumption that the uncertainty is independent between each state-action pair.
In Chapter~\ref{ch:6-rambo} we consider a robust MDP formulation of offline RL. 
To arrive at a scalable solution, we propose a model-based algorithm which is inspired by Robust Adversarial RL. 
Our approach treats the environment model itself as the adversary to be trained.

\section{Bayes-Adaptive Reinforcement Learning}
\label{sec:lit_bamdps}

In the previous section, we discussed methods for finding policies that are robust to worst-case uncertainty over the MDP.
An alternative approach is to instead find a policy which can~\textit{adapt} to the true underlying MDP.

The model-based Bayesian formulation of reinforcement learning~\cite{ghavamzadeh2015bayesian} assumes that there is a distribution over the transition function of the underlying MDP.
One way of modelling this problem is to consider the~\textit{Bayes-adaptive} MDP (BAMDP)~\cite{duff2002optimal}. 
The BAMDP augments the state space of the original MDP with the belief distribution over the underlying MDP. 
By integrating over the distribution of plausible transition functions, the transition function between augmented states in the BAMDP can be computed.
After observing a transition, the belief over the MDP is updated.
This forms a type of belief MDP~\cite{kaelbling1998planning}, and therefore the BAMDP can be viewed as a specific type of POMDP.

Acting optimally in the BAMDP results in an optimal tradeoff between exploration and exploitation: exploratory actions are only taken if they improve the expected return over the problem horizon, given the prior over MDPs.
This property is referred to as Bayes-optimality.
Thus, Bayes-adaptive MDPs can be thought of as bridging planning and reinforcement learning: like RL, the goal is to tradeoff exploration and exploitation to learn the best behaviour in an unknown MDP, however the BAMDP represents this as a planning problem in a known (belief) MDP~\cite{poupart2006analytic}. 

Exact solutions to BAMDPs based on value iteration do not scale beyond the very smallest of problems~\cite{poupart2006analytic}.
Algorithms based on Monte Carlo tree search for POMDPs~\cite{silver2010monte} have been adapted to BAMDPs~\cite{guez2012efficient}.
These methods utilise~\textit{root sampling}, in which an MDP instantiation is sampled from the root belief at each trial.
The sampled MDP is used to simulate each trial, and this approach results in the correct distribution over belief states in the BAMDP~\cite{guez2012efficient}. 
Reinforcement learning algorithms have been adapted to BAMDPs, and also utilise root sampling~\cite{guez2014bayes, zintgraf2020varibad}. 

One of the challenges of this approach is modelling the posterior distribution over MDPs.
Earlier efforts considered Dirichlet distributions over transition probabilities~\cite{poupart2006analytic}, or Gaussian processes~\cite{guez2014bayes}.
VariBAD~\cite{zintgraf2020varibad} proposes to use deep latent variable models to learn to approximate the posterior distribution over MDPs, resulting in a more scalable approach.

Other approaches to the Bayesian RL problem do not attempt to generate Bayes-optimal behaviour and do not consider the BAMDP model.
One line of work is based on Thompson sampling~\cite{osband2013more, russo2018tutorial}.
In this approach, for each episode an MDP is sampled and the agent acts optimally with respect to that MDP for the entire episode. 
The posterior over MDPs is refined between episodes, rather than at each step per the BAMDP approach.
Another approach is to apply exploration bonuses which depend on the level of uncertainty in the model for that state-action pair~\cite{kolter2009near, sorg2010variance}. 
%


\paragraph{Relevance to this thesis}
In Chapter~\ref{ch:5-risk-bamdps} we propose to optimise for risk in the BAMDP model to address both epistemic uncertainty and aleatoric uncertainty under the same framework.
By avoiding risk in a BAMDP, the risk of a poor outcome occurring due to either: a) a lack of knowledge about the MDP,  or b) the inherent stochasticity of the MDP, are both mitigated.

\section{Offline Reinforcement Learning}
\label{sec:offline_rl}
In this section, we first provide an overview of the motivation and problem formulation of offline reinforcement learning. 
We then give an overview of the algorithmic challenges and existing approaches to the problem.

\subsection{Motivation and Problem Setting}
One of the major limitations of reinforcement learning is the need for extensive online exploration.
For many applications, exploration can be expensive or dangerous~\cite{garcia2015comprehensive}, making RL impractical.
For applications where online exploration is not possible, a reasonable alternative is to instead leverage large pre-existing datasets to optimise policies.

~\textit{Offline} (or batch) RL~\cite{lange2012batch, levine2020offline} refers to the setting where we only have access to a fixed dataset.
The goal is to find a performant policy using this data alone.
Researchers have applied offline RL to domains such as healthcare~\cite{nie2021learning, shortreed2011informing}, natural language processing~\cite{kandasamy2017batch, jaques2019way}, and robotics~\cite{kumar2021workflow, mandlekar2020iris, rafailov2021offline}

In contrast to learning from demonstration, or imitation learning~\cite{argall2009survey}, there is no assumption that the data is collected by an expert policy.
However, unlike imitation learning, offline RL requires that the dataset is labelled with rewards to direct the learnt policy towards the desired behaviour. 
Therefore, offline RL is suitable for applications where the data that has been collected is noisy, and the reward function for the desired behaviour can easily be specified.

A drawback of offline RL is that the best achievable performance  can be considerably worse than the optimal performance for the MDP: if the dataset does not cover high-value regions of the state space relevant to optimal behaviour, one cannot hope to obtain a performant policy~\cite{xiao2022curse}.
A middle-ground between online and offline RL is the notion of~\textit{deployment efficient} RL~\cite{matsushima2020deployment} (also called RL with a low switching cost~\cite{bai2019provably}). 
In this problem formulation, the number of times a new policy is a deployed for data collection is limited. 
This is motivated by the consideration that deploying each new policy might be a  costly or risky procedure.
For example, each new policy may require safety validation prior to deployment.
Reducing the number of times the policy is switched lessens this burden, and makes training an RL agent in the real world more practical.

Another possibility is to use offline RL to initialise a policy, and then allow a limited amount of online exploration to fine-tune the policy~\cite{kalashnikov2018scalable, kostrikov2021offline, xie2021policy}.
This approach can substantially reduce the amount of online exploration required to find a near-optimal policy~\cite{kalashnikov2018scalable}.

In the remainder of this section we focus on algorithms for the fully offline formulation of RL where no further data collection is possible.
This is the most suitable approach domains where any form of random exploration in the real environment is not possible, and we do not have access to a sufficiently accurate simulator.

\subsection{Algorithms}
Offline RL poses several challenges that necessitate the design of specific algorithms.
Many reinforcement learning algorithms can learn from~\textit{off-policy} data, i.e. data that was collected by a different policy.
However, in practice, off-policy online RL algorithms do not perform well in the offline setting.

The main challenge of offline RL is the issue of distributional shift~\cite{levine2020offline}.
While the policy, value function, or model might be trained accurately under one distribution (the distribution of data in the dataset), it will be evaluated under a different distribution (the distribution generated by the output policy in the real environment). 
For example, if a value function is trained to achieve low Bellman error over the dataset, this does not mean that the value function is accurate  under the data distribution generated by the new policy. 

This issue is exacerbated by the use of function approximation, and the fact that the policy is trained to maximise the return.
Consider again the perspective of value-based methods.
The Q-values for some state-action pairs will inevitably be over-estimated. 
The policy is trained to maximise the value function, and therefore the policy learns to systematically exploit the largest errors in the value function. 
In the online setting, this is not an issue as the distribution of state-action pairs generated by the policy are sampled in the environment, and this experience is used to correct inaccuracies.
However, in the offline setting these inaccuracies are not corrected, and this can result in drastically over-optimistic value estimates and policies that perform poorly in the real environment~\cite{kumar2019stabilizing}.

For model-based approaches, which learn a dynamics model from the dataset,  there is the analogous issue of~\textit{model exploitation}. The policy learns to select actions that exploit errors in the learnt model, rather than actions that are likely to perform well on the real environment, as evidenced by the dataset~\cite{clavera2018model, kidambi2020morel, kurutach2018model}.

As a result of these issues, the na\"ive application of off-policy online RL algorithms generally does not perform well in the offline setting, and may produce erroneous actions when tested in the real environment.
Earlier works on offline RL do not attempt to avoid the aforementioned issues~\cite{ernst2005tree, lagoudakis2003least, riedmiller2005neural}. 
More recently, the focus of offline RL research has been  to develop algorithms to mitigate this issue of distributional shift. 
We will focus on these more recent works and give an overview of both model-free and model-based approaches.

\paragraph{Model-free offline RL algorithms}
The simplest approach to model-free offline RL is to apply an online RL algorithm, but with the additional constraint that the learnt policy is similar to the behaviour policy that generated the dataset~\cite{fujimoto2019off, kostrikov2022offline, wu2019behavior, jaques2019way, siegel2020keep, fujimoto2021minimalist}.
Most methods encourage the learnt policy to approximately match the distribution over actions generated by the behaviour policy.
However, this can be highly restrictive and result in suboptimal behaviour.
Therefore,~\cite{kumar2019stabilizing} argues for constraining the learnt policy to take actions in the~\textit{support} of the behaviour policy. 
However, support constraints generally require approximations to implement~\cite{wu2022supported}.
Therefore, subsequent works have shown that other techniques for ensuring that the policy does not leave the data distribution work better in practice~\cite{kumar2020conservative}.

Another approach is to incorporate conservatism into the value function update to prevent overestimation of the value of state-action pairs that are not present in the dataset~\cite{cheng2022adversarially, kumar2020conservative, xie2021bellman}. For example, the value function update for CQL~\cite{kumar2020conservative} penalises the Q-values of state-action pairs that are not contained in the dataset.

Another line of work uses uncertainty quantification to obtain more robust value estimates~\cite{agarwal2020optimistic, an2021uncertainty, kumar2019stabilizing}.
The approach proposed by~\cite{an2021uncertainty} is inspired by clipped double Q-learning~\cite{fujimoto2018addressing}, which uses the minimum over two separately initialised Q-functions as the final Q-value estimate.
 However,~\cite{an2021uncertainty} instead uses the minimum over a larger ensemble of many Q-networks to arrive at a pessimistic estimate of the value function in the true environment.

One-step offline RL approaches avoid the need for off-policy evaluation by only performing a single iteration of policy iteration~\cite{brandfonbrener2021offline, peng2019advantage}.
The Q-function is estimated for the behaviour policy, and policy improvement is performed under this Q-function.
This approach works reasonably well in comparison to approaches which interleave policy evaluation and improvement~\cite{brandfonbrener2021offline}. 
The justification for one-step approaches is that iterative approaches propagate errors through the Bellman equation.
This issue of compounding error is avoided by performing policy iteration only once.

A final class of model-free offline RL algorithms are approaches based on policy gradients with importance sampling that are tailored towards the fully offline setting~\cite{liu2019off, nachum2019algaedice}.

\paragraph{Model-based offline RL algorithms}
Model-based approaches learn a model of the environment and generate synthetic data from that model~\cite{sutton1991dyna} to optimise a policy using either planning~\cite{argenson2021modelbased} or RL algorithms~\cite{kidambi2020morel, yu2020mopo, yu2021combo}.
By training a policy on additional synthetic data, model-based approaches can potentially generalise more broadly, or be more easily adapted to new tasks~\cite{ball2021augmented, yu2020mopo}

 A simple approach to prevent model exploitation is to constrain the policy to be similar to the behaviour policy, in the same fashion as some model-free approaches~\cite{cang2021behavioral,  matsushima2020deployment, swazinna2021overcoming}. 
However, like the model-free case, such constraints may prevent the algorithm from finding the best policy covered by the dataset.
 
 Another approach is to apply reward penalties for executing state-action pairs with high uncertainty in the environment model~\cite{kidambi2020morel, yang2021pareto, yu2020mopo}. 
 If the uncertainty estimate is accurate and the reward penalties are defined appropriately, this optimises a lower bound on the value function in the true environment\cite{kidambi2020morel, yu2020mopo}.
 However, in practice the uncertainty estimates for neural network models are often unreliable~\cite{gawlikowski2021survey, lu2022revisiting,  ovadia2019can, yu2021combo}. 
 COMBO~\cite{yu2021combo} obviates the need for uncertainty estimation in model-based offline RL by adapting model-free techniques~\cite{kumar2020conservative} to regularise the value function for out-of-distribution samples.

Most approaches to model-based offline RL use maximum likelihood estimates of the MDP trained using standard supervised learning~\cite{argenson2021modelbased,matsushima2020deployment, swazinna2021overcoming, yu2020mopo,yu2021combo}. 
However, other methods have been proposed to learn models which are more suitable for offline policy optimisation. 
One approach is to reweight the loss function for training the model to ensure the model is accurate under the distribution of states and actions generated by the policy~\cite{lee2020representation, rajeswaran2020game, hishinuma2021weighted}. 
This reduces the possibility that the policy can learn to exploit errors in the model. 
A~\textit{reverse} model is learnt by~\cite{wang2021offline}, which predicts the dynamics of the MDP backwards in time. 
The motivation for doing this is that it enables the generation of synthetic rollouts that always end at states which are within the dataset.
This encourages the policy to learn corrective actions that stay within the distribution of the dataset.
This stands in contrast to forward models which may generate trajectories that leave the dataset distribution. 

\paragraph{Relevance to this thesis} We propose two new algorithms for model-based offline RL in Chapters~\ref{ch:6-rambo} and~\ref{ch:7-1r2r}.
In Chapter~\ref{ch:6-rambo}, we formulate the problem of offline RL as a Robust MDP, and propose an algorithm inspired by Robust Adversarial RL.
In this instantiation of Robust Adversarial RL, we consider the adversary to be the environment model itself.
The model is adversarially trained to generate pessimistic synthetic transitions for state-action pairs which are out of the dataset distribution.
This is a novel approach for preventing model exploitation without using uncertainty estimation or policy constraints.

In Chapter~\ref{ch:7-1r2r} we propose to optimise for risk-aversion to solve the problem of offline RL.
Risk-aversion to epistemic uncertainty solves the problem of model exploitation and distributional shift by discouraging the policy from visiting areas that are not covered by the dataset, where the epistemic uncertainty in the model is high.
Our approach also achieves risk-aversion to aleatoric uncertainty under the same framework.
Unlike Chapter~\ref{ch:6-rambo}, the approach in Chapter~\ref{ch:7-1r2r} requires uncertainty estimation to quantify the uncertainty in the model.

\section{Learning from Demonstration}
\label{sec:lfd}

As discussed in the previous section, in many domains, learning a policy entirely from scratch using RL is infeasible due to the need for extensive exploration.
An alternative approach is to utilise expert demonstrations to learn a policy~\cite{schaal1996learning}.
This is referred to as either learning from demonstration, or imitation learning.

The two major paradigms of learning from demonstration are behaviour cloning, and inverse reinforcement learning~\cite{osa2018algorithmic}. 
On the one hand, behaviour cloning directly imitates the actions in the dataset using supervised learning~\cite{pomerleau1991efficient}.
On the other hand, inverse reinforcement learning finds a reward function for which the policy of the expert is optimal~\cite{ziebart2008maximum}.

Even with expert demonstrations, the performance of behaviour cloning can be poor due to errors compounding as soon as the learner makes any mistake and begins to deviate from the distribution of states present in the demonstrations~\cite{ross2010efficient}. 
A policy learned via behaviour cloning does not know how to recover from such situations.
This issue of compounding error is avoided by inverse RL methods~\cite{ho2016generative}. 
However, inverse RL approaches usually require the MDP to be solved for the current reward function in an inner loop. 
This may require running RL in an inner loop, so this approach can be very inefficient and still require extensive exploration of the environment~\cite{osa2018algorithmic}.

To mitigate the issue of compounding error in behaviour cloning, DAGGER~\cite{ross2011reduction} proposes to collect more expert actions under the distribution of states generated by the current policy.
This ensures that the learner sees corrective actions provided by the expert.
If the reward function for the environment is known, another approach is to use learning from demonstration to initialise a policy, which is then fine-tuned using RL.
Initialising a policy using demonstrations can drastically reduce the amount of exploration required for RL to find a near-optimal policy~\cite{kober2009learning, sun2017deeply}.

In a similar vein to DAGGER, other works have proposed approaches to request expert demonstrations only if they are necessary.
The ``ask for help'' framework introduced by~\cite{clouse1996integrating} only asks for human input when the learner is uncertain.
In~\cite{clouse1996integrating}, the agent nominally performs tabular Q-learning. If all actions in a state have a similar Q-value, the agent is deemed uncertain, and asks the human which action to take. 
In~\cite{chernova2009interactive}, the authors consider policies defined by a classifier.
For states near a decision boundary of the classifier, a demonstration is requested. 
In \cite{del2018not}, Gaussian process policies are learnt from demonstrations. A demonstration is requested if the variance of the output of the Gaussian process policy is too high.
All of these approaches require a hand-tuned threshold for uncertainty.

~\paragraph{Relevance to this thesis} In Chapter~\ref{ch:8-lfd}, we consider an interactive learning from demonstration setting where expert demonstrations can be requested throughout training.
Rather than developing a new algorithm for learning from demonstration, we focus on a new problem formulation.
We assume that there is some cost associated with the ``human effort'' required to recover the system from a failed episode, as well as to provide a demonstration. 
Based on this cost function, the goal is to minimise the human effort required throughout the training process. 
Unlike many of the aforementioned approaches, this avoids the need to hand-tune a threshold of uncertainty at which demonstrations should be requested.
Instead, our approach uses the human cost objective to determine when it is worthwhile to request a demonstration.

Furthermore, we use behaviour cloning to initialise the policy prior to applying the offline RL algorithm we propose in  Chapter~\ref{ch:6-rambo}.
We found that initialising the policy in this manner resulted in a small performance improvement.

%% file: text/ch3-risk-mdps.tex
\chapter{\label{ch:3-risk-mdps}Planning for Risk-Aversion and Expected Value in MDPs}

In this chapter, we consider the case where the MDP is known exactly. 
This is suitable for domains where there is substantial historical data that can be used to estimate the transition and reward function of the MDP, and it is safe to assume that the MDP will not change over time.
We consider risk-aversion to the inherent stochasticity in the MDP, i.e. aleatoric uncertainty.

In this work, we presume that the objective is to optimise the static conditional value at risk (CVaR) of the total cost received. 
Static CVaR for some confidence level, $\alpha$, is the mean of the worst $\alpha$-portion of total costs.
As discussed in Section~\ref{sec:lit_risk}, CVaR is commonly used in machine learning and robotics research due to the fact that it is coherent, and easy to interpret.

In this work, we focus on the tradeoff between optimising for a risk-sensitive objective and the expected performance.
For most domains, risk-sensitive optimisation results in more conservative behaviour and worse performance in expectation than risk-neutral approaches. 
We develop an algorithm to obtain the best possible expected value subject to the constraint that CVaR remains optimal.
This approach ensures that the optimal risk-sensitive behaviour is maintained, but the expected value is improved where possible as a secondary objective.

Our approach builds upon an existing planning approach for CVaR~\cite{chow2015risk} that can be viewed as a two-player zero-sum game.
Our algorithm first executes a policy that is optimal for CVaR. 
Note that CVaR is the expectation of total costs that exceed the value at risk (VaR). 
If, during execution, a stage is reached where there exists a policy that is guaranteed not to exceed the VaR, then we switch to executing the policy with the best expected value that is guaranteed not to exceed the VaR.
Our approach only modifies the distribution over returns that are less than the VaR, and therefore the optimal CVaR is maintained.

We evaluate this approach on several domains, including a domain based on real-world driving data, and show that we can significantly improve the expected cost obtained compared to an existing state-of-the-art algorithm, while maintaining optimal CVaR performance.

The summary in Table~\ref{tab:summary}, which we have repeated below, shows that this chapter considers the most restrictive setting in this thesis. 
It assumes full knowledge of the MDP, and is only applicable to tabular MDPs.
The risk-sensitive objective addressed in this chapter takes into consideration aleatoric uncertainty.
However, because the MDP is assumed to be fully known, this approach does not take into consideration epistemic uncertainty.

\begin{table}[h!tb]
	\renewcommand{\arraystretch}{1.3}
	\hyphenpenalty=100000
	\footnotesize
	\caption*{\textbf{Table~\ref{tab:summary}:} \tablecaption }
	\resizebox{\columnwidth}{!}{%
		\begin{tabular}{l|a{1.7cm}|b{1.7cm}|b{1.7cm}|b{1.7cm}|b{1.7cm}|b{1.9cm}|}
			\hhline{~|-|-|-|-|-|-|}
			&  \textbf{Chapter~\ref{ch:3-risk-mdps}} &  \textbf{Chapter~\ref{ch:4-regret-mdps}}  & \textbf{Chapter~\ref{ch:5-risk-bamdps}} & \textbf{Chapter~\ref{ch:6-rambo}} & \textbf{Chapter~\ref{ch:7-1r2r}} & \textbf{Chapter~\ref{ch:8-lfd}} \\ \hline
			\multicolumn{1}{|m{1.9cm}|}{\textbf{Prior knowledge}} & MDP fully known & MDP uncertainty set & Prior over MDPs & Fixed dataset & Fixed dataset & Expert demonstrator \\ \hline
			\multicolumn{1}{|m{1.9cm}|}{\textbf{Considers aleatoric uncertainty}} & Yes & No & Yes &  No & Yes & No \\ \hline
			\multicolumn{1}{|m{1.9cm}|}{\textbf{Considers epistemic uncertainty}} & No & Yes & Yes & Yes & Yes & Yes \\ \hline
			\multicolumn{1}{|m{1.9cm}|}{\textbf{Risk or robustness}} & Risk &  Robustness  & Risk & Robustness & Risk & N/A \\ \hline
			\multicolumn{1}{|m{1.9cm}|}{\textbf{Paradigm}} & Tabular &  Tabular  & Tabular & Function Approx. & Function Approx. & Function Approx. \\ \hline
		\end{tabular}
	}%
\end{table}

\FloatBarrier

\noindent\rule{6cm}{0.5pt} 

\medskip
\noindent Marc Rigter, Paul Duckworth, Bruno Lacerda and Nick Hawes (2022). Planning for risk-aversion and expected value in MDPs. \textit{International Conference on Automated Planning and Scheduling} (ICAPS).
\medskip

\includepdf[pages=-]{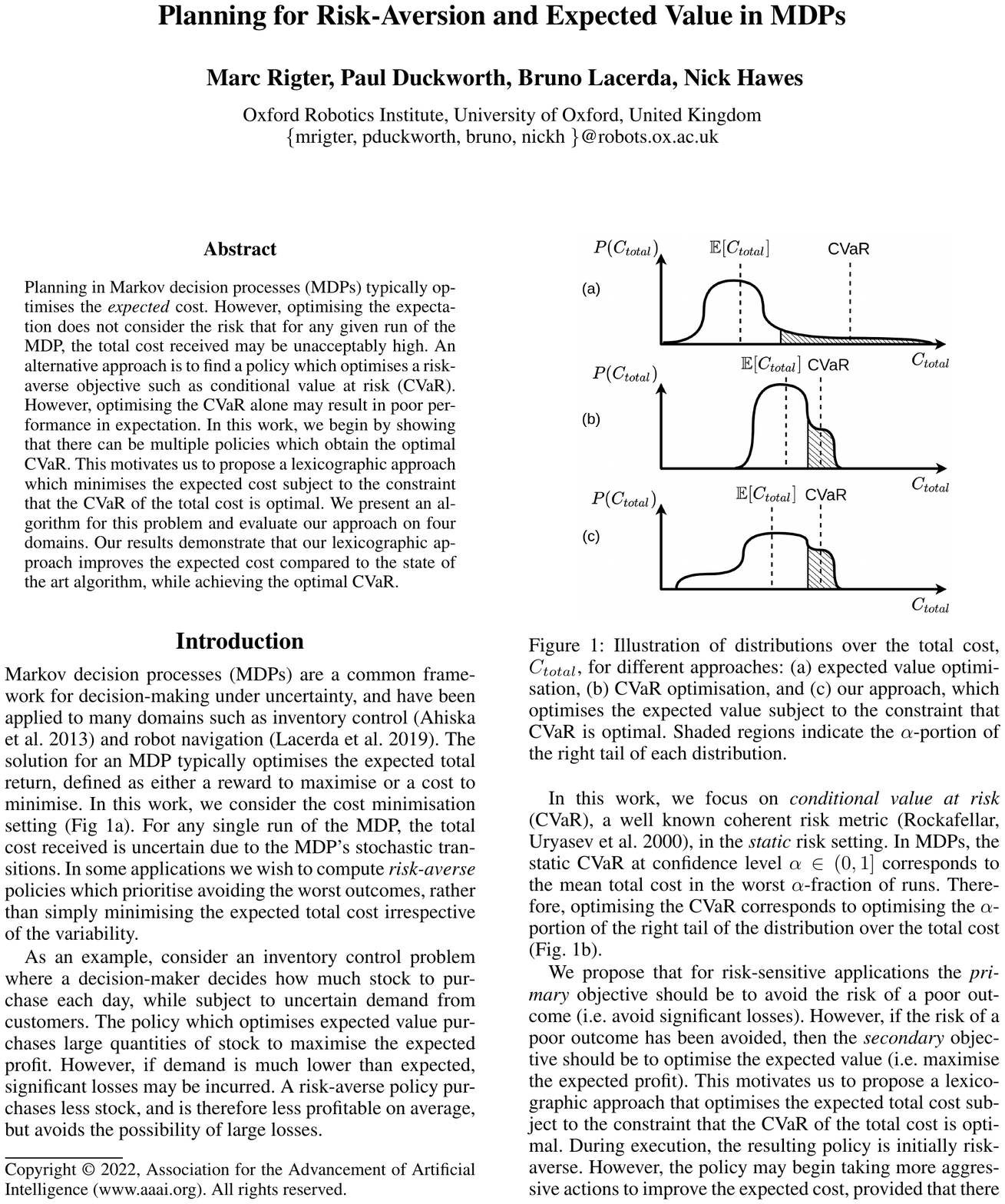}

\section{Limitations and Future Work}
There are several limitations to this work that prevent it from being more widely applicable.
First, improving the expected value of the policy while maintaining the optimal CVaR is not possible for all problems.
Our approach requires that for some reachable histories, it will be possible to take an alternate course of action that guarantees that the VaR of the original policy will not be exceeded.
Whether this is possible depends on the structure of the specific problem.
For example, we found that for some start-goal pairs of the autonomous navigation domain, our method resulted in no improvement.
As seen in the paper, the improvement is only modest for the final version of the autonomous navigation domain that we used.

To guarantee that the CVaR remains optimal, our approach relies on exact solutions to the MDP that iterate over the entire state space.
Therefore, it is unclear how our approach can be combined with approximate methods for solving MDPs such as sample-based search or reinforcement learning, where we will not be able to provide such a guarantee.
Therefore, when considering more scalable approaches, it may be more sensible to focus on weighted combinations of CVaR and expected value~\cite{petrik2012an}, or risk measures that do not neglect the expected value~\cite{wang2000class}.

This research hints at a direction for possible future work.
We have shown that by switching between two policies that optimise for different objectives, we are able to derive an overall strategy that performs better for our chosen objective.
An interesting direction for future work could be to investigate other approaches for composing together different policies with different objectives.
For example, one could imagine optimising several policies for different levels of risk-aversion, and switching between the policies in a hierarchical manner.
It will be a challenging question to determine in which situations, and for which optimisation objectives, better performance can be achieved by systematically  switching between the policies with different levels of risk-aversion.


%% file: text/ch4-regret-mdps.tex

\chapter{\label{ch:4-regret-mdps} Minimax Regret Optimisation for Robust Planning in Uncertain Markov Decision Processes}
\chaptermark{Minimax Regret Optimisation for Robust Planning}

\begin{table}[b!]
	\renewcommand{\arraystretch}{1.3}
	\hyphenpenalty=100000
	\footnotesize
	\caption*{\textbf{Table~\ref{tab:summary}:} \tablecaption }
	\resizebox{\columnwidth}{!}{%
		\begin{tabular}{l|b{1.7cm}|a{1.7cm}|b{1.7cm}|b{1.7cm}|b{1.7cm}|b{1.9cm}|}
			\hhline{~|-|-|-|-|-|-|}
			&  \textbf{Chapter~\ref{ch:3-risk-mdps}} &  \textbf{Chapter~\ref{ch:4-regret-mdps}}  & \textbf{Chapter~\ref{ch:5-risk-bamdps}} & \textbf{Chapter~\ref{ch:6-rambo}} & \textbf{Chapter~\ref{ch:7-1r2r}} & \textbf{Chapter~\ref{ch:8-lfd}} \\ \hline
			\multicolumn{1}{|m{1.9cm}|}{\textbf{Prior knowledge}} & MDP fully known & MDP uncertainty set & Prior over MDPs & Fixed dataset & Fixed dataset & Expert demonstrator \\ \hline
			\multicolumn{1}{|m{1.9cm}|}{\textbf{Considers aleatoric uncertainty}} & Yes & No & Yes &  No & Yes & No \\ \hline
			\multicolumn{1}{|m{1.9cm}|}{\textbf{Considers epistemic uncertainty}} & No & Yes & Yes & Yes & Yes & Yes \\ \hline
			\multicolumn{1}{|m{1.9cm}|}{\textbf{Risk or robustness}} & Risk &  Robustness  & Risk & Robustness & Risk & N/A \\ \hline
			\multicolumn{1}{|m{1.9cm}|}{\textbf{Paradigm}} & Tabular &  Tabular  & Tabular & Function Approx. & Function Approx. & Function Approx. \\ \hline
		\end{tabular}
	}%
\end{table}

In the previous chapter, we assumed that the MDP representing the environment was known exactly.
For the remaining chapters of this thesis, we no longer make such a strong assumption.
As shown by Table~\ref{tab:summary}, this, and all subsequent chapters, consider epistemic uncertainty.
This means that we explicitly take into account a lack of knowledge about the environment.

The characteristics of the problem setting in this chapter are summarised in Table~\ref{tab:summary}.
In this chapter, we represent the uncertainty over the MDP in the form of a discrete set of possible MDPs.
The true MDP is assumed to be one of the MDPs within this set.
We consider the minimax regret objective, meaning that our goal is to find a~\emph{single} policy with the lowest maximum sub-optimality across the set of possible MDPs.
Here, sub-optimality in each MDP refers to the expected value of the policy compared to the optimal expected value for each MDP.
To achieve low maximum regret the policy must be near-optimal, in terms of expected value, in all of the plausible MDPs.
Because the expected value is considered within each MDP, the approach presented in this chapter does not consider aleatoric uncertainty.
Furthermore, because this work considers the~\emph{worst-case} sub-optimality in the uncertainty set, it achieves robustness to all scenarios in the uncertainty set.

We propose a dynamic programming approach for approximating the minimax regret.
We first solve for the optimal expected value in each plausible MDP.
Then, for each MDP we treat the difference between the optimal Q-value for a given action, and the optimal expected value, (i.e. the ``advantage'': $Q^*(s, a) - V^*(s)$) as the cost incurred for executing the action in that MDP.
This cost represents the regret associated with each action in each MDP.
Our approach performs minimax optimisation of this regret-based cost as an approximation for the minimax regret.
To perform this optimisation, we use robust dynamic programming~\cite{iyengar2005robust, nilim2005robust} with the regret-based cost that we propose.\footnote{In the original publication, we claimed that this approach was exact for independent/rectangular uncertainty sets. The authors later realised that this is incorrect, and this is indicated by the footnotes in the paper and in corrigendum included in the appendices for this paper. The authors have contacted AAAI to resolve this issue.}

Dynamic programming approaches to this type of minimax optimisation problem require that the MDP uncertainties are independent between each state-action pair (this type of uncertainty set is also called ``rectangular'').
However, for many problems there exist correlations between the uncertainties.
For example, consider navigating the ocean in the presence of unknown currents. 
Knowing the direction of the current at one state informs what the current will be like at nearby states.
However, these correlations cannot be modelled due to the independence assumption.

To help overcome this limitation we propose to define the policy using options, where each option is executed for $n$-steps.
Within each option, dependencies between uncertainties are captured.
In the ocean navigation problem, this means that within each option the direction of the uncertain ocean current is fixed.
This requires solving a mixed integer linear program to optimise minimax regret for each option.
Between options, the uncertainties are assumed to be independent so that we can utilise dynamic programming and solve for each option separately.
Depending on the value of $n$, this approach enables more realistic modelling of uncertainty at the expense of greater computation requirements.

We test our approach on problems where there are dependencies between the uncertainties, including a domain based on real ocean current forecasts. 
The results demonstrate that the method we propose outperforms existing approaches to the problem formulation addressed in this work.
Additionally, the performance improves as the number of steps in each option is increased. 
This validates that our options-based approach enables more accurate modelling of the correlations between uncertainties.

\noindent\rule{6cm}{0.5pt} 

\medskip
\noindent  Marc Rigter, Bruno Lacerda, and Nick Hawes (2021). Minimax regret optimisation for robust planning in uncertain Markov decision processes. \textit{AAAI Conference on Artificial Intelligence} (AAAI).
\medskip

\includepdf[pages=-]{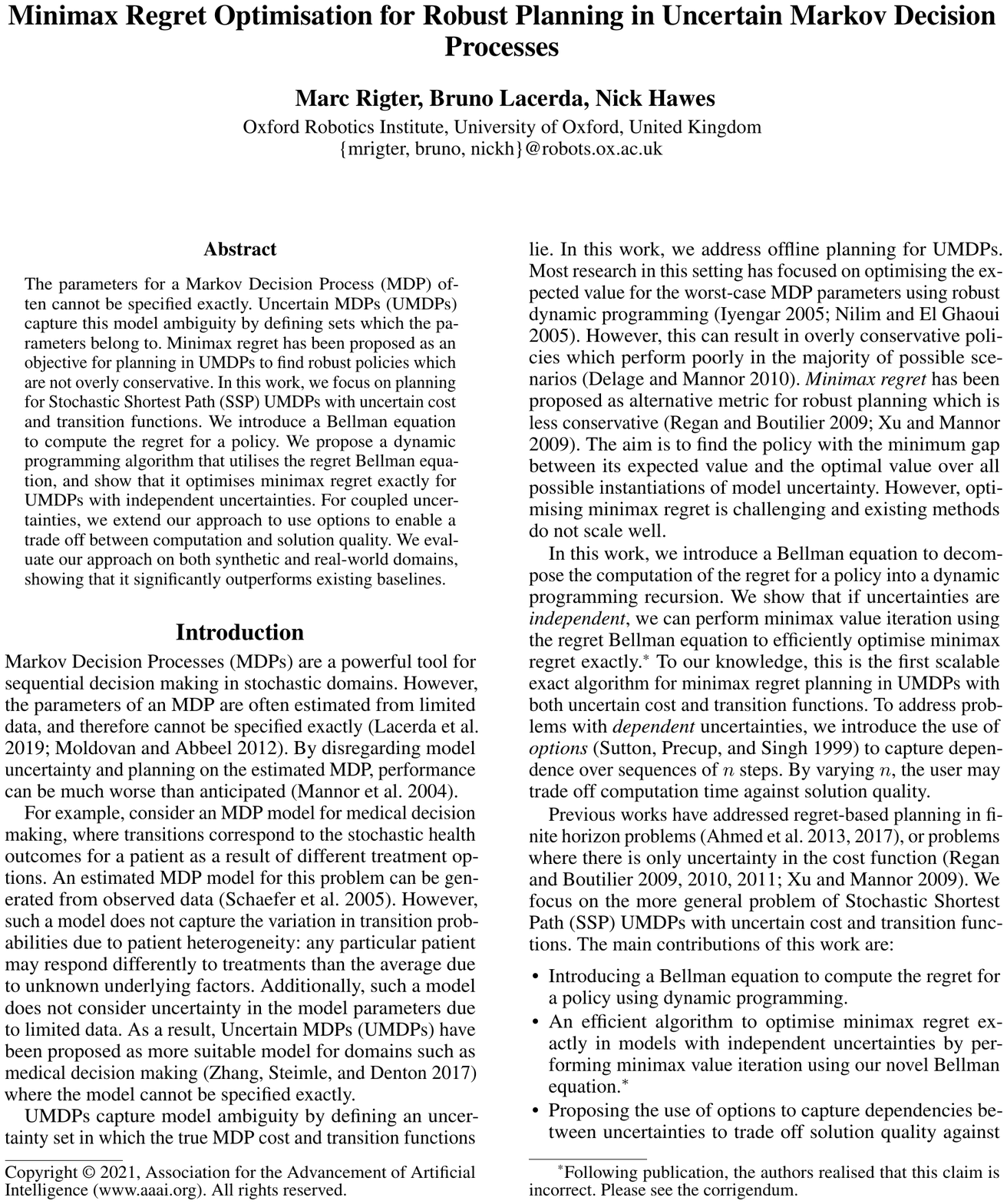}

\section{Limitations and Future Work}
\label{sec:minimax_reg_future_work}
A key issue with this piece of work is that some of the theoretical analysis presented in the original version of this paper was later found to be incorrect, as indicated in the footnotes, and in the corrigendum included in the appendices.
Specifically, the approach we proposed does not give an exact solution for minimax regret for independent/rectangular uncertainty sets.
As noted, we have contacted the publisher to amend this.
Thankfully, this does not appear to be a major shortcoming as the independent uncertainty assumption is usually unrealistic, and problems with independent uncertainties are not the focus of this work.
Furthermore, our experimental results show that the approach we proposed empirically outperforms existing approaches.

One of the limitations of our approach is that finding each option policy requires solving a mixed integer linear program. 
This prevents our method from scaling to longer option policies that capture greater dependencies between the uncertainties, as these sub-problems become too complex to solve.

An obvious next step for this work would be to try and apply the proposed ideas to larger problems by utilising methods based on reinforcement learning with function approximation.
Rather than specifying an MDP uncertainty set, for more complex problems it would be more natural to consider some unknown parameter(s) that govern the dynamics.
For example, in robotics an example of an unknown parameter could be the friction coefficient for the contact between the robot and the ground. 

To adapt our approach for approximating minimax regret, we could first find the optimal Q-function for a number of samples of the unknown parameter using reinforcement learning.
Then, we could optimise a final policy to minimise the regret-based cost that we proposed in this work.
The regret-based cost would be computed by evaluating the maximum sub-optimality of each action across the optimal Q-functions for each of the plausible parameter values.
Optimising a policy based on our proposed regret-based cost would approximate the minimax regret criterion.
The first step of finding the optimal Q-function for a range of parameter values would be computationally demanding for large problems.
However, this could be alleviated by including the parameter value in the state space, and learning a single Q-function, thereby enabling generalisation between different parameter values.

It would be interesting to investigate whether this proposed approach outperforms RL approaches that optimise for the best expected value in the worst-case scenario~\cite{pinto2017robust, tamar2014scaling, mankowitz2020robust}, that might suffer from being overly conservative.

This proposed approach would implicitly assume that the uncertainties are independent between each step, allowing for the value of the unknown parameter to change each step.
To alleviate this assumption, it would be interesting to investigate whether the idea that we proposed to utilise options with fixed uncertainty can be adapted to the deep RL setting.


%% file: text/ch5-risk-bamdps.tex
\chapter{\label{ch:5-risk-bamdps} Risk-Averse Bayes-Adaptive Reinforcement Learning}

\begin{table}[b!]
	\renewcommand{\arraystretch}{1.3}
	\hyphenpenalty=100000
	\footnotesize
	\centering
	\caption*{\textbf{Table~\ref{tab:summary}:} \tablecaption }
	\resizebox{0.9\columnwidth}{!}{%
		\begin{tabular}{l|b{1.7cm}|b{1.7cm}|a{1.7cm}|b{1.7cm}|b{1.7cm}|b{1.9cm}|}
			\hhline{~|-|-|-|-|-|-|}
			&  \textbf{Chapter~\ref{ch:3-risk-mdps}} &  \textbf{Chapter~\ref{ch:4-regret-mdps}}  & \textbf{Chapter~\ref{ch:5-risk-bamdps}} & \textbf{Chapter~\ref{ch:6-rambo}} & \textbf{Chapter~\ref{ch:7-1r2r}} & \textbf{Chapter~\ref{ch:8-lfd}} \\ \hline
			\multicolumn{1}{|m{1.9cm}|}{\textbf{Prior knowledge}} & MDP fully known & MDP uncertainty set & Prior over MDPs & Fixed dataset & Fixed dataset & Expert demonstrator \\ \hline
			\multicolumn{1}{|m{1.9cm}|}{\textbf{Considers aleatoric uncertainty}} & Yes & No & Yes &  No & Yes & No \\ \hline
			\multicolumn{1}{|m{1.9cm}|}{\textbf{Considers epistemic uncertainty}} & No & Yes & Yes & Yes & Yes & Yes \\ \hline
			\multicolumn{1}{|m{1.9cm}|}{\textbf{Risk or robustness}} & Risk &  Robustness  & Risk & Robustness & Risk & N/A \\ \hline
			\multicolumn{1}{|m{1.9cm}|}{\textbf{Paradigm}} & Tabular &  Tabular  & Tabular & Function Approx. & Function Approx. & Function Approx. \\ \hline
		\end{tabular}
	}%
\end{table}

In the previous chapter, we assumed that the prior knowledge about the environment was specified by a set of candidate MDPs, and that this knowledge did not improve after interacting with the environment during execution.
In this chapter, we now assume that the knowledge about the environment can be improved during the course of each episode.
Additionally, the previous chapter disregarded aleatoric uncertainty, as it considered the expected value in each plausible MDP.
In this chapter, we now propose to address both aleatoric and epistemic uncertainty under a single framework by optimising a risk-averse objective.
These characteristics are summarised in Table~\ref{tab:summary}.

Our approach builds upon the Bayesian formulation of model-based reinforcement learning, where we model a  belief distribution over the unknown MDP.
We consider the Bayes-Adaptive MDP (BAMDP) which augments the state space of the original MDP with the belief over the MDP.
In this work, we propose to optimise a risk-sensitive objective, specifically CVaR in the BAMDP.

Our motivation for this approach is that it mitigates both epistemic and aleatoric uncertainty under a single framework, as indicated in Table~\ref{tab:summary}.
The belief distribution over MDPs captures the epistemic uncertainty about the MDP dynamics.
The dynamics of the BAMDP are influenced by both the belief distribution and the aleatoric uncertainty within each plausible MDP.
Therefore, optimising for risk in the BAMDP must avoid poor outcomes that are possible due to either source of uncertainty.
To make this point more concretely, we show that CVaR in a BAMDP is equivalent to the expected value under an adversarial perturbation to both the prior distribution over MDPs, and the transition probabilities within each plausible MDP.

We extend an existing approach to CVaR optimisation~\cite{chow2015risk}.
We model the problem as a turn-based two-player zero-sum game, which is played over the fully-observable BAMDP.
In the game, the adversary modifies the transition probabilities in the BAMDP to reduce the expected return of the agent.
This makes it more likely that the agent experiences transitions in the BAMDP that either a) receive a low rewards, or b) result in the agent transitioning to a belief state where the worst possible MDPs are more likely. 
This results in risk-averse behaviour from the agent, as the agent acts optimally under the assumption that bad transitions, or bad MDPs are more likely than they are under the prior belief.

We propose an algorithm based on two-player Monte Carlo tree search to approximately solve this game. 
Due to the continuous action space of the adversary, this problem is computationally challenging.
To improve scalability we use progressive widening with Bayesian optimisation~\cite{mern2020bayesian} to gradually expand the action space of the adversary.
We demonstrate that this approach outperforms existing baselines on two small domains.

\noindent\rule{6cm}{0.5pt} 

\medskip
\noindent Marc Rigter, Bruno Lacerda and Nick Hawes (2021). Risk-averse Bayes-adaptive reinforcement learning. \textit{Advances in Neural Information Processing Systems} (NeurIPS).
\medskip

\includepdf[pages=-]{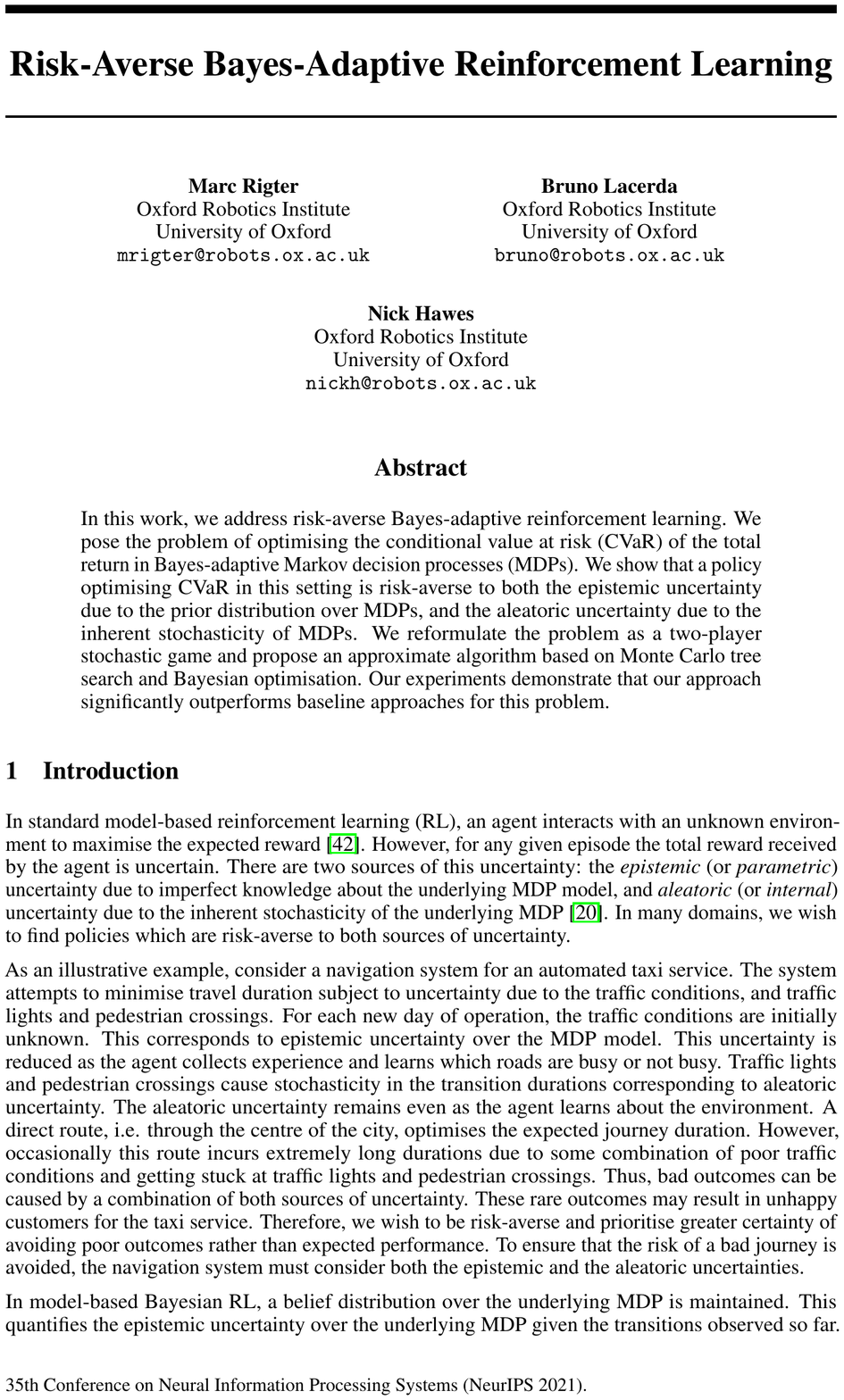}

\section{Limitations and Future Work}
There are two key limitations to this work.
The first is that it is necessary to specify a suitable prior over the MDP.
If a completely uninformative prior is used, this is unlikely to be very meaningful when it comes to optimising risk.
It is also difficult to know how to represent the prior such that the posterior can be computed in a scalable way.
These issues were not the focus of this piece of research, so we used simple Dirichlet priors for the toy problems we tested on.
A more appropriate approach in future work may instead be to use approaches that learn to represent the prior and posterior update over MDPs after repeated interactions with samples of MDPs~\cite{doshi2016hidden, zintgraf2020varibad}, rather than specifying this a priori.

The second major limitation of this work is that the two-player Monte Carlo tree search algorithm that we proposed is unlikely to scale to larger problems.
Indeed, we found that scaling our method beyond the toy problems presented in the paper was difficult.
This is because the adversary has a large and continuous action space for perturbing the transition probabilities.
Furthermore, because we optimise the static risk objective, the adversary's optimal strategy depends on it's ``budget'' for perturbing the transition probabilities over the entire trajectory.
These two factors make it is very difficult to find the correct policy for the adversary using our approach.
If the adversary is suboptimal, then the agent policy does not correctly optimise CVaR in response.

There are two possible approaches to alleviating these scalability issues.
The first would be to consider the dynamic perspective of risk, rather than static risk.
For dynamic risk, we would not have to keep track of the ``budget'' of the adversary to perturb the transition probabilities. 
This is because in the dynamic perspective of risk, the risk measure is applied recursively at each step.
This means that we can separately consider an adversarially modified transition distribution at each step, rather than over the entire trajectory.
The downside of the dynamic risk approach is that it would be very difficult to interpret what the resulting algorithm achieves, especially in the BAMDP setting. 

Another avenue to improving scalability would be to utilise deep RL algorithms for optimising CVaR based on either policy gradients~\cite{tamar2015policy} or distributional RL~\cite{dabney2018implicit}.
These algorithms could be combined with methods for BAMDPs where the posterior update is learned via sampling~\cite{zintgraf2020varibad}.
This approach would stay true to our original aim of jointly mitigating risk due to both the uncertainty over the unknown MDP, and the inherent randomness of MDPs, but would leverage scalable RL methods achieve this goal.


%% file: text/ch6-rambo.tex

\chapter{\label{ch:6-rambo} RAMBO-RL: Robust Adversarial Model-Based Offline Reinforcement Learning}
\chaptermark{Robust Adversarial Model-Based Offline Reinforcement Learning}

\begin{table}[b!]
	\renewcommand{\arraystretch}{1.3}
	\hyphenpenalty=100000
	\footnotesize
	\caption*{\textbf{Table~\ref{tab:summary}:} \tablecaption }
	\resizebox{\columnwidth}{!}{%
		\begin{tabular}{l|b{1.7cm}|b{1.7cm}|b{1.7cm}|a{1.7cm}|b{1.7cm}|b{1.9cm}|}
			\hhline{~|-|-|-|-|-|-|}
			&  \textbf{Chapter~\ref{ch:3-risk-mdps}} &  \textbf{Chapter~\ref{ch:4-regret-mdps}}  & \textbf{Chapter~\ref{ch:5-risk-bamdps}} & \textbf{Chapter~\ref{ch:6-rambo}} & \textbf{Chapter~\ref{ch:7-1r2r}} & \textbf{Chapter~\ref{ch:8-lfd}} \\ \hline
			\multicolumn{1}{|m{1.9cm}|}{\textbf{Prior knowledge}} & MDP fully known & MDP uncertainty set & Prior over MDPs & Fixed dataset & Fixed dataset & Expert demonstrator \\ \hline
			\multicolumn{1}{|m{1.9cm}|}{\textbf{Considers aleatoric uncertainty}} & Yes & No & Yes &  No & Yes & No \\ \hline
			\multicolumn{1}{|m{1.9cm}|}{\textbf{Considers epistemic uncertainty}} & No & Yes & Yes & Yes & Yes & Yes \\ \hline
			\multicolumn{1}{|m{1.9cm}|}{\textbf{Risk or robustness}} & Risk &  Robustness  & Risk & Robustness & Risk & N/A \\ \hline
			\multicolumn{1}{|m{1.9cm}|}{\textbf{Paradigm}} & Tabular &  Tabular  & Tabular & Function Approx. & Function Approx. & Function Approx. \\ \hline
		\end{tabular}
	}%
\end{table}

The two  previous pieces of work assumed that prior information about the MDP was specified either in the form of a set of plausible MDPs, or a suitable prior over MDPs.
In the following two chapters, we address the offline reinforcement learning setting, where the uncertainty over the MDP is specified by a fixed dataset of experience in the environment.
This source of knowledge about the environment is indicated in Table~\ref{tab:summary}.
This is arguably a more pragmatic problem formulation, as most applications of sequential decision-making algorithms in the real world will need to be based on data collected from the environment.

Offline RL avoids the need for the extensive online exploration required by standard online RL methods by leveraging pre-existing datasets.
We consider model-based approaches, which learn a model of the environment dynamics using the dataset, and use that model to generate synthetic data to train a policy.

As discussed in Section~\ref{sec:offline_rl}, the issue with directly applying this approach is that the policy may learn to choose actions that exploit errors in the dynamics model.
Instead, we would like to ensure that the policy takes actions that are likely to perform well in the real environment, as evidenced by the dataset.
The contribution of this work is to propose a new approach to address the issue of model exploitation without the need for explicit uncertainty estimation.

We address an existing formulation of model-based offline RL that poses the problem as solving a type of robust MDP, where the goal is the find the policy with the best expected value in the worst-case instantiation of the MDP.
The uncertainty set for the robust MDP is defined such that it includes all models that have a low prediction error with respect to the dataset.
This problem formulation has been analysed theoretically in prior work~\cite{uehara2022pessimistic}.
As indicated by Table~\ref{tab:summary}, this chapter therefore proposes a robust approach to addressing epistemic uncertainty.
Because the expected value is optimised within the worst-case MDP, this approach does not take into consideration aleatoric uncertainty.

For the remainder of this thesis, we no longer focus on tabular MDPs. 
Instead, we develop methods that employ deep RL to utilise powerful neural network function approximators and achieve scalability to problems with large or continuous state spaces.
The use of function approximation in the remaining chapters of this thesis is indicated by the ``Paradigm'' row of Table~\ref{tab:summary}.

To arrive at a scalable solution to the robust MDP formulation that we address, we propose an approach inspired by Robust Adversarial RL~\cite{pinto2017robust}.
We alternate between optimising the policy to increase the expected reward in the model, and optimising the model to decrease it.
However, the model is constrained such that it must still accurately predict the transitions within the dataset.
Training the model in this way means that it predicts accurate transitions for state-action pairs covered by the dataset, but generates pessimistic synthetic  transitions for areas far from the dataset.
This prevents the policy from exploiting errors in the model, because in regions where there is little or no data, the model generates pessimistic transitions.

We evaluate our approach on standard benchmarks for offline RL, and compare our algorithm to a number of existing methods. 
The results demonstrate that our approach achieves state-of-the-art performance on these benchmarks.
We also analyse the characteristics of our approach on a toy example.
We find that initially, the value function is optimistic and pessimism is introduced gradually as the model is trained adversarially.
Our results suggest that this initial optimism makes it less likely that the policy becomes stuck in a poor local minimum, compared to the most similar existing approach.

\noindent\rule{6cm}{0.5pt} 

\medskip
\noindent Marc Rigter, Bruno Lacerda and Nick Hawes (2022). RAMBO-RL: Robust adversarial model-based offline reinforcement learning. \textit{Advances in Neural Information Processing Systems} (NeurIPS).
\medskip

\includepdf[pages=-]{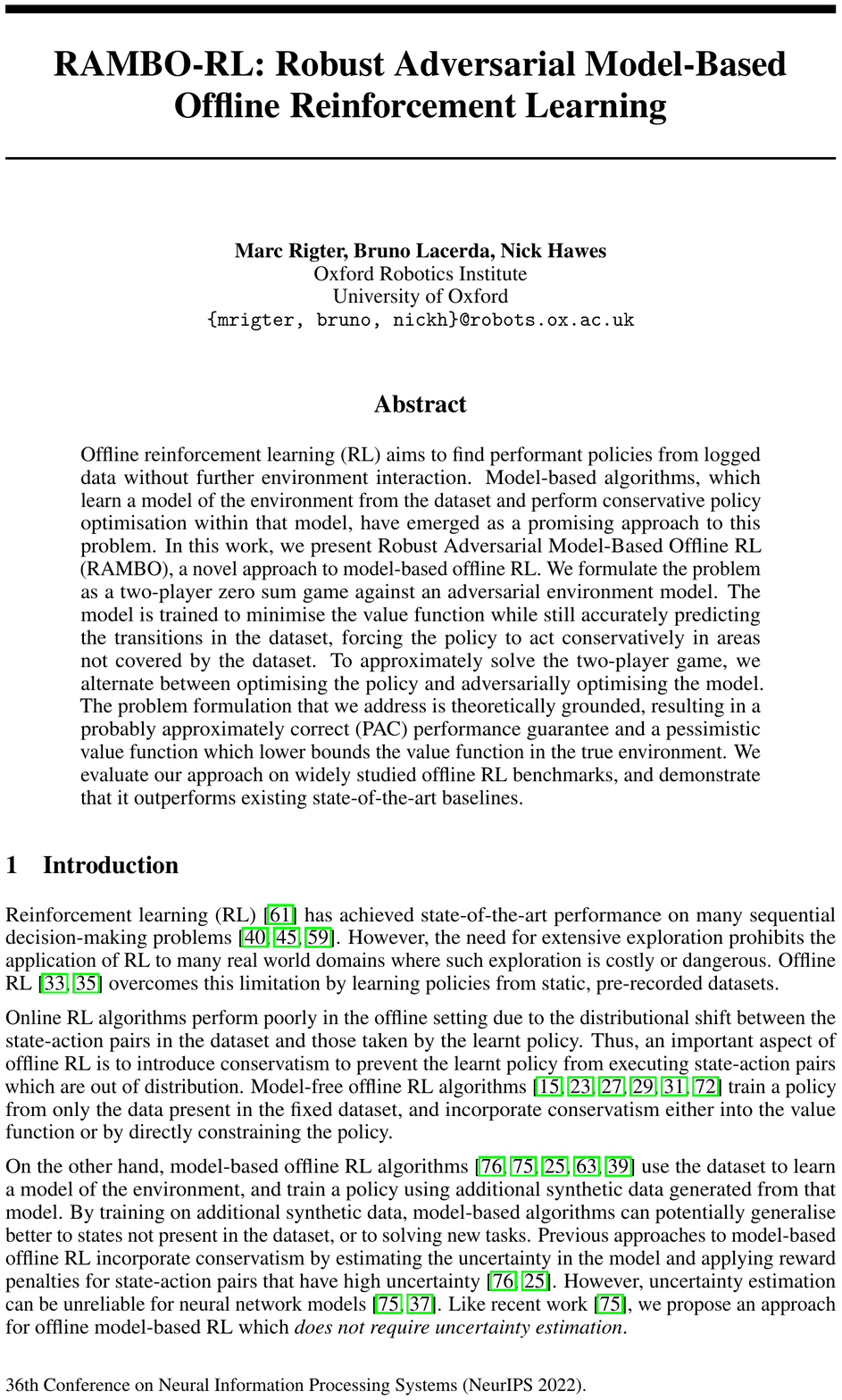}

\section{Additional Results}
As with the other chapters, the entirety of the appendices for this chapter are included at the end of this document.
However, here we include results from Appendix C.1 and C.3 because they give additional insight into our approach.
Appendix C.1 highlights differences in our approach compared to COMBO~\cite{yu2021combo}, the most similar prior method.
Appendix C.3 provides a visualisation of how our approach adversarially modifies the dynamics model, and how this regularises the Q-values for out-of-distribution actions.

\vspace{40pt}
\noindent \textbf{\large Single Transition Example}
\paragraph{Description} In this domain, the state and action spaces are one-dimensional. The agent executes a single action, $a \in [-1, 1]$, from initial state $s_0$. After executing the action, the agent transitions to $s'$ and receives a reward equal to the successor state, i.e. $r(s') = s'$, and the episode terminates.

The actions in the dataset are sampled uniformly from $a \in [-0.75, 0.7]\ \cup\ [-0.15, -0.1]\ \cup\ [0.1, 0.15]\ \cup\ [0.7, 0.75]$. In the MDP for this domain, the transition distribution for successor states is as follows:
\begin{itemize}
	\item $s' \sim \mathcal{N}(\mu =1,\ \sigma = 0.2)$, for $a \in [-0.8,\ -0.65]$.
	\item $s' \sim \mathcal{N}(\mu =0.5,\ \sigma = 0.2)$, for $a \in [-0.2,\ -0.05]$.
	\item $s' \sim \mathcal{N}(\mu =1.25,\ \sigma = 0.2)$, for $a \in [0.05,\ 0.2]$.
	\item $s' \sim \mathcal{N}(\mu =1.5,\ \sigma = 0.2)$, for $a \in [0.65,\ 0.8]$.
	\item $s' = 0.5$, for all other actions.
\end{itemize}

\begin{figure}[htb]
	\centering
	\includegraphics[width=6cm]{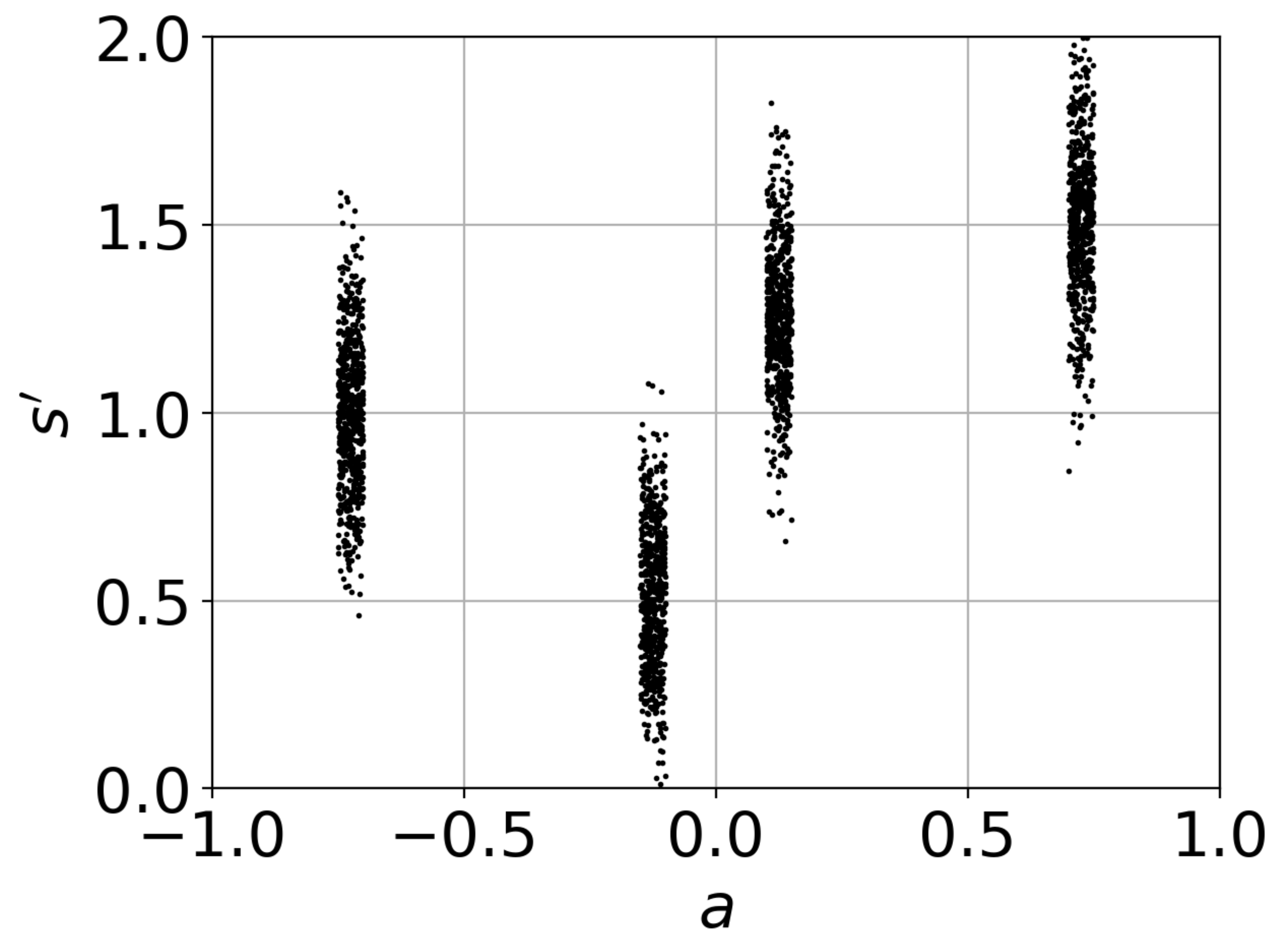}
	\caption{Transition data for the Single Transition Example after executing action $a$ from $s_0$. \label{fig:ste_example}}
\end{figure}

The transitions to successor states from the actions in the dataset are illustrated in Figure~\ref{fig:ste_example}.

\paragraph{Comparison between RAMBO and COMBO} To generate the comparison, we run both algorithms for 50 iterations. To choose the regularisation parameter for COMBO, we sweep over $\beta \in \{0.1, 0.2, 0.5, 1, 5\}$ and choose the parameter with the best performance. Note that 0.5, 1, and 5 are the values used in the original COMBO paper~\cite{yu2021combo}, so we consider lower values of regularisation than in the original paper. For RAMBO we use $\lambda=$3e-2. We used the implementation of COMBO from~\href{https://github.com/takuseno/d3rlpy}{github.com/takuseno/d3rlpy}.

\begin{figure}[htb]
	\centering
	\begin{subfigure}[t]{.48\textwidth}
		\centering
		\includegraphics[width=6cm]{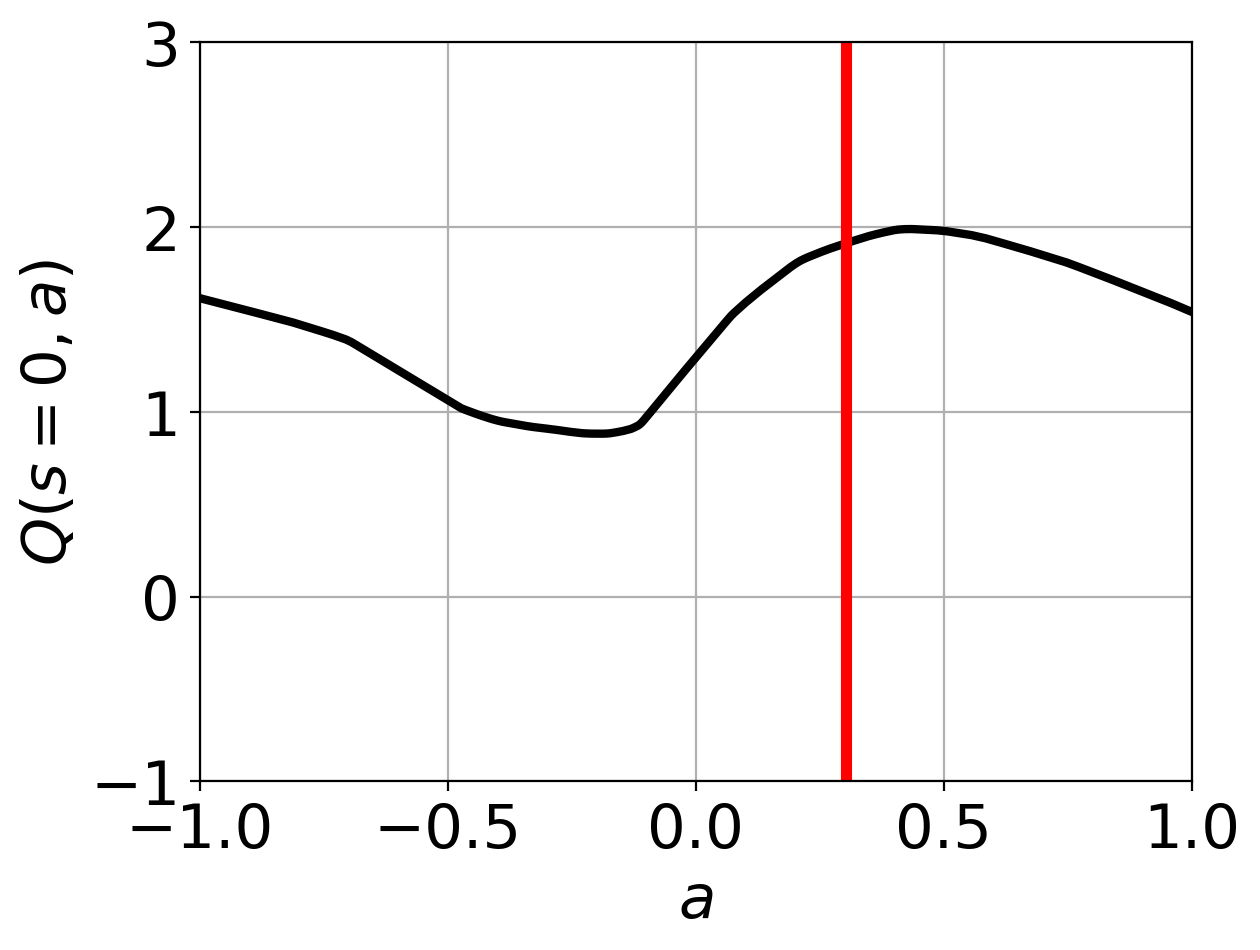}
		\caption{5 iterations}
		\label{fig:rambo_5}
	\end{subfigure}%
	\hfill
	\begin{subfigure}[t]{.48\textwidth}
		\centering
		\includegraphics[width=6cm]{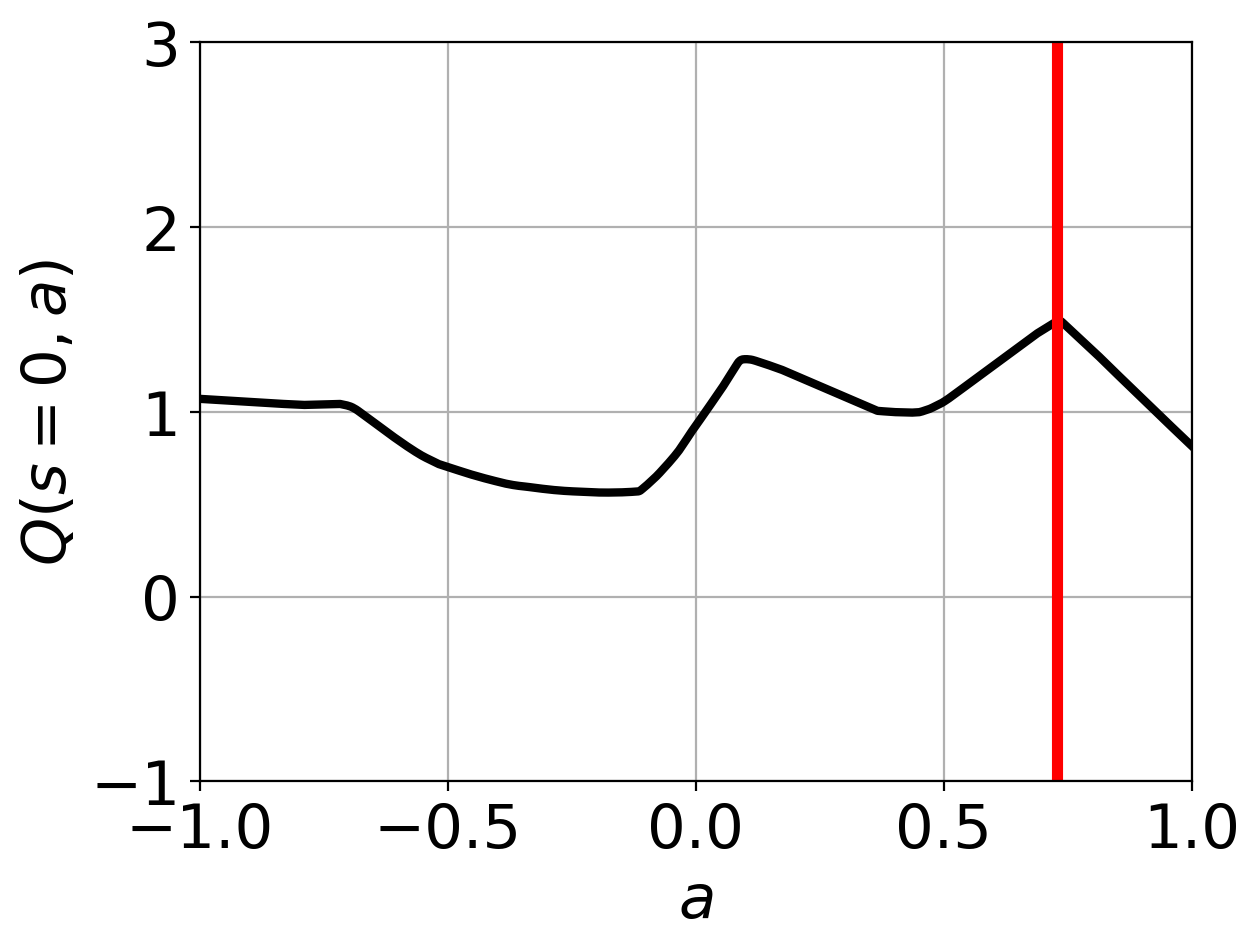}
		\caption{50 iterations}
		\label{fig:rambo_50}
	\end{subfigure}
	\caption{$Q$-values at the initial state during the training of RAMBO on the Single Transition Example. The red line indicates the mean action of the policy. }
	\label{fig:ste_rambo}
\end{figure}

Figure~\ref{fig:ste_rambo} shows the $Q$-values produced by RAMBO throughout training. We can see that after 5 iterations, the $Q$-values for actions which are outside of the dataset are initially~\textit{optimistic}, and over-estimate the true values. However, after 50 iterations the $Q$-values for out of distribution actions have been regularised by RAMBO. The action selected by the policy after 50 iterations is the optimal action in the dataset, $a \in [0.7, 0.75]$. Thus, we see that RAMBO introduces pessimism~\textit{gradually} as the adversary is trained to modify the transitions to successor states.

\begin{figure}[htb]
	\centering
	\begin{subfigure}[t]{.48\textwidth}
		\centering
		\includegraphics[width=6cm]{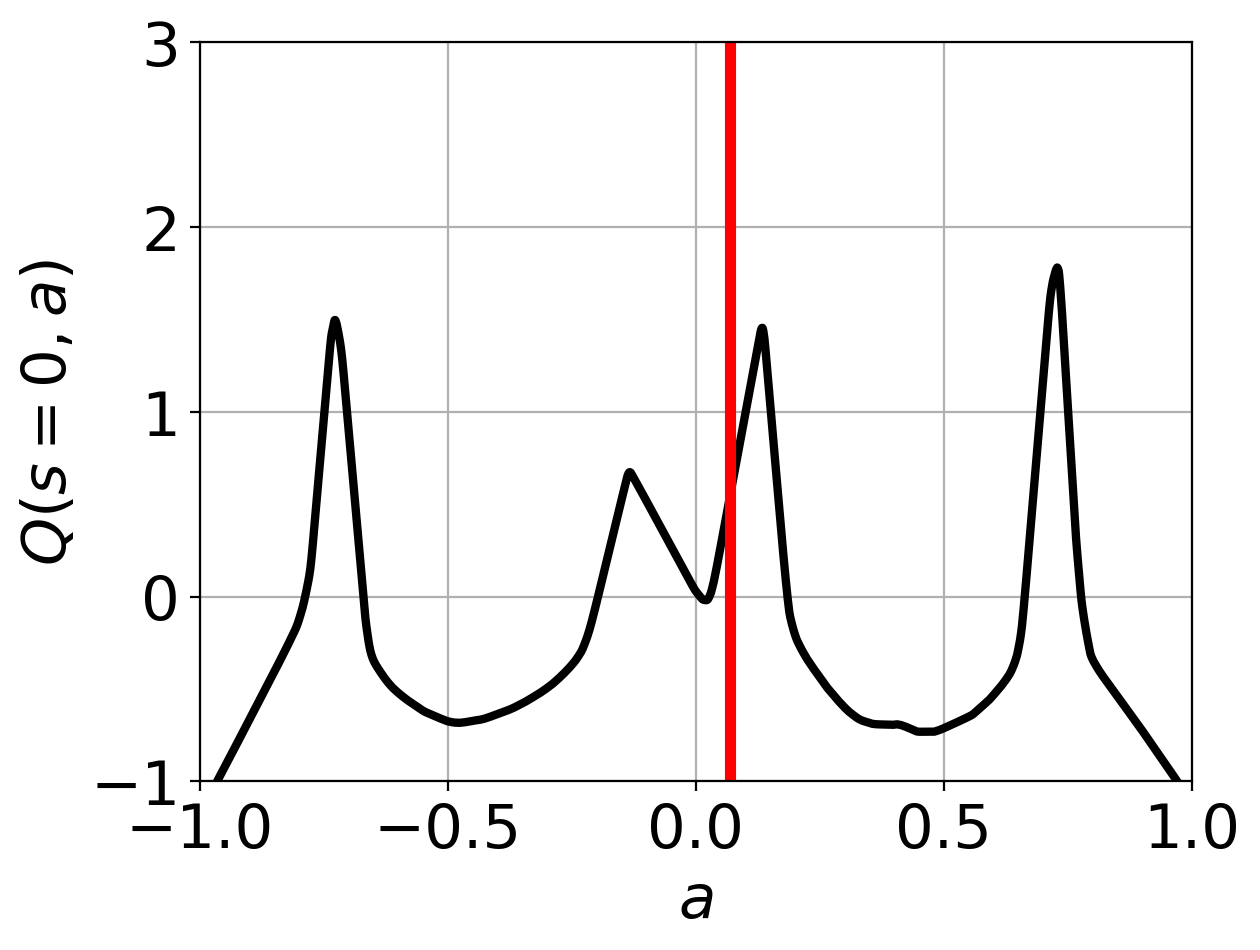}
		\caption{5 iterations}
		\label{fig:combo_5}
	\end{subfigure}%
	\hfill
	\begin{subfigure}[t]{.48\textwidth}
		\centering
		\includegraphics[width=6cm]{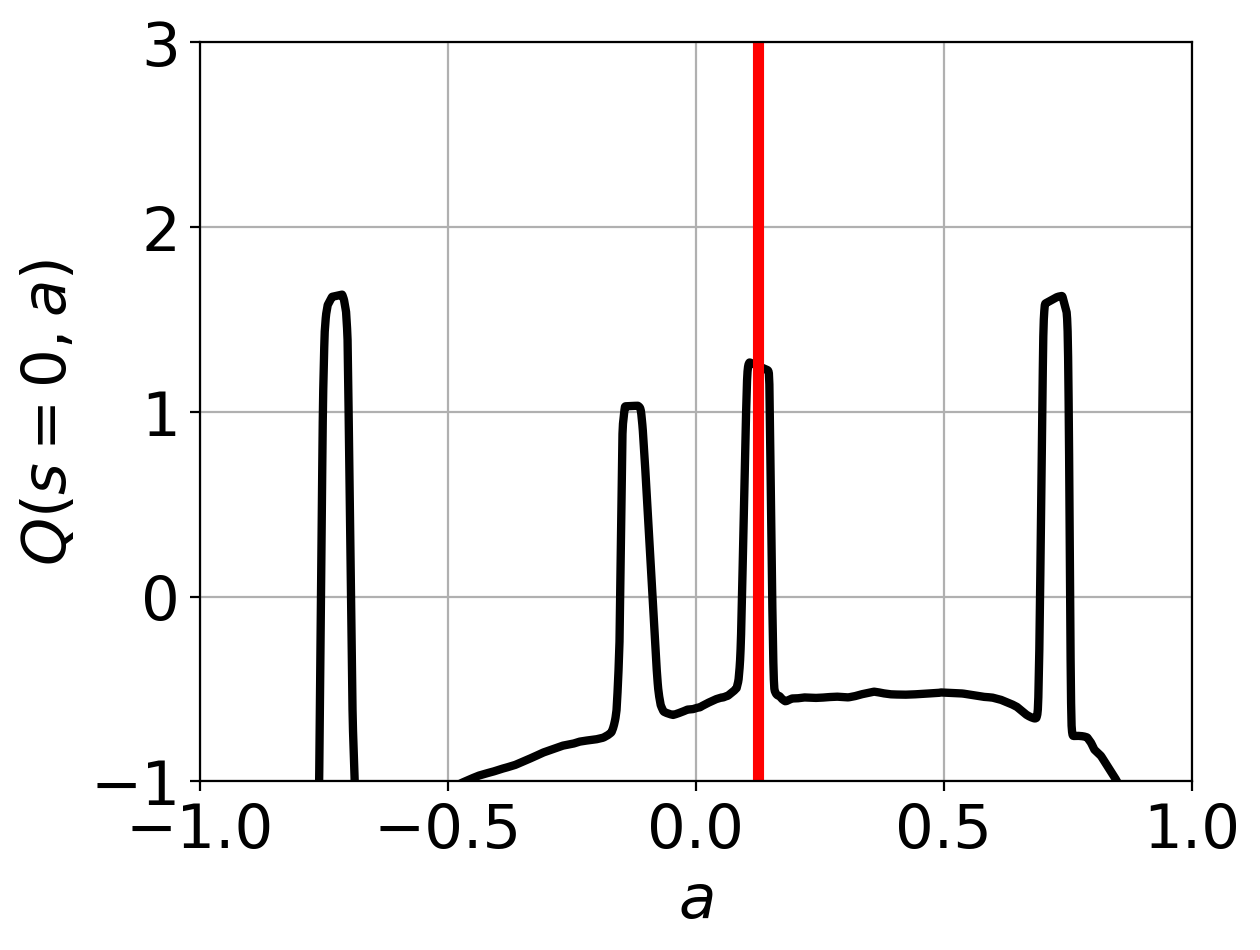}
		\caption{50 iterations}
		\label{fig:combo_50}
	\end{subfigure}
	\caption{$Q$-values at the initial state during the training of COMBO on the Single Transition Example ($\beta = 0.2$). The red line indicates the mean action of the policy.}
	\label{fig:ste_combo}
\end{figure}

For COMBO, the best performance averaged over 20 seeds was obtained for $\beta = 0.2$, and therefore we report results for this value of the regularisation parameter. Figure~\ref{fig:ste_combo} illustrates the $Q$-values produced by COMBO throughout training (with $\beta = 0.2$). We see that at both 5 iterations and 50 iterations, the value estimates for actions outside of the dataset are highly pessimistic. In the run illustrated for COMBO, the action selected by the policy after 50 iterations is $a \in [0.1, 0.15]$, which is not the optimal action in the dataset. Figure~\ref{fig:ste_combo} shows that the failure to find an optimal action is due to the gradient-based policy optimisation becoming stuck in a local maxima of the $Q$-function. 

These results highlight a difference in the behaviour of RAMBO compared to COMBO. For RAMBO, pessimism is introduced gradually as the adversary is trained to modify the transitions to successor states. For COMBO, pessimism is introduced at the outset as it is part of the value function update throughout the entirety of training. Additionally, the regularisation of the $Q$-values for out of distribution actions appears to be less aggressive for RAMBO than for COMBO.

The results averaged over 20 seeds in Table 3 of the paper show that RAMBO consistently performs better than COMBO for this problem.  This suggests that the gradual introduction of pessimism produced by RAMBO means that the policy optimisation procedure is less likely to get stuck in poor local maxima for this example. The downside of this behaviour is that it may take more iterations for RAMBO to find a performant policy. 
Modifying other algorithms to gradually introduce pessimism may be an interesting direction for future research.

\FloatBarrier

\vspace{40pt}
\noindent \textbf{\large Visualisation of Adversarially Trained Dynamics Model}

To visualise the influence of training the transition dynamics model in an adversarial manner as proposed by RAMBO, we consider the following simple example MDP. In the example, the state space ($S$) and action space ($A$) are 1-dimensional with, $A = [-1, 1]$. For this example, we assume that the reward function is known and is equal to the current state, i.e. $R(s, a) = s$, meaning that greater values of $s$ have greater expected value. The true transition dynamics to the next state $s'$ depend on the action but are the same regardless of the initial state, $s$.

\begin{figure}[htb]
	\centering
	\includegraphics[width=5.5cm]{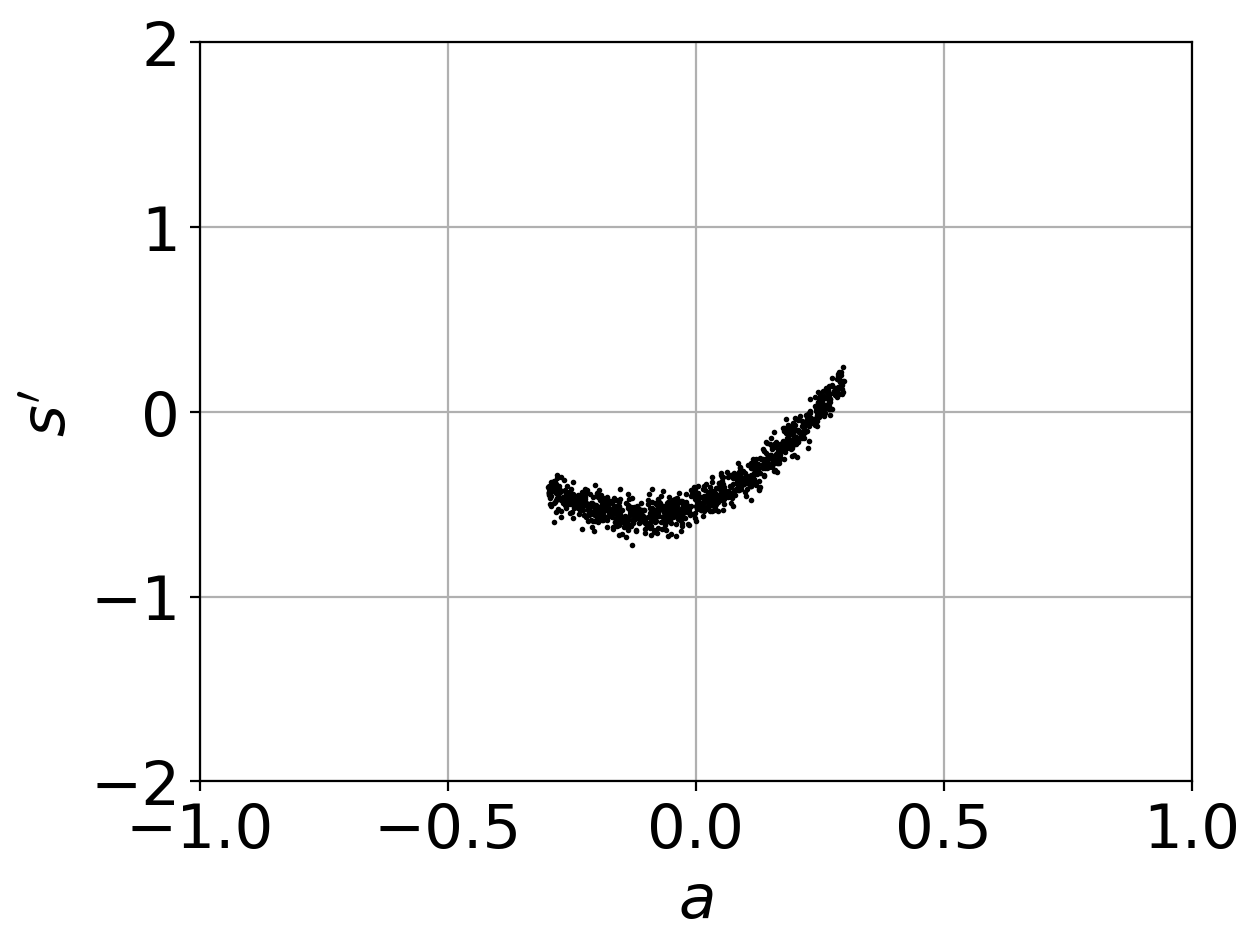}
	\caption{Transition data for illustrative example. The transition function is the same regardless of the initial state. The actions in the dataset are sampled from $a \in [-0.3, 0.3]$. \label{fig:example_data}}
\end{figure}

In Figure~\ref{fig:example_data} we plot the data present in the offline dataset for this MDP. Note that the actions in the dataset are sampled from a subset of the action space: $a \in [-0.3, 0.3]$.
Because greater values of $s'$ correspond to greater expected value, the transition data indicates that the optimal action covered by the dataset is $a = 0.3$.

\begin{figure}[htb]
	\centering
	\begin{subfigure}[t]{.48\textwidth}
		\centering
		\includegraphics[width=5.5cm]{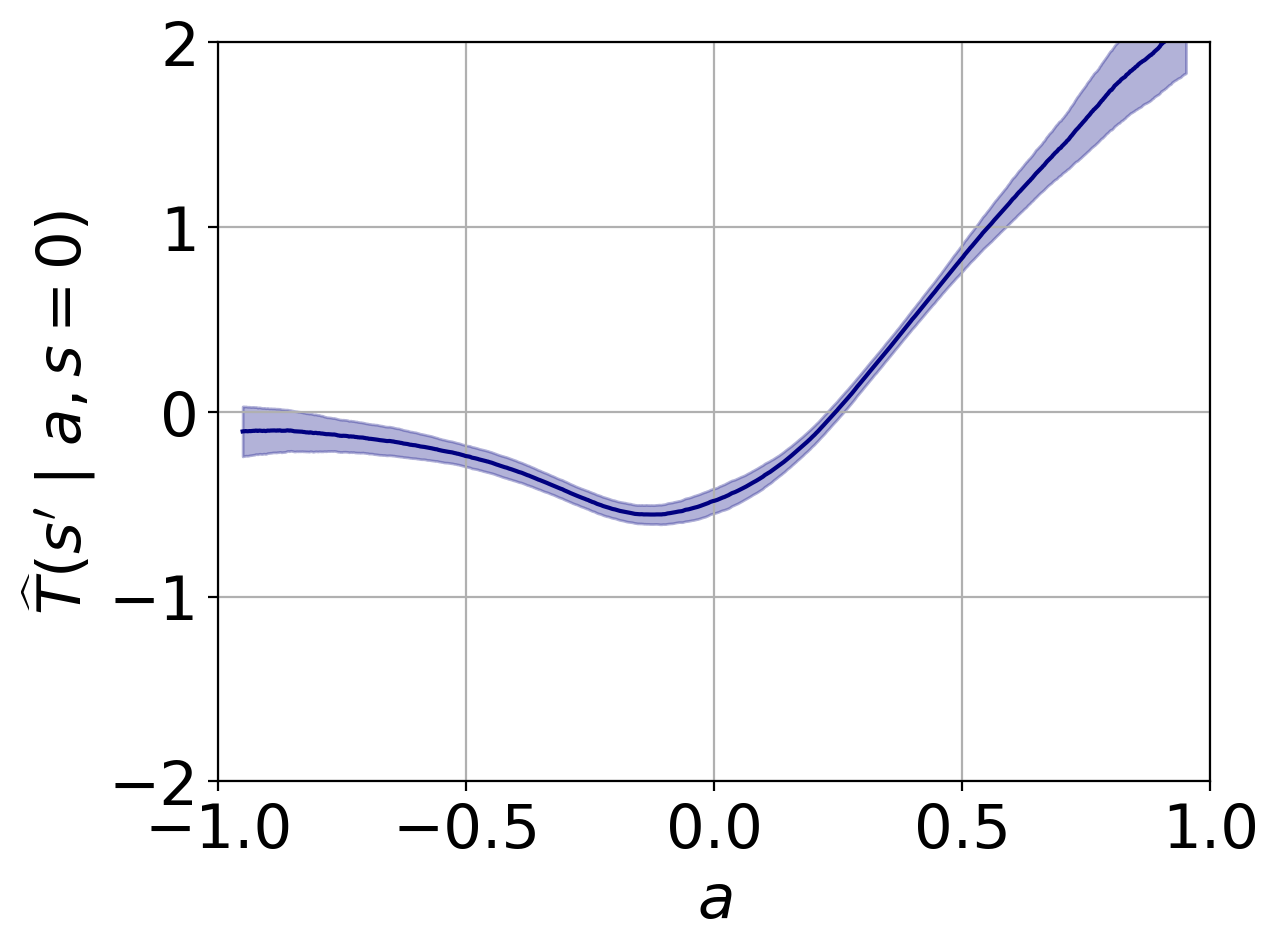}
		\caption{Maximum likelihood estimate of the transition function, which is used by MBPO. Shaded area indicates $\pm$1 SD of samples generated by the ensemble.}
		\label{fig:mle_trans}
	\end{subfigure}%
	\hfill
	\begin{subfigure}[t]{.48\textwidth}
		\centering
		\includegraphics[width=5.5cm]{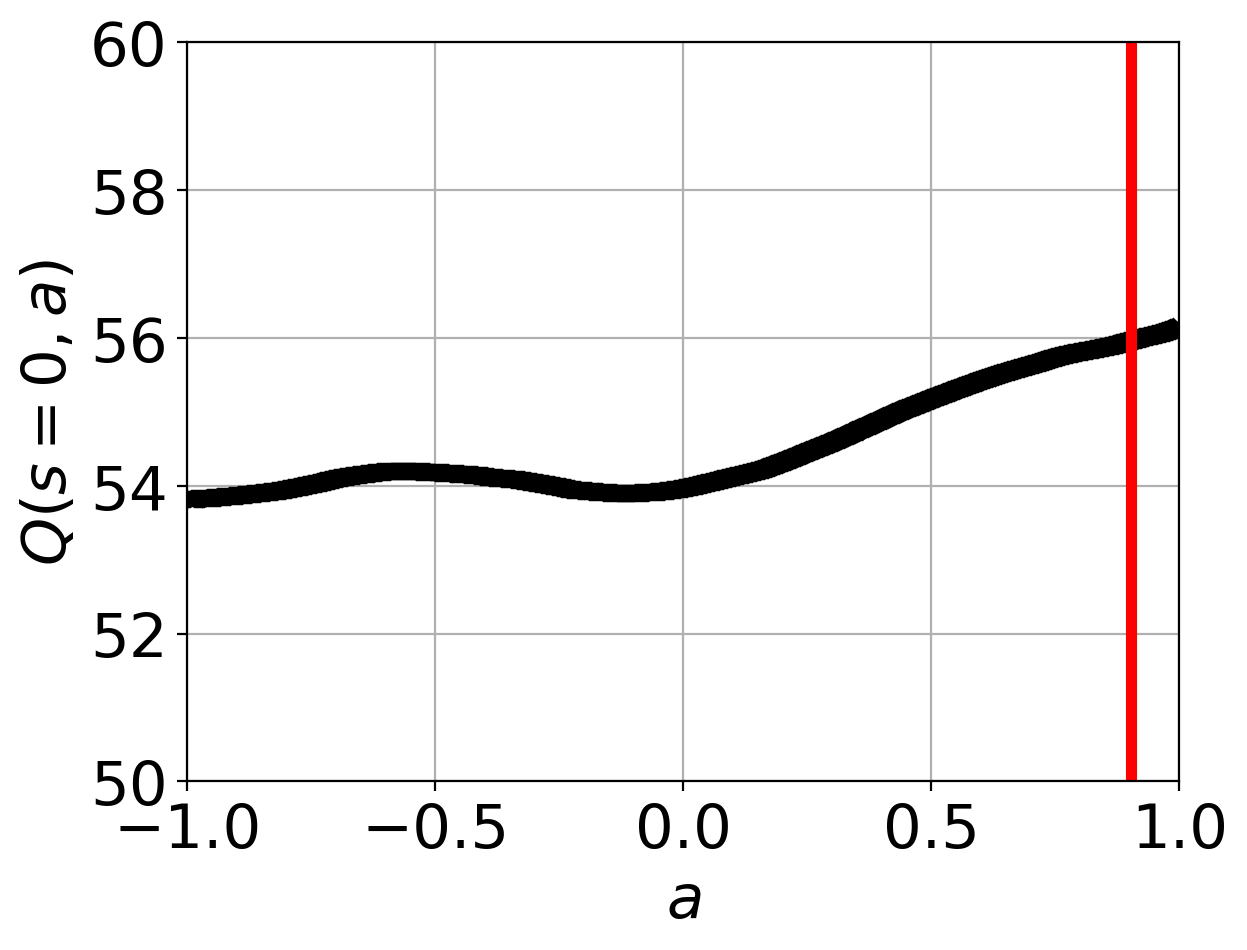}
		\caption{$Q$-values after running SAC for 15 iterations. The values are significantly over-estimated. The red line indicates the mean action taken by the SAC policy.}
		\label{fig:sac_mbpo}
	\end{subfigure}
	\caption{Plots generated by running na\"ive MBPO on the illustrative MDP example. }
	\label{fig:mbpo_output}
\end{figure}

In Figure~\ref{fig:mbpo_output} we plot the output of running na\"ive model-based policy optimisation (MBPO) on this illustrative offline RL example.
Figure~\ref{fig:mle_trans} illustrates the MLE transition function used by MBPO. 
The transition function fits the dataset for $a \in [-0.3, 0.3]$. Outside of the dataset, it predicts that an action of $a \approx 1$ transitions to the best successor state.
Figure~\ref{fig:sac_mbpo} shows that applying a reinforcement learning algorithm (SAC) to this model results in the value function being over-estimated, and $a \approx 1$ being predicted as the best action.
This illustrates that the policy learns to exploit the inaccuracy in the model and choose an out-of-distribution action.

\begin{figure}[htb]
	\centering
	\begin{subfigure}[t]{.48\textwidth}
		\centering
		\includegraphics[width=5.5cm]{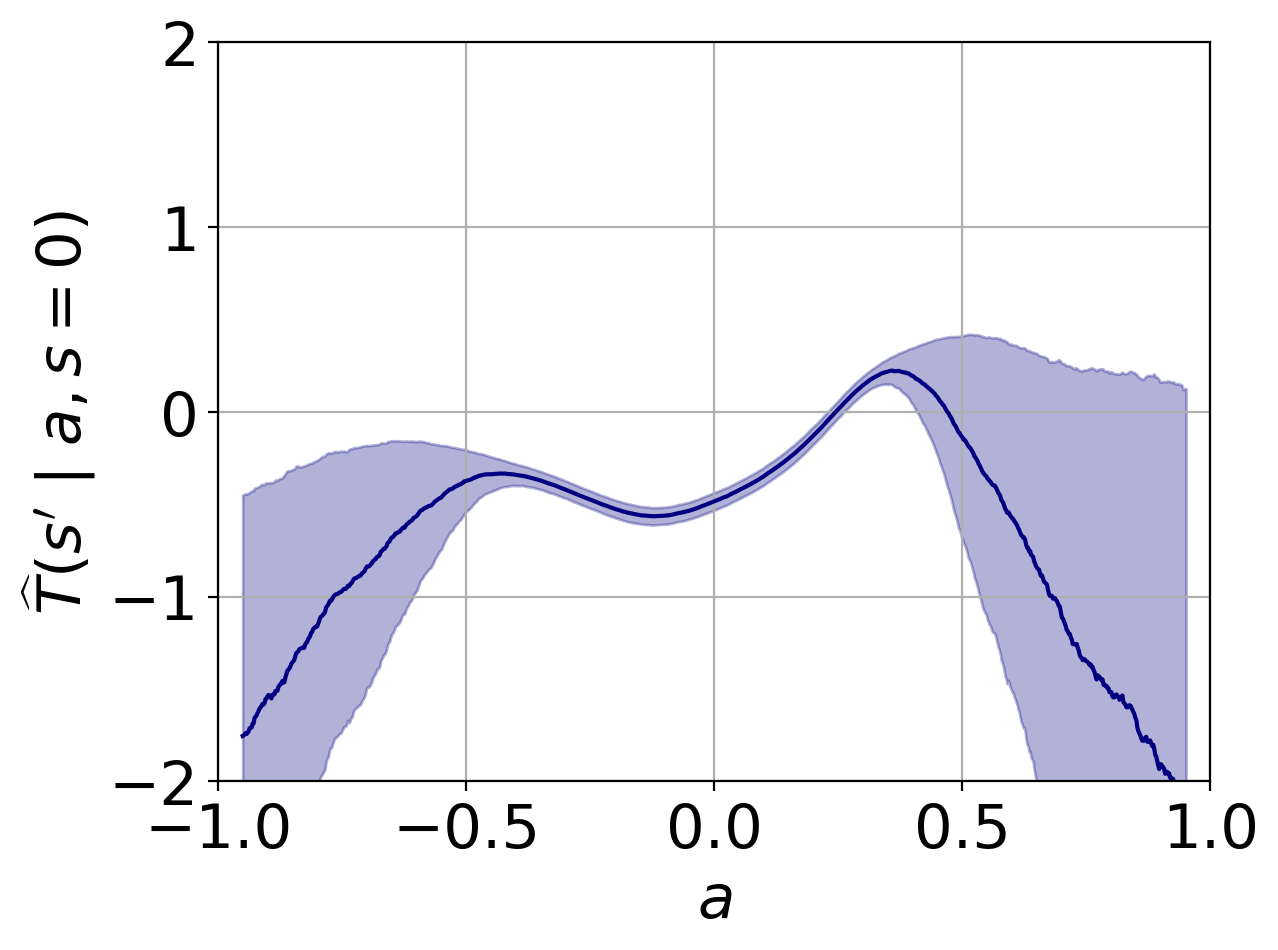}
		\caption{Model which is adversarially trained using RAMBO. Shaded area indicates $\pm$1 SD of samples generated by the ensemble. The transition function accurately predicts the dataset for in-distribution actions, but predicts transitions to low value states for actions which are out of distribution. }
		\label{fig:trans_rambo}
		\vfill
	\end{subfigure}%
	\hfill
	\begin{subfigure}[t]{.48\textwidth}
		\centering
		\includegraphics[width=5.5cm]{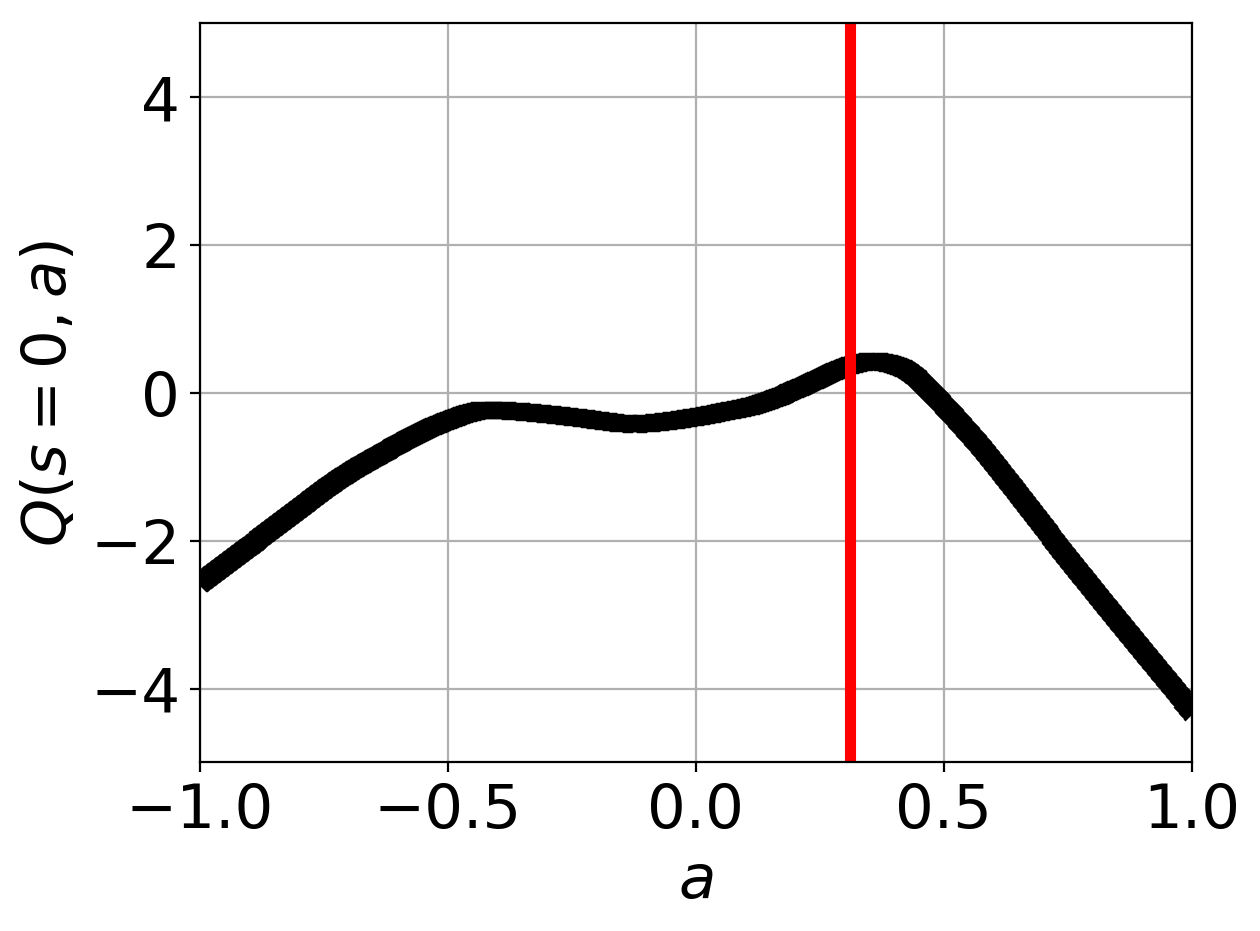}
		\caption{$Q$-values from SAC critic. The red line indicates the mean action taken by the policy trained using SAC, which is $a \approx 0.3$ (the best in-distribution action). Training on pessimistic synthetic transitions regularises the $Q$-values for out-of-distribution actions.}
		\label{fig:rambo_sac}
		\vfill
	\end{subfigure}
	\caption{Output generated by running RAMBO for 15 iterations on the illustrative MDP example, using an adversary weighting of $\lambda = $3e-2.}
	\label{fig:rambo_output}
\end{figure}

\FloatBarrier

In Figure~\ref{fig:rambo_output} we show plots generated after running RAMBO on this example for 15 iterations, using an adversary weighting of $\lambda =\ $3e-2.
Figure~\ref{fig:trans_rambo} shows that the transition function still accurately predicts the transitions within the dataset. 
This is because choosing $\lambda \ll 1 $ means that the MLE loss dominates for state-action pairs within the dataset.
Due to the adversarial training, for actions $a \notin [-0.3, 0.3]$ which are out of the dataset distribution, the transition function generates pessimistic transitions to low values of $s$, which have low expected value.

As a result, low values are predicted for out-of-distribution actions, and the SAC agent illustrated in~\ref{fig:rambo_sac} learns to take action $a \approx 0.3$, which is the best action which is within the distribution of the dataset.
This demonstrates that by generating pessimistic synthetic transitions for out-of-distribution actions, RAMBO enforces conservatism and prevents the learnt policy from taking state-action pairs which have not been observed in the dataset.

\section{Limitations and Future Work}
One of the limitations of RAMBO is that we found that it did not perform well on the harder AntMaze benchmarks, along with all the other model-based methods that we compared against.
Understanding why model-based methods perform less well on these more difficult domains is an important question for future work.
It is unclear if this is simply because the learnt model is not accurate enough, or there is some deeper limitation.

Recent work~\cite{wang2021offline} suggests that model-based methods are overly aggressive and leave the dataset distribution, resulting in the robot contacting the walls of the maze.
One would expect that the robust MDP formulation that we address should prevent this from happening.
However, the state-action space of AntMaze domain has a greater number of dimensions  than the other benchmarks (>50 vs $\sim$20).
Adversarially training the model may be slow for high-dimension problems, as there is a large space of state-action pairs for which the model needs to be modified to generate pessimistic transitions.
Because of this, it might be possible that RAMBO fails to ensure that the policy remains within the dataset distribution for these problems.
A first step to investigate this as a possible issue would be to visualise the trajectories generated by the policy, to identify whether the policy may be leaving the data distribution.

Another avenue for future work could be investigating rollouts that are generated backwards through time~\cite{wang2021offline}.
The advantage of a model that generates reverse rollouts is that it can be used to guarantee that that all simulated trajectories end at states contained within the dataset. 
This approach has been shown to improve the performance of model-based methods on the AntMaze benchmarks~\cite{wang2021offline}.
It would be interesting to combine the regularisation approach that we propose in RAMBO with a reverse model to see whether this improves performance.

Another limitation of our approach is that we make use of a simple model that predicts a Gaussian distribution over successor states.
This type of model is suitable for the MuJoCo benchmarks that we tested on which are deterministic.
However, many real-world problems are stochastic, and these stochastic outcomes may have multimodal distributions.
Therefore, combining model-based offline RL algorithms with dynamics models that can more accurately predict stochastic outcomes~\cite{hafner2020mastering} is an important direction for future work.


%% file: text/ch7-1r2r.tex
\chapter{\label{ch:7-1r2r} One Risk to Rule Them All: A Risk-Sensitive Perspective on Model-Based Offline Reinforcement Learning}
\chaptermark{One Risk to Rule Them All}

In the previous chapter we presented RAMBO, which aims to ensure that the learnt policy remains within regions covered by the dataset by generating pessimistic synthetic transitions for out-of-distribution state-action pairs.
RAMBO is based on a robust MDP formulation, which optimises the policy in a worst-case instantiation of plausible MDP models.
We can view this approach as achieving robustness towards the epistemic uncertainty resulting from the limited dataset.
RAMBO optimises for the expected value in the worst-case MDP instantiation, and therefore is neutral towards aleatoric uncertainty.

One of the key advantages of offline RL compared to online RL is that it is more suitable for safety-critical domains where online exploration is too dangerous or costly.
In many safety-critical domains in which offline RL is applicable, we want the policy to be risk-averse towards aleatoric uncertainty, and prioritise avoiding the worst outcomes that might occur due to environment stochasticity. 
However, the vast majority of offline RL research, including RAMBO, considers risk-neutral algorithms.
This motivates the work in this chapter, where we develop a risk-sensitive approach to offline RL.

\begin{table}[tb!]
	\renewcommand{\arraystretch}{1.3}
	\hyphenpenalty=100000
	\footnotesize
	\caption*{\textbf{Table~\ref{tab:summary}:} \tablecaption }
	\resizebox{\columnwidth}{!}{%
		\begin{tabular}{l|b{1.7cm}|b{1.7cm}|b{1.7cm}|b{1.7cm}|a{1.7cm}|b{1.9cm}|}
			\hhline{~|-|-|-|-|-|-|}
			&  \textbf{Chapter~\ref{ch:3-risk-mdps}} &  \textbf{Chapter~\ref{ch:4-regret-mdps}}  & \textbf{Chapter~\ref{ch:5-risk-bamdps}} & \textbf{Chapter~\ref{ch:6-rambo}} & \textbf{Chapter~\ref{ch:7-1r2r}} & \textbf{Chapter~\ref{ch:8-lfd}} \\ \hline
			\multicolumn{1}{|m{1.9cm}|}{\textbf{Prior knowledge}} & MDP fully known & MDP uncertainty set & Prior over MDPs & Fixed dataset & Fixed dataset & Expert demonstrator \\ \hline
			\multicolumn{1}{|m{1.9cm}|}{\textbf{Considers aleatoric uncertainty}} & Yes & No & Yes &  No & Yes & No \\ \hline
			\multicolumn{1}{|m{1.9cm}|}{\textbf{Considers epistemic uncertainty}} & No & Yes & Yes & Yes & Yes & Yes \\ \hline
			\multicolumn{1}{|m{1.9cm}|}{\textbf{Risk or robustness}} & Risk &  Robustness  & Risk & Robustness & Risk & N/A \\ \hline
			\multicolumn{1}{|m{1.9cm}|}{\textbf{Paradigm}} & Tabular &  Tabular  & Tabular & Function Approx. & Function Approx. & Function Approx. \\ \hline
		\end{tabular}
	}%
\end{table}

The approach in this chapter is summarised in Table~\ref{tab:summary}.
Like the previous chapter, we address offline RL so the prior knowledge about the environment is in the form of a fixed dataset.
Unlike the previous chapter, which only considered epistemic uncertainty, in this chapter we propose a risk-sensitive approach to avoid outcomes that are plausible due to~\emph{either} epistemic or aleatoric uncertainty.
As shown in Table~\ref{tab:summary}, this is a similar core approach to that in Chapter~\ref{ch:5-risk-bamdps}.
However, there are a number of differences.
First, in Chapter~\ref{ch:5-risk-bamdps}, we considered the Bayes-adaptive MDP setting, where a known prior over MDPs represents the epistemic uncertainty.
In this chapter we learn a model from the offline dataset which approximates the epistemic uncertainty.
Second, in Chapter~\ref{ch:5-risk-bamdps} we proposed an approach for tabular problems. %
In this chapter, we apply function approximation to achieve a more scalable solution.
Finally, in Chapter~\ref{ch:5-risk-bamdps} we aimed to find a policy that adapts to the true underlying MDP during execution. 
In this chapter, we find a Markovian policy offline and do not consider online adaptation.

Our approach in this chapter requires uncertainty estimation.
Specifically, we assume that we can compute an approximation to the posterior distribution over MDP models, given the dataset.
In practice, we use an ensemble of neural networks to represent the epistemic uncertainty over the MDP model.
We propose to optimise the dynamic risk under this posterior distribution over models.
Like Chapter~\ref{ch:5-risk-bamdps}, this means that the policy is trained to be risk-averse towards both the epistemic uncertainty over the true MDP model, in addition to the aleatoric uncertainty predicted by each plausible MDP model.

To achieve risk-aversion, we consider the dynamic risk perspective, where the desired risk measure is applied recursively at each time step.
In the paper, we discuss the motivation for choosing to consider dynamic rather than static risk in this context.
The dynamic risk is equivalent to the expected value under a transition distribution that is perturbed independently at each step.
In our approach, to generate synthetic data for training the policy, we sample multiple candidate successor states from the ensemble.
We re-weight the distribution of successor states according to the chosen risk measure, and this re-weighted distribution over transitions is added to the replay buffer for training the policy.

In the offline RL setting, risk-aversion to epistemic uncertainty discourages the policy from visiting areas not covered by the dataset, as in these areas epistemic uncertainty is high.
Thus, risk-aversion to epistemic uncertainty mitigates the issue of distributional shift which is a core focus of many works on offline RL.
Risk-aversion to aleatoric uncertainty ensures that poor outcomes are avoided in stochastic environments.

Our experiments show that our algorithm achieves competitive performance relative to state-of-the-art baselines on deterministic benchmarks, and outperforms existing approaches for risk-sensitive offline RL on stochastic domains.

\noindent\rule{6cm}{0.5pt} 

\medskip
\noindent Marc Rigter, Bruno Lacerda and Nick Hawes (2022). One Risk to Rule Them All: A Risk-Sensitive Perspective on Model-Based Offline Reinforcement Learning. \textit{arXiv preprint arXiv:2212.00124}.
\medskip

\includepdf[pages=-]{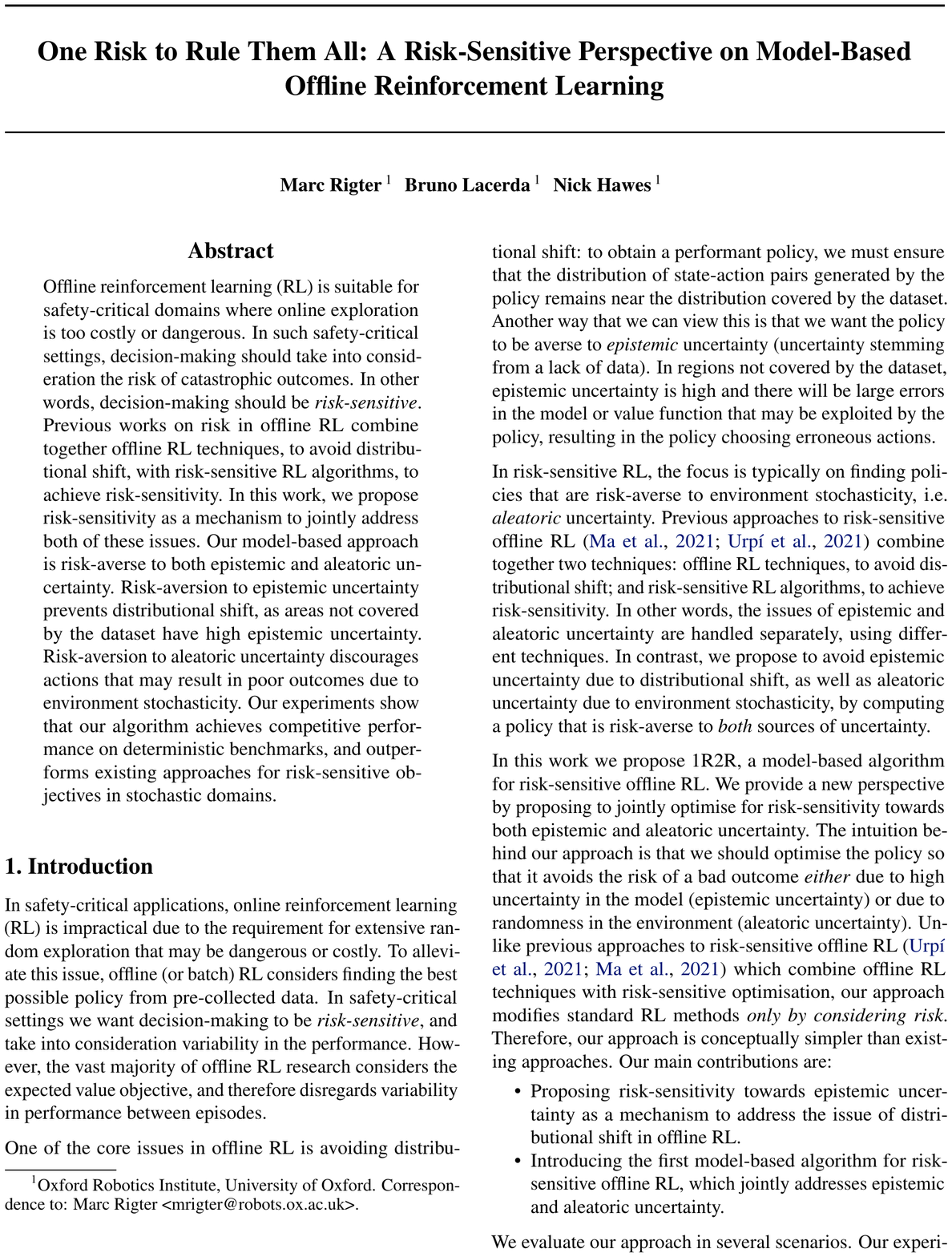}

\section{Limitations and Future Work}
We chose to base our approach on dynamic risk, which evaluates risk recursively at each step, rather than static risk, which evaluates risk across episodes.
As discussed in the paper, our motivation for this decision was that: a) dynamic risk is more suitable for avoiding distributional shift, as it enforces aversion to uncertainty at each time step, thus discouraging the policy from leaving the data-distribution at any step; and b) optimising dynamic risk is more straightforward for our model-based algorithm, as it only requires the transition function to be modified independently at each step.
The main disadvantage of this approach is that it is difficult to interpret what the algorithm ultimately optimises.
This makes it unclear how we should choose the risk measure to apply.
In our work, we circumvented this issue by treating the risk measure as a hyperparameter, which we tuned for the best performance for a static risk objective.
However, it would be preferable for the user to be able to choose a desired static risk objective upfront, and avoid this tuning procedure.
This is especially important for offline RL where hyperparameter tuning is a key issue: ideally the hyperparameters should be chosen offline without evaluation in the real environment.

A key direction for future work is improving the models of uncertainty.
We used a uniform distribution over a neural network ensemble, where the transition distribution for each model is Gaussian.
This prohibits our approach from being suitable for problems with multimodal stochasticity.
Furthermore, neural network ensembles may poorly estimate the uncertainty over the MDP dynamics~\cite{lu2022revisiting}.
Our risk-sensitive approach is agnostic to the source of uncertainty, motivated by the intuition that the worst plausible outcomes should be avoided, regardless of the source of uncertainty.
However, this approach may fail if the uncertainty estimates are poorly calibrated.
For example, if the epistemic uncertainty is underestimated, the agent may not be sufficiently discouraged from entering regions of the state-action space whether the model is highly inaccurate.

To overcome this issue, one approach is to adjust the level of aversion to each source of uncertainty, so that it is possible to obtain strong performance in the presence of poorly calibrated uncertainty estimates.
This could be achieved by making use of~\emph{composite} risk measures~\cite{eriksson2022sentinel}.
The disadvantage of this approach is that it would increase the number of hyperparameters, as there would be a separate hyperparameter related to each source of uncertainty.
Thus, adapting 1R2R to use composite risk measures may result in an even greater reliance on online hyperparameter tuning.

A final shortcoming of our work is that in some of the domains we created, particularly the Stochastic MuJoCo domains, there is a strong correlation between average performance and the performance for the risk-sensitive objective.
This might be because the optimal risk-sensitive policy in these domains is similar (or the same) as the optimal expected value policy.
This made it difficult to assess whether our algorithm generates different behaviour than risk-neutral algorithms. 
To confirm that our algorithm generates risk-sensitive behaviour, we created the Currency Exchange domain, which has a distinct tradeoff between risk and expected performance.
However, this is a toy domain which is not representative of real world problems.
A useful direction for future work would be to develop more realistic benchmarks for safety-critical domains, which are tailored towards evaluating risk-sensitivity and/or the robustness of offline RL algorithms.
This would be analogous to Safety Gym~\cite{ray2019benchmarking}, a set of benchmarks that evaluate safe exploration in online RL.

 

%% file: text/ch8-lfd.tex
\chapter{\label{ch:8-lfd} A Framework for Learning from Demonstration with Minimal Human Effort}
\chaptermark{Learning from Demonstration with Minimal Human Effort}

Offline RL, which we have addressed in the previous two chapters, is one approach to circumventing the extensive exploration required by online RL.
 In this final piece of work, we consider the alternative perspective of utilising expert demonstrations to quickly train a policy.
In this chapter, we fine-tune the policy using online RL, but little exploration is required as our approach predominantly relies on demonstrations to direct the policy towards good actions.
Most works on learning from demonstration assume that a fixed batch of demonstrations is provided upfront.
Our work proposes a novel interactive setting where demonstrations are requested as necessary, in an effort to minimise the human input required during training.

We focus on robotics tasks with a binary outcome of success or failure.
Failing to complete a task requires the system to be reset by a human operator. 
Examples of such failures could include the robot getting stuck during an inspection task, or dropping a package in a factory setting.
Furthermore, we assume that the robot is being trained in a real environment, such as a factory, and therefore we cannot control the order in which tasks appear (i.e. we cannot perform curriculum learning). 

\begin{table}[hb!t]
	\renewcommand{\arraystretch}{1.3}
	\hyphenpenalty=100000
	\caption*{\textbf{Table~\ref{tab:summary}:} \tablecaption }
	\footnotesize
	\resizebox{\columnwidth}{!}{%
		\begin{tabular}{l|b{1.7cm}|b{1.7cm}|b{1.7cm}|b{1.7cm}|b{1.7cm}|a{1.9cm}|}
			\hhline{~|-|-|-|-|-|-|}
			&  \textbf{Chapter~\ref{ch:3-risk-mdps}} &  \textbf{Chapter~\ref{ch:4-regret-mdps}}  & \textbf{Chapter~\ref{ch:5-risk-bamdps}} & \textbf{Chapter~\ref{ch:6-rambo}} & \textbf{Chapter~\ref{ch:7-1r2r}} & \textbf{Chapter~\ref{ch:8-lfd}} \\ \hline
			\multicolumn{1}{|m{1.9cm}|}{\textbf{Prior knowledge}} & MDP fully known & MDP uncertainty set & Prior over MDPs & Fixed dataset & Fixed dataset & Expert demonstrator \\ \hline
			\multicolumn{1}{|m{1.9cm}|}{\textbf{Considers aleatoric uncertainty}} & Yes & No & Yes &  No & Yes & No \\ \hline
			\multicolumn{1}{|m{1.9cm}|}{\textbf{Considers epistemic uncertainty}} & No & Yes & Yes & Yes & Yes & Yes \\ \hline
			\multicolumn{1}{|m{1.9cm}|}{\textbf{Risk or robustness}} & Risk &  Robustness  & Risk & Robustness & Risk & N/A \\ \hline
			\multicolumn{1}{|m{1.9cm}|}{\textbf{Paradigm}} & Tabular &  Tabular  & Tabular & Function Approx. & Function Approx. & Function Approx. \\ \hline
		\end{tabular}
	}%
\end{table}
\FloatBarrier

Our goal in this piece of work is to develop a framework for interactively training the policy, while minimising the human effort required.
We assume that human effort is required to generate expert demonstrations, and to reset the system from failures.
To model this problem, we assume that there is a cost for generating a demonstration, as well as a cost for resetting the robot from failure. 
The goal is to minimise the expected cost associated with human effort accumulated over many episodes.

Table~\ref{tab:summary} summarises this chapter in relation to the other chapters in this thesis.
In this chapter, the source of knowledge used to train the policy is the access to the expert demonstrator.
We seek to optimise the human-related cost in expectation, and therefore this approach does not consider aleatoric uncertainty.
Our approach does consider epistemic uncertainty.
The epistemic uncertainty stems from the fact that we do not know which tasks the current policy performs well at, as opposed to the tasks where the policy would benefit from more demonstrations.
As indicated by Table~\ref{tab:summary}, our approach in this work is neither risk-averse nor robust to this epistemic uncertainty.
Instead, we use ``optimism in the face of uncertainty'' to explore areas of high epistemic uncertainty to obtain a greater understanding of the performance of the current policy.
Furthermore, the approach that we present in this chapter is entirely model-free, in contrast to all of the other chapters of this thesis.

In our proposed approach, we formulate the problem as a contextual multi-armed bandit.
The context for the problem is represented by the initial state of the environment for each episode, and this encodes the difficulty of the current task.
For example, the context could be the size and position of a package which is to be manoeuvred. 
Based on experience from previous episodes, we train a classifier that attempts to predict the probability that the current policy will succeed at the task, as well as provide an estimate of the uncertainty in this prediction.
To capture the fact that the policy is changing over time, we only use a window of recent experiences to make this prediction.
Furthermore, we also allow for the possibility that there are multiple autonomous controllers to choose from, including the policy currently being trained.

For each new episode, we use the classifier to estimate the failure cost associated with each controller, given the context for that episode.
Based on this estimate, we use an upper-confidence bound algorithm to decide whether autonomous control should be selected or whether a human demonstration should be requested.
The result is that demonstrations are requested for cases where the policy does not perform well, and the system chooses to rely on the autonomous policy in situations where it reliably succeeds.
In situations where there is high uncertainty over the performance of the policy, the upper-confidence bound algorithm encourages trialling the policy to better understand it's current performance.

We test our approach on a navigation domain and a robot arm manipulation domain. 
Our experiments verify that our approach reduces the human cost incurred over many episodes relative to baselines. 
We also validate our approach on a real-world robotic task.

\noindent\rule{6cm}{0.5pt} 

\medskip
\noindent Marc Rigter, Bruno Lacerda, and Nick Hawes (2020). A framework for learning from demonstration with minimal human effort. \textit{IEEE Robotics and Automation Letters} (RAL).

\includepdf[pages=-]{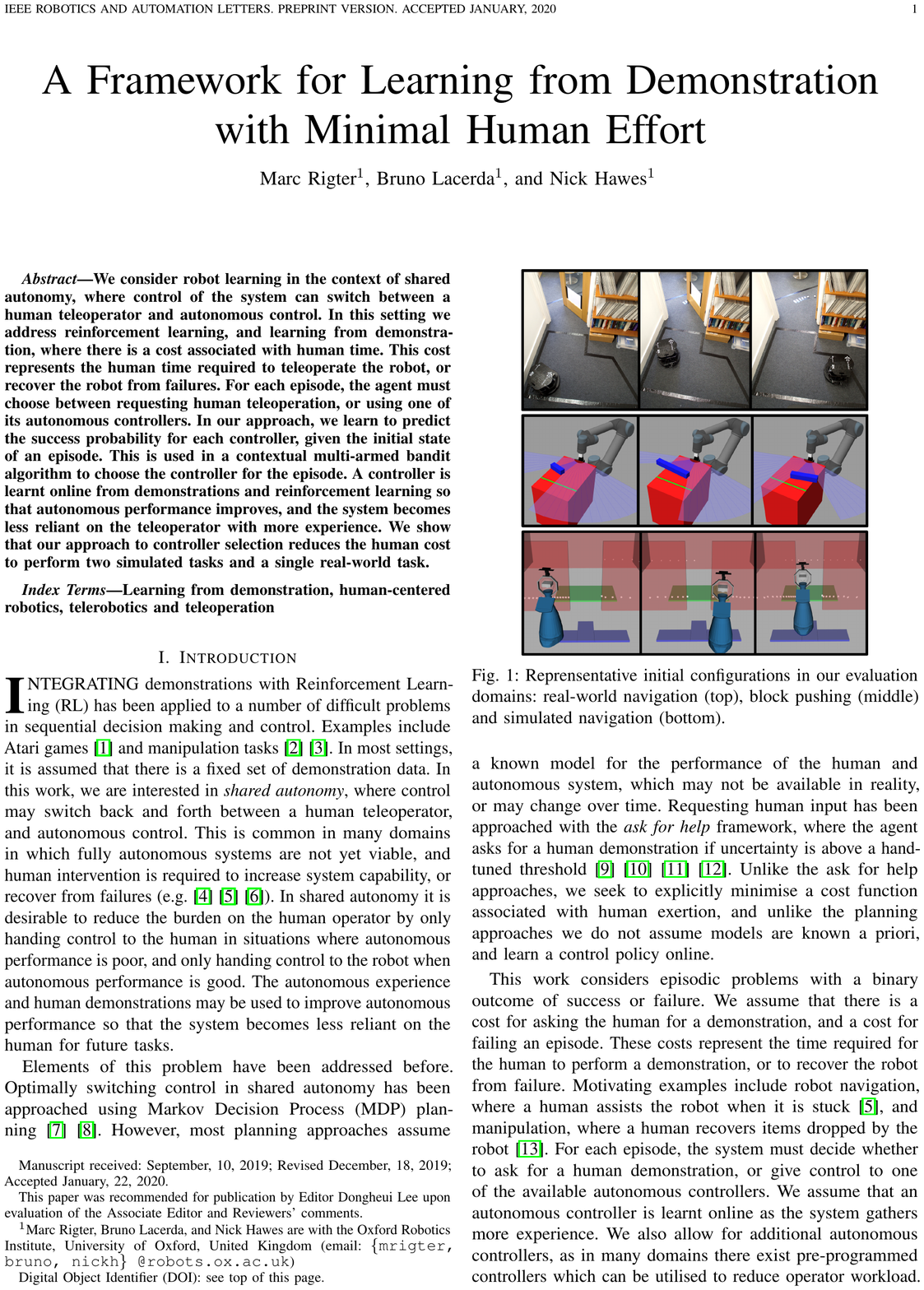}

\section{Limitations and Future Work}
There are several limitations to the approach that we proposed in this work.
One is the simplistic approach that we took towards modelling the human in our framework.
We assumed that the workload for the human can be modelled as fixed costs for resetting failures and for demonstrations.
However, in reality the perceived workload for the human operator might depend on many factors. 
For example, periodically interrupting the human to request demonstrations may be more inconvenient than providing a large number of demonstrations upfront.
Furthermore, requesting help from the human may be a greater burden if the human is currently busy.
A more flexible approach that takes into account these additional human factors could be a challenging, but interesting direction for future work.
Such research would likely require trials with real human operators to understand how the approach could be tailored towards human preferences.

The main limitation of the algorithm we proposed is that it is myopic in the sense that it determines what will minimise the cost based only on the current performance of the policy.
It does not take into account the fact that the policy is able to improve, so providing a demonstration at the current episode will improve the policy in the long run, and prevent cost incurred due to future failures.
Quantifying the improvement in the policy that can be expected by gathering additional demonstrations (as well as the associated reduction in future cost) appears to be a very difficult problem. 
However, this would be an interesting direction to try and improve upon the framework that we proposed.

A recent piece of work~\cite{vats2022synergistic} tries to address this. 
The authors train a classifier to predict which tasks can be solved after receiving a demonstration for a given input task. 
This can then be used to weigh the long-term benefit of requesting additional demonstrations.
However, this approach requires a large dataset of demonstrations across many tasks to train the classifier, which appears to undermine the original goal of reducing the human workload.
A simpler alternative may be to include a reward bonus for receiving a demonstration to reflect the improvement in long-term cost.
However, appropriately designing the reward bonuses would be very difficult.

A key attribute of our approach is the use of uncertainty estimation.
We use uncertainty estimates to determine which tasks we are certain the robot will fail at, and therefore requires a demonstration.
We also use the uncertainty estimates to determine which tasks we are unsure about the performance of the current policy, and therefore it might be worth attempting to use the policy.
We used a simple kernel-based method for uncertainty estimation.
In future work, our approach could be combined with uncertainty estimation techniques for powerful deep learning models to address more complex problems.

A final limitation of our approach is that it is only applicable to tasks with a binary outcome of success or failure.
A direction for future work would be to develop an approach for tasks where the outcome for an episode is summarised by the cumulative reward received.
This would lead to a multi-objective problem, where the human effort cost must be traded off against the reward received.
Another direction would be to address problems where control may switch back and forth between human and demonstrator multiple times within an episode~\cite{costen2022shared}.


%% file: text/ch9-conclusion.tex
\chapter{\label{ch:9-conclusion} Conclusion}

This thesis has contributed towards developing algorithms for sequential decision-making that are able to reckon with uncertainty in a number of different problem settings.
In Chapter~\ref{ch:3-risk-mdps}, we considered problems where the MDP is fully specified and we wish to optimise CVaR to be risk-averse to the stochastic outcomes of the MDP.
We developed an approach to achieve the best possible expected performance whilst ensuring CVaR is optimal.
In Chapters~\ref{ch:4-regret-mdps} and~\ref{ch:5-risk-bamdps} we considered problems where there is uncertainty over the MDP specified by a set of MDPs, or a prior distribution over MDPs, respectively.
In Chapter~\ref{ch:4-regret-mdps} we developed approximations to finding a policy with the lowest maximum sub-optimality, or regret, over the set of MDPs.
In Chapter~\ref{ch:5-risk-bamdps} we considered the optimisation of CVaR in the presence of the uncertain distribution over MDPs to mitigate risk due to both epistemic and aleatoric uncertainty.
In Chapters~\ref{ch:6-rambo} and~\ref{ch:7-1r2r}, we proposed new algorithms for offline RL.
Our approach in Chapter~\ref{ch:6-rambo} is based on developing a scalable approach to solving a type of robust MDP.
In Chapter~\ref{ch:7-1r2r}, we proposed an approach that tackles the problem of offline RL by finding a policy that is risk-averse to both epistemic and aleatoric uncertainty, in a similar spirit to Chapter~\ref{ch:5-risk-bamdps}.
Finally, in Chapter~\ref{ch:8-lfd} we considered the learning from demonstration setting where we must reason about the uncertain performance of the learnt policy to minimise the human effort required to train the policy. 

\section{Discussion}
In this thesis we have considered both tabular approaches, and approaches that utilise deep learning.
There are clear tradeoffs between the two approaches.
On the one hand, tabular approaches can often guarantee an exact solution, and that solution can be easily interpreted.
However, these approaches cannot be scaled beyond small problems. 
On the other hand, deep RL is considerably more scalable, but has the downside that it can be sensitive to hyperparameters, may require abundant training data, and lacks guarantees on the performance of the final policy.
It remains unclear whether there is a satisfactory middle ground to this tradeoff, but we think that conducting research under both paradigms is worthwhile.
Addressing tabular problems is often a better way to understand the core elements of the problem at hand, and is a useful foundation from which to try and develop algorithms that are more scalable.

One of the core arguments that we have made throughout this thesis is that when designing sequential decision-making algorithms, we should take into consideration both epistemic and aleatoric uncertainty.
Much research on both planning and RL disregards epistemic uncertainty by either assuming access to a perfect model, or to a perfect simulator from which unlimited experience can be obtained.
However, for many real problems there is no perfect model or simulator, and we only have access to limited data from the environment.
Therefore, there will always be inaccuracies in any model, simulator, or value function derived from this data.
By explicitly reasoning about model uncertainty in the model-based approaches that we have considered, we can ensure that the policies we generate make sound decisions even in the presence of limited knowledge about the environment.

Furthermore, going beyond optimising for the expectation and considering risk-sensitive objectives is crucial for mitigating aleatoric risk in stochastic environments.
Much of the current research in sequential decision-making, particularly in deep RL, focuses on deterministic benchmark domains. 
We think that more emphasis should be placed on evaluating the performance of algorithms, including their variability in performance, in highly stochastic domains which are more representative of the real world.

Rather than considering each uncertainty source in isolation, like much of the previous work in this area, we think that more research should investigate approaches that jointly consider both epistemic and aleatoric uncertainty. 
After all, our goal should be to develop policies that avoid catastrophic outcomes, irrespective of the source of uncertainty that led to that outcome. 
Focussing on this goal may improve progress toward developing algorithms that can be deployed safely in the real world in stochastic and safety-critical environments with access to limited data.

\section{Future Work}
To close, we discuss some concrete research directions that lead on from the work  in this thesis.
In each chapter, we have already discussed the limitations of each individual piece of work, and suggested future work to address those limitations.
In this final section, we focus on directions for future work that draw connections between multiple chapters.

\paragraph{Better Models} An orthogonal direction to the research presented in this thesis is to develop better machine learning models for capturing uncertainty.
Throughout this thesis, we have used methods for representing epistemic uncertainty over the MDP model that are either difficult to scale (discrete sets of MDPs, Dirichlet distributions), or may be unreliable (ensembles of neural networks).
Furthermore, to capture aleatoric uncertainty, our neural network models used Gaussian distributions over successor states which may be highly inaccurate for some problems, particularly those with multi-modal dynamics.
Therefore, developing machine learning models that are scalable, and can accurately model both epistemic and aleatoric uncertainty, is crucial to improving the practical effectiveness of the approaches presented in this work.
Integrating recent advances in these areas~\cite{gawlikowski2021survey} with the algorithms presented in this thesis is a straightforward next step for future work. 

\paragraph{Adaptive Risk-Averse Offline RL} In Chapter~\ref{ch:5-risk-bamdps} we generated policies that are risk-averse to epistemic and aleatoric uncertainty, but adapt their behaviour as epistemic uncertainty is reduced during online execution.
In Chapter~\ref{ch:7-1r2r} we addressed risk-aversion to epistemic and aleatoric uncertainty for offline RL, but trained a Markovian policy which did not allow for adaptivity.
Throughout this thesis, we have argued that risk-averse algorithms are crucial for the safety-critical applications in which offline RL is most applicable.
Meanwhile, concurrent works have argued for training adaptive policies for offline RL~\cite{dorfman2021offline, ghosh2022offline}, as this helps to overcome the limitations of a fixed dataset by enabling the policy to quickly adapt to the true environment.

Thus, a natural direction for future work is to combine the  concepts of risk-sensitivity and adaptiveness in the context of offline RL, and seek to train risk-averse policies that are adaptive. 
Adaptivity might help enable the generation of more performant policies from very limited datasets, while incorporating risk-sensitivity might make these approaches more applicable to safety-critical environments. 

\paragraph{Minimax Regret for Offline RL} In Chapter~\ref{ch:6-rambo}, we developed an algorithm to solve a robust MDP formulation of offline RL, where we found the policy with the best worst-case expected value across a set of candidate MDPs.
In Chapter~\ref{ch:4-regret-mdps}, we addressed the minimax regret criterion for an uncertainty set of MDPs.
The argument for considering the minimax regret criterion, as opposed to maximin expected value (i.e. robust MDPs), is that minimax regret is less conservative but can still be considered robust as it optimises for the worst-case sub-optimality.

A possible direction for future work is to address the minimax regret objective in the deep offline RL setting. 
This would involve developing a more scalable approach for optimising minimax regret, something that we have already discussed in Section~\ref{sec:minimax_reg_future_work}.
A potential advantage of this approach is that it might result in more aggressive and performant policies than existing approaches which are usually very pessimistic and consider either worst-case optimisation (such as Chapter~\ref{ch:6-rambo}), or other methods that lower-bound the value function in the true environment~\cite{kumar2020conservative, yu2021combo}.
By considering the worst-case regret across all plausible environments, one would expect that the resulting policy should still be robust to environment uncertainty.

\paragraph{Pareto Fronts for Risk-Sensitive Policies} In Chapter~\ref{ch:3-risk-mdps} we explored the fact that optimising for risk alone may not achieve the best possible tradeoff with the expected value.
In Chapter~\ref{ch:3-risk-mdps} we proposed to optimise for the best expected value, given that performance is optimal for a chosen risk-level.
However, it is not clear how to choose the level of risk to optimise for that obtains the desired tradeoff between these two objectives.

Therefore, a possible future direction is to instead focus on developing scalable algorithms that can efficiently generate a Pareto front of policies that obtain different tradeoffs between risk and expected value.
After observing the behaviours of the different policies, the desired tradeoff may be selected.
This approach might additionally be useful in the offline RL setting, where it has been shown that generating a Pareto front of diverse policies with different levels of pessimism and selecting the best policy can be used to obtain very strong performance~\cite{yang2021pareto}.
Generating and selecting between multiple candidate policies with different levels of risk-aversion may be a promising alternative.

\paragraph{Robustness and Risk-Sensitivity for Human-Robot Interaction} In Chapter~\ref{ch:8-lfd} we briefly touched upon human-robot interaction, in the form of interactively training a policy using demonstrations.
In Chapter~\ref{ch:8-lfd}, we did not consider any form of risk-sensitivity or robustness, and furthermore we assumed that the human demonstrator never made any mistakes.
In future work, we think that tasks involving human-robot interaction and collaboration are a natural domain in which to apply some of the ideas developed in this thesis.

There are a number of ways in which risk-sensitivity and/or robustness could be applied to this setting.
One possible direction would be to consider robustness or risk-sensitivity towards human behaviour.
Human behaviour is both unpredictable, and variable between humans, and therefore cannot easily be specified by a hand-crafted model~\cite{costen2022shared}.
Therefore, when developing algorithms for human robot-interaction, the human should be modelled using data-driven approaches.
Optimising for a risk-sensitive or robust objective in the presence of an uncertain human model that is derived from data, in the spirit of Chapters~\ref{ch:6-rambo} and~\ref{ch:7-1r2r}, may be an effective means to handle the high degree of unpredictability of humans in collaborative tasks.

Furthermore, adaptive policies like those obtained in Chapter~\ref{ch:5-risk-bamdps} will be useful for adapting to the unique characteristics of each human, when interacting with multiple humans.
Recent work~\cite{costen2022shared} has begun to explore this direction.

\paragraph{Further Applications} We wish to apply the ideas in this thesis to other real-world problems, aside from human-robot interaction. 
In robotics, some possibilities include applying the offline RL algorithms developed in this thesis to real systems~\cite{gurtler2023benchmarking}, or using algorithms for robust or risk-sensitive policy optimisation to improve sim-to-real transfer~\cite{zhao2020sim}.
Other possible application domains include finance and healthcare.
In these domains, accurate simulators do not exist, and incorporating risk and uncertainty into decision-making is paramount.

The goal of this thesis is to make a small contribution towards the widespread application of planning and reinforcement learning algorithms to real world problems.
Our hope is that if, and when, these algorithms become widely used, their benefits are distributed equitably.

%% file: text/appendices.tex

\renewcommand{\appendixpagename}{Appendices of Chapter~\ref{ch:3-risk-mdps}}
\renewcommand{\appendixtocname}{Appendices of Chapter~\ref{ch:3-risk-mdps}}
\appendixpage
\addappheadtotoc

\includepdf[pages=-]{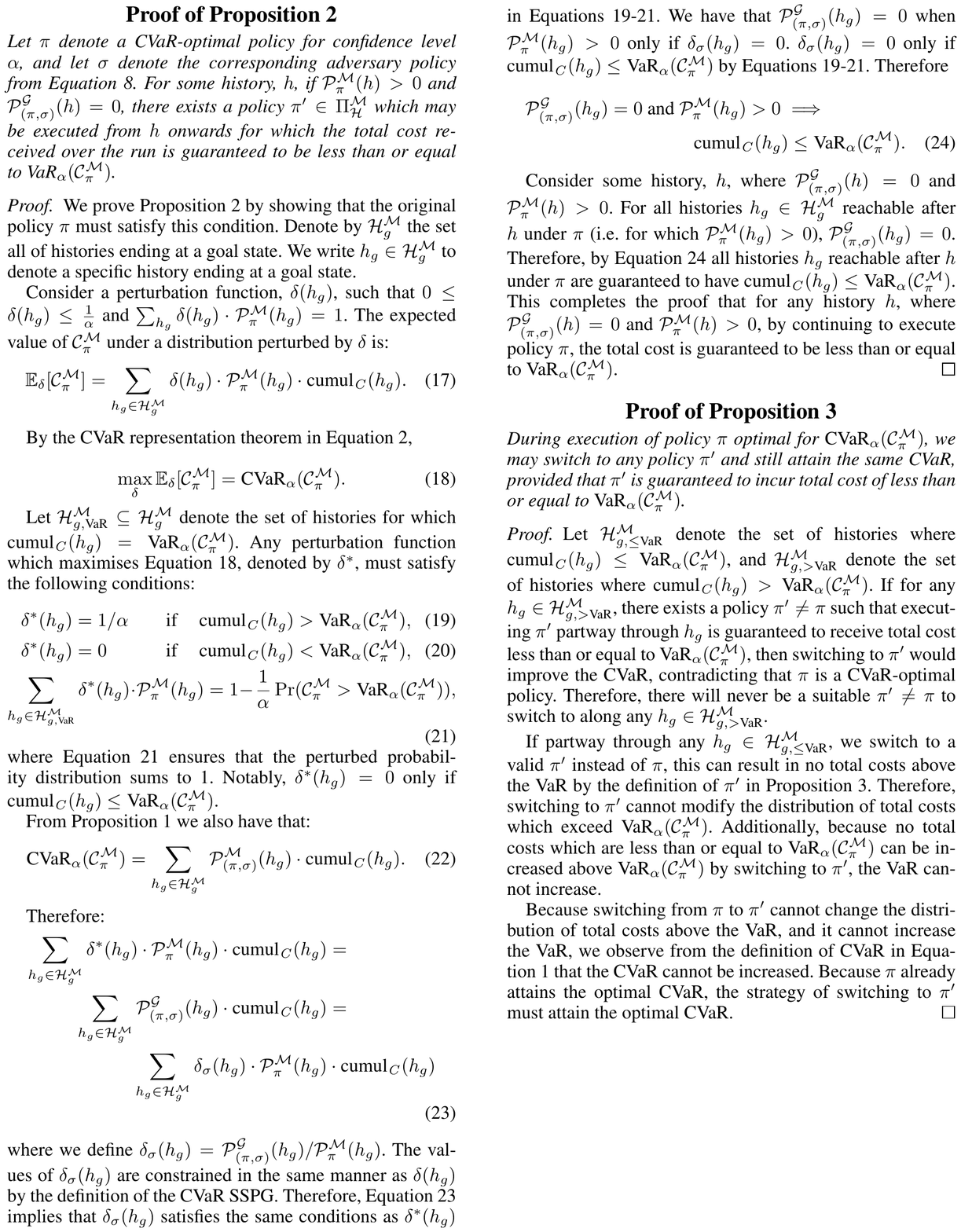}

\renewcommand{\appendixpagename}{Appendices of Chapter~\ref{ch:4-regret-mdps}}
\renewcommand{\appendixtocname}{Appendices of Chapter~\ref{ch:4-regret-mdps}}
\appendixpage
\addappheadtotoc

\includepdf[pages=-]{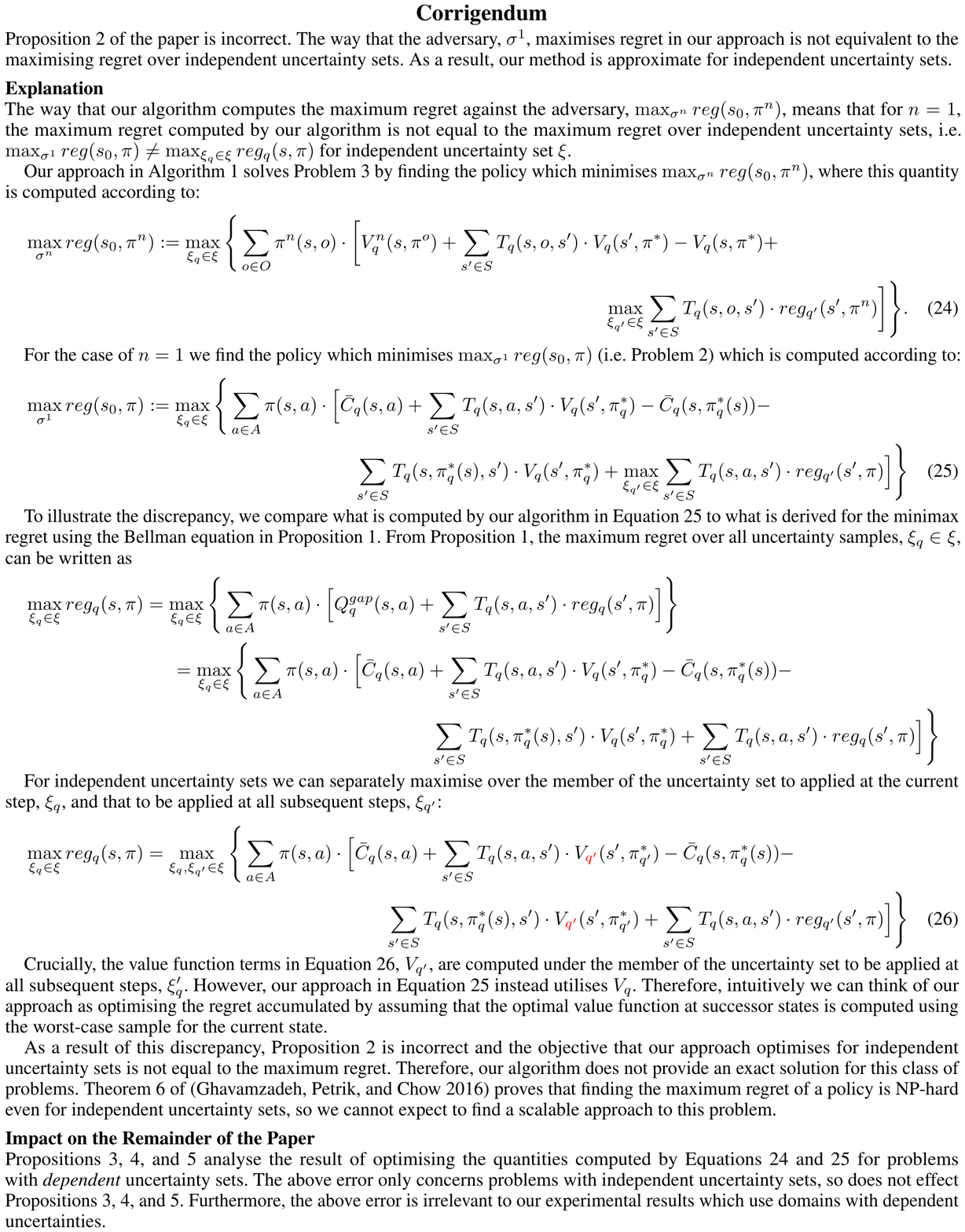}

\renewcommand{\appendixpagename}{Appendices of Chapter~\ref{ch:5-risk-bamdps}}
\renewcommand{\appendixtocname}{Appendices of Chapter~\ref{ch:5-risk-bamdps}}
\appendixpage
\addappheadtotoc

\includepdf[pages=-]{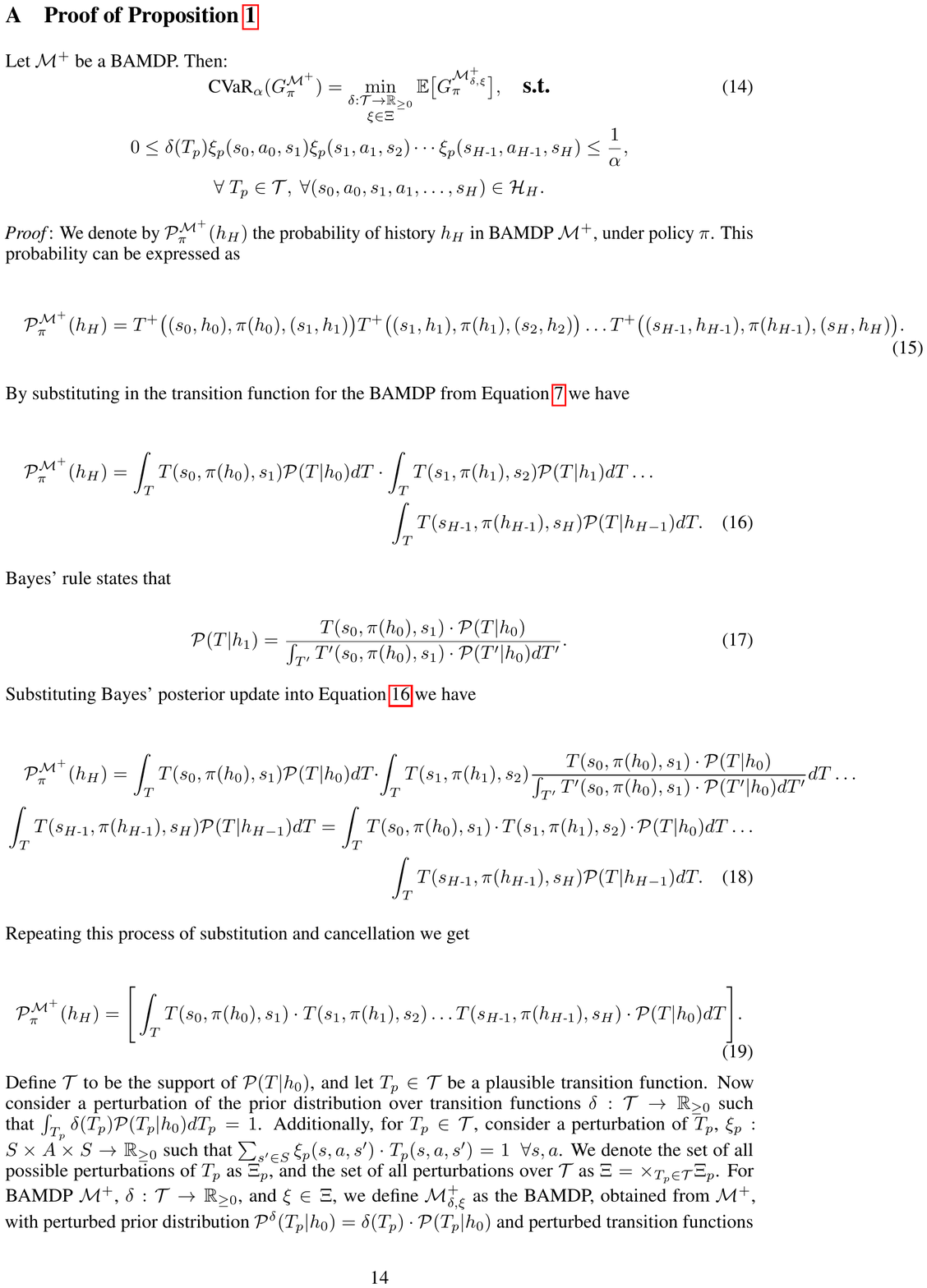}

\renewcommand{\appendixpagename}{Appendices of Chapter~\ref{ch:6-rambo}}
\renewcommand{\appendixtocname}{Appendices of Chapter~\ref{ch:6-rambo}}
\appendixpage
\addappheadtotoc

\includepdf[pages=-]{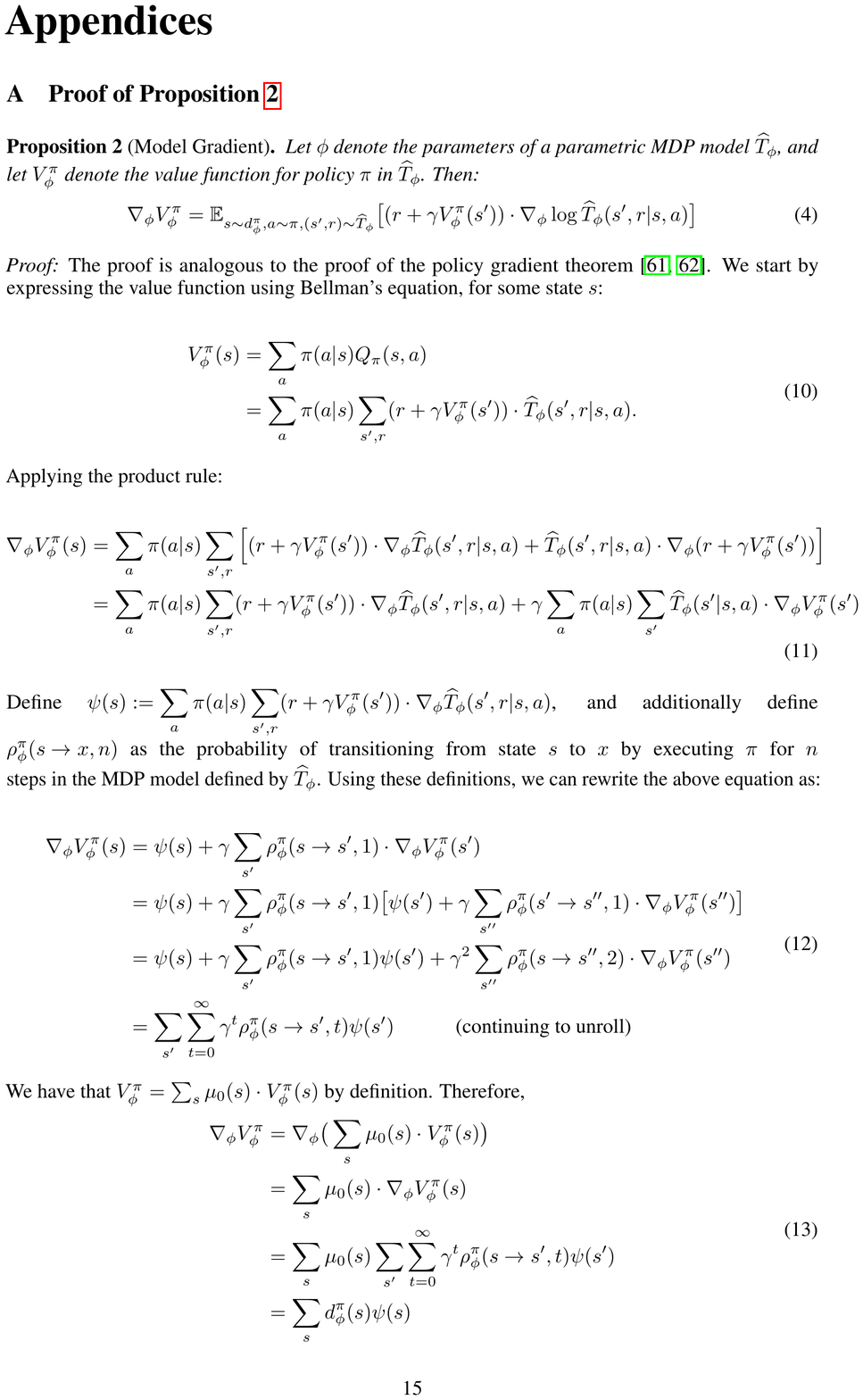}

\renewcommand{\appendixpagename}{Appendices of Chapter~\ref{ch:7-1r2r}}
\renewcommand{\appendixtocname}{Appendices of Chapter~\ref{ch:7-1r2r}}
\appendixpage
\addappheadtotoc

\includepdf[pages=-]{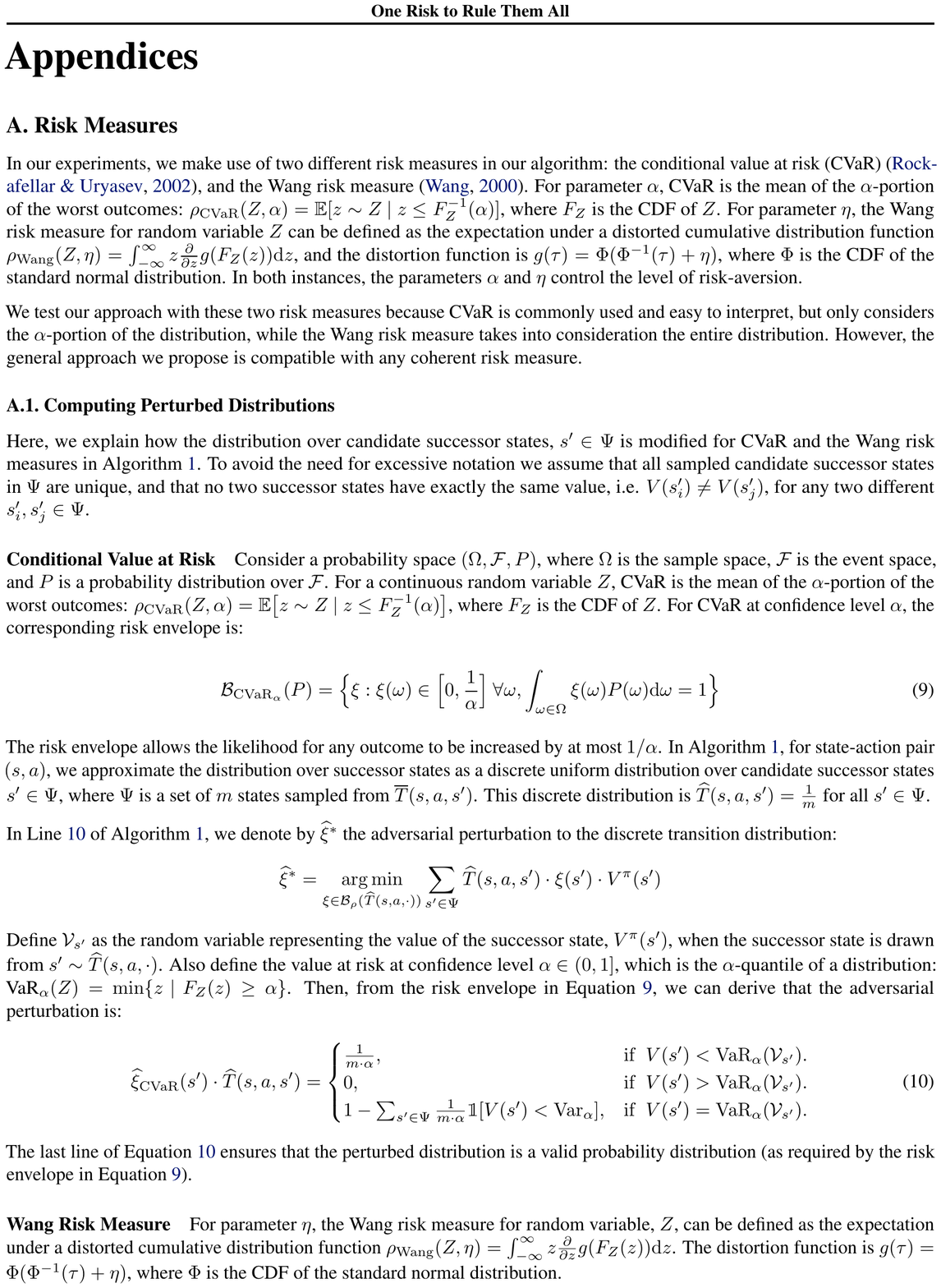}


%% file: thesis.bbl
\begin{thebibliography}{100}

\bibitem{maddern20171}
Will Maddern, Geoffrey Pascoe, Chris Linegar, and Paul Newman.
\newblock 1 year, 1000 km: The {Oxford} {RobotCar} dataset.
\newblock {\em The International Journal of Robotics Research}, 36(1):3--15,
  2017.

\bibitem{tseng2017deep}
Huan-Hsin Tseng, Yi~Luo, Sunan Cui, Jen-Tzung Chien, Randall~K Ten~Haken, and
  Issam~El Naqa.
\newblock Deep reinforcement learning for automated radiation adaptation in
  lung cancer.
\newblock {\em Medical physics}, 44(12):6690--6705, 2017.

\bibitem{charbonnier2022scalable}
Flora Charbonnier, Thomas Morstyn, and Malcolm~D McCulloch.
\newblock Scalable multi-agent reinforcement learning for distributed control
  of residential energy flexibility.
\newblock {\em Applied Energy}, 314:118825, 2022.

\bibitem{spooner2018market}
Thomas Spooner, John Fearnley, Rahul Savani, and Andreas Koukorinis.
\newblock Market making via reinforcement learning.
\newblock {\em International Conference on Autonomous Agents and Multiagent
  Systems}, 2018.

\bibitem{frey2017future}
Carl~Benedikt Frey and Michael~A Osborne.
\newblock The future of employment: How susceptible are jobs to
  computerisation?
\newblock {\em Technological Forecasting and Social Change}, 114:254--280,
  2017.

\bibitem{majumdar2020should}
Anirudha Majumdar and Marco Pavone.
\newblock How should a robot assess risk? {Towards} an axiomatic theory of risk
  in robotics.
\newblock In {\em Robotics Research}, pages 75--84. Springer, 2020.

\bibitem{berkenkamp2017safe}
Felix Berkenkamp, Matteo Turchetta, Angela Schoellig, and Andreas Krause.
\newblock Safe model-based reinforcement learning with stability guarantees.
\newblock {\em Advances in Neural Information Processing Systems}, 30, 2017.

\bibitem{khan2009minimal}
Omar Khan, Pascal Poupart, and James Black.
\newblock Minimal sufficient explanations for factored {Markov} decision
  processes.
\newblock In {\em Proceedings of the International Conference on Automated
  Planning and Scheduling}, volume~19, pages 194--200, 2009.

\bibitem{hafner2020mastering}
Danijar Hafner, Timothy Lillicrap, Mohammad Norouzi, and Jimmy Ba.
\newblock Mastering atari with discrete world models.
\newblock {\em International Conference on Learning Representations}, 2021.

\bibitem{mannor2007bias}
Shie Mannor, Duncan Simester, Peng Sun, and John~N Tsitsiklis.
\newblock Bias and variance approximation in value function estimates.
\newblock {\em Management Science}, 53(2):308--322, 2007.

\bibitem{knight1921risk}
Frank~Hyneman Knight.
\newblock {\em Risk, uncertainty and profit}, volume~31.
\newblock Houghton Mifflin, 1921.

\bibitem{iyengar2005robust}
Garud~N Iyengar.
\newblock Robust dynamic programming.
\newblock {\em Mathematics of Operations Research}, 30(2):257--280, 2005.

\bibitem{nilim2005robust}
Arnab Nilim and Laurent El~Ghaoui.
\newblock Robust control of {Markov} decision processes with uncertain
  transition matrices.
\newblock {\em Operations Research}, 53(5):780--798, 2005.

\bibitem{mannor2016robust}
Shie Mannor, Ofir Mebel, and Huan Xu.
\newblock Robust {MDPs} with k-rectangular uncertainty.
\newblock {\em Mathematics of Operations Research}, 41(4):1484--1509, 2016.

\bibitem{dilokthanakul2018deep}
Nat Dilokthanakul and Murray Shanahan.
\newblock Deep reinforcement learning with risk-seeking exploration.
\newblock In {\em International Conference on Simulation of Adaptive Behavior},
  pages 201--211. Springer, 2018.

\bibitem{rigter2020framework}
Marc Rigter, Bruno Lacerda, and Nick Hawes.
\newblock A framework for learning from demonstration with minimal human
  effort.
\newblock {\em IEEE Robotics and Automation Letters}, 5(2):2023--2030, 2020.

\bibitem{rigter2021minimax}
Marc Rigter, Bruno Lacerda, and Nick Hawes.
\newblock Minimax regret optimisation for robust planning in uncertain {Markov}
  decision processes.
\newblock In {\em Proceedings of the AAAI Conference on Artificial
  Intelligence}, volume~35, pages 11930--11938, 2021.

\bibitem{rigter2021risk}
Marc Rigter, Bruno Lacerda, and Nick Hawes.
\newblock Risk-averse {Bayes}-adaptive reinforcement learning.
\newblock {\em Advances in Neural Information Processing Systems},
  34:1142--1154, 2021.

\bibitem{rigter2022planning}
Marc Rigter, Paul Duckworth, Bruno Lacerda, and Nick Hawes.
\newblock Planning for risk-aversion and expected value in {MDPs}.
\newblock {\em International Conference on Automated Planning and Scheduling},
  2022.

\bibitem{rigter2022rambo}
Marc Rigter, Bruno Lacerda, and Nick Hawes.
\newblock {RAMBO-RL}: Robust adversarial model-based offline reinforcement
  learning.
\newblock {\em Advances in Neural Information Processing Systems}, 2022.

\bibitem{rigter2022one}
Marc Rigter, Bruno Lacerda, and Nick Hawes.
\newblock One risk to rule them all: A risk-sensitive perspective on
  model-based offline reinforcement learning.
\newblock {\em arXiv preprint arXiv:2212.00124}, 2022.

\bibitem{ishida2019robot}
Shu Ishida, Marc Rigter, and Nick Hawes.
\newblock Robot path planning for multiple target regions.
\newblock In {\em 2019 European Conference on Mobile Robots (ECMR)}, pages
  1--6. IEEE, 2019.

\bibitem{rigter2022optimal}
Marc Rigter, Danial Dervovic, Parisa Hassanzadeh, Jason Long, Parisa Zehtabi,
  and Daniele Magazzeni.
\newblock Optimal admission control for multiclass queues with time-varying
  arrival rates via state abstraction.
\newblock {\em AAAI Conference on Artificial Intelligence}, 2022.

\bibitem{costen2022shared}
Clarissa Costen, Marc Rigter, Bruno Lacerda, and Nick Hawes.
\newblock Shared autonomy systems with stochastic operator models.
\newblock In {\em International Joint Conferences on Artificial Intelligence},
  2022.

\bibitem{gautier2022risk}
Anna Gautier, Marc Rigter, Bruno Lacerda, Nick Hawes, and Michael Wooldridge.
\newblock Risk-constrained planning for multi-agent systems with shared
  resources.
\newblock {\em International Conference on Autonomous Agents and Multiagent
  Systems}, 2022.

\bibitem{costen2022planning}
Clarissa Costen, Marc Rigter, Bruno Lacerda, and Nick Hawes.
\newblock Planning with hidden parameter polynomial {MDPs}.
\newblock {\em AAAI Conference on Artificial Intelligence}, 2022.

\bibitem{puterman1994markov}
Martin~L Puterman.
\newblock {\em Markov decision processes: Discrete stochastic dynamic
  programming}.
\newblock John Wiley \& Sons, 1994.

\bibitem{sutton2018reinforcement}
Richard~S Sutton and Andrew~G Barto.
\newblock {\em Reinforcement learning: An introduction}.
\newblock MIT press, 2018.

\bibitem{bellman1966dynamic}
Richard Bellman.
\newblock Dynamic programming.
\newblock {\em Science}, 153(3731):34--37, 1966.

\bibitem{kolobov2012planning}
Andrey Kolobov et~al.
\newblock {\em Planning with Markov decision processes: An AI perspective}.
\newblock Morgan \& Claypool Publishers, 2012.

\bibitem{browne2012survey}
Cameron~B Browne, Edward Powley, Daniel Whitehouse, Simon~M Lucas, Peter~I
  Cowling, Philipp Rohlfshagen, Stephen Tavener, Diego Perez, Spyridon
  Samothrakis, and Simon Colton.
\newblock A survey of {Monte Carlo} tree search methods.
\newblock {\em IEEE Transactions on Computational Intelligence and AI in
  games}, 4(1):1--43, 2012.

\bibitem{sutton1999policy}
Richard~S Sutton, David McAllester, Satinder Singh, and Yishay Mansour.
\newblock Policy gradient methods for reinforcement learning with function
  approximation.
\newblock {\em Advances in Neural Information Processing Systems}, 12, 1999.

\bibitem{knox2021reward}
W~Bradley Knox, Alessandro Allievi, Holger Banzhaf, Felix Schmitt, and Peter
  Stone.
\newblock Reward (mis)design for autonomous driving.
\newblock {\em arXiv preprint arXiv:2104.13906}, 2021.

\bibitem{amodei2016concrete}
Dario Amodei, Chris Olah, Jacob Steinhardt, Paul Christiano, John Schulman, and
  Dan Man{\'e}.
\newblock Concrete problems in {AI} safety.
\newblock {\em arXiv preprint arXiv:1606.06565}, 2016.

\bibitem{smith2021exponential}
Kevin~M Smith and Margaret~P Chapman.
\newblock On exponential utility and conditional value-at-risk as risk-averse
  performance criteria.
\newblock {\em arXiv preprint arXiv:2108.01771}, 2021.

\bibitem{hwangbo2018per}
Jemin Hwangbo, Joonho Lee, and Marco Hutter.
\newblock Per-contact iteration method for solving contact dynamics.
\newblock {\em IEEE Robotics and Automation Letters}, 3(2):895--902, 2018.

\bibitem{garcia2015comprehensive}
Javier Garc{\i}a and Fernando Fern{\'a}ndez.
\newblock A comprehensive survey on safe reinforcement learning.
\newblock {\em Journal of Machine Learning Research}, 16(1):1437--1480, 2015.

\bibitem{kumar2021workflow}
Aviral Kumar, Anikait Singh, Stephen Tian, Chelsea Finn, and Sergey Levine.
\newblock A workflow for offline model-free robotic reinforcement learning.
\newblock {\em Conference on Robot Learning}, 2021.

\bibitem{tang2021model}
Shengpu Tang and Jenna Wiens.
\newblock Model selection for offline reinforcement learning: Practical
  considerations for healthcare settings.
\newblock In {\em Machine Learning for Healthcare Conference}, pages 2--35.
  PMLR, 2021.

\bibitem{argall2009survey}
Brenna~D Argall, Sonia Chernova, Manuela Veloso, and Brett Browning.
\newblock A survey of robot learning from demonstration.
\newblock {\em Robotics and autonomous systems}, 57(5):469--483, 2009.

\bibitem{osa2018algorithmic}
Takayuki Osa, Joni Pajarinen, Gerhard Neumann, J~Andrew Bagnell, Pieter Abbeel,
  Jan Peters, et~al.
\newblock An algorithmic perspective on imitation learning.
\newblock {\em Foundations and Trends in Robotics}, 7(1-2):1--179, 2018.

\bibitem{howard1972risk}
Ronald~A Howard and James~E Matheson.
\newblock Risk-sensitive {Markov} decision processes.
\newblock {\em Management Science}, 18(7):356--369, 1972.

\bibitem{browne1995optimal}
Sid Browne.
\newblock Optimal investment policies for a firm with a random risk process:
  Exponential utility and minimizing the probability of ruin.
\newblock {\em Mathematics of operations research}, 20(4):937--958, 1995.

\bibitem{emmer2015best}
Susanne Emmer, Marie Kratz, and Dirk Tasche.
\newblock What is the best risk measure in practice? {A} comparison of standard
  measures.
\newblock {\em Journal of Risk}, 2015.

\bibitem{artzner1999coherent}
Philippe Artzner, Freddy Delbaen, Jean-Marc Eber, and David Heath.
\newblock Coherent measures of risk.
\newblock {\em Mathematical finance}, 9(3):203--228, 1999.

\bibitem{sobel1982variance}
Matthew~J Sobel.
\newblock The variance of discounted {Markov} decision processes.
\newblock {\em Journal of Applied Probability}, 19(4):794--802, 1982.

\bibitem{tamar2012policy}
Dotan Di~Castro, Aviv Tamar, and Shie Mannor.
\newblock Policy gradients with variance related risk criteria.
\newblock {\em International Conference on Machine Learning}, 2012.

\bibitem{sollis2009value}
Robert Sollis.
\newblock Value at risk: A critical overview.
\newblock {\em Journal of Financial Regulation and Compliance}, 2009.

\bibitem{danielsson2005subadditivity}
Jon Danielsson, Bj{\o}rn~N Jorgensen, Sarma Mandira, Gennady Samorodnitsky, and
  Casper~G De~Vries.
\newblock Subadditivity re-examined: The case for value-at-risk.
\newblock Technical report, Cornell University Operations Research and
  Industrial Engineering, 2005.

\bibitem{favre2002mean}
Laurent Favre and Jos{\'e}-Antonio Galeano.
\newblock Mean-modified value-at-risk optimization with hedge funds.
\newblock {\em The Journal of Alternative Investments}, 5(2):21--25, 2002.

\bibitem{rockafellar2000optimization}
R~Tyrrell Rockafellar, Stanislav Uryasev, et~al.
\newblock Optimization of conditional value-at-risk.
\newblock {\em Journal of risk}, 2:21--42, 2000.

\bibitem{ahmadi2012entropic}
Amir Ahmadi-Javid.
\newblock Entropic value-at-risk: A new coherent risk measure.
\newblock {\em Journal of Optimization Theory and Applications},
  155(3):1105--1123, 2012.

\bibitem{tamar2015policy}
Aviv Tamar, Yinlam Chow, Mohammad Ghavamzadeh, and Shie Mannor.
\newblock Policy gradient for coherent risk measures.
\newblock In {\em Advances in Neural Information Processing Systems}, pages
  1468--1476, 2015.

\bibitem{wang2000class}
Shaun~S Wang.
\newblock A class of distortion operators for pricing financial and insurance
  risks.
\newblock {\em Journal of Risk and Insurance}, pages 15--36, 2000.

\bibitem{chow2014algorithms}
Yinlam Chow and Mohammad Ghavamzadeh.
\newblock Algorithms for {CVaR} optimization in {MDPs}.
\newblock In {\em Advances in Neural Information Processing Systems}, pages
  3509--3517, 2014.

\bibitem{chow2015risk}
Yinlam Chow, Aviv Tamar, Shie Mannor, and Marco Pavone.
\newblock Risk-sensitive and robust decision-making: A {CVaR} optimization
  approach.
\newblock In {\em Advances in Neural Information Processing Systems}, pages
  1522--1530, 2015.

\bibitem{tamar2014optimizing}
Aviv Tamar, Yonatan Glassner, and Shie Mannor.
\newblock Optimizing the {CVaR} via sampling.
\newblock In {\em AAAI Conference on Artificial Intelligence}, page
  2993–2999, 2015.

\bibitem{bauerle2011markov}
Nicole B{\"a}uerle and Jonathan Ott.
\newblock Markov decision processes with average-value-at-risk criteria.
\newblock {\em Mathematical Methods of Operations Research}, 74(3):361--379,
  2011.

\bibitem{basel2013fundamental}
Basel Committee et~al.
\newblock Fundamental review of the trading book: A revised market risk
  framework.
\newblock {\em Consultative Document, October}, 2013.

\bibitem{boda2006time}
Kang Boda and Jerzy~A Filar.
\newblock Time consistent dynamic risk measures.
\newblock {\em Mathematical Methods of Operations Research}, 63(1):169--186,
  2006.

\bibitem{ruszczynski2010risk}
Andrzej Ruszczy{\'n}ski.
\newblock Risk-averse dynamic programming for {Markov} decision processes.
\newblock {\em Mathematical programming}, 125(2):235--261, 2010.

\bibitem{gagne2021two}
Christopher Gagne and Peter Dayan.
\newblock Two steps to risk sensitivity.
\newblock {\em Advances in Neural Information Processing Systems},
  34:22209--22220, 2021.

\bibitem{borkar2010risk}
Vivek Borkar and Rahul Jain.
\newblock Risk-constrained {Markov} decision processes.
\newblock In {\em 49th IEEE Conference on Decision and Control (CDC)}, pages
  2664--2669. IEEE, 2010.

\bibitem{prashanth2014policy}
LA~Prashanth.
\newblock Policy gradients for {CVaR}-constrained {MDPs}.
\newblock In {\em International Conference on Algorithmic Learning Theory},
  pages 155--169. Springer, 2014.

\bibitem{chow2017risk}
Yinlam Chow, Mohammad Ghavamzadeh, Lucas Janson, and Marco Pavone.
\newblock Risk-constrained reinforcement learning with percentile risk
  criteria.
\newblock {\em The Journal of Machine Learning Research}, 18(1):6070--6120,
  2017.

\bibitem{tang2019worst}
Yichuan~Charlie Tang, Jian Zhang, and Ruslan Salakhutdinov.
\newblock Worst cases policy gradients.
\newblock {\em Conference on Robot Learning}, 2020.

\bibitem{tamar2016sequential}
Aviv Tamar, Yinlam Chow, Mohammad Ghavamzadeh, and Shie Mannor.
\newblock Sequential decision making with coherent risk.
\newblock {\em IEEE Transactions on Automatic Control}, 62(7):3323--3338, 2016.

\bibitem{keramati2020being}
Ramtin Keramati, Christoph Dann, Alex Tamkin, and Emma Brunskill.
\newblock Being optimistic to be conservative: Quickly learning a {CVaR}
  policy.
\newblock In {\em AAAI Conference on Artificial Intelligence}, volume~34, pages
  4436--4443, 2020.

\bibitem{bellemare2017distributional}
Marc~G Bellemare, Will Dabney, and R{\'e}mi Munos.
\newblock A distributional perspective on reinforcement learning.
\newblock In {\em International Conference on Machine Learning}, pages
  449--458. PMLR, 2017.

\bibitem{morimura2010nonparametric}
Tetsuro Morimura, Masashi Sugiyama, Hisashi Kashima, Hirotaka Hachiya, and
  Toshiyuki Tanaka.
\newblock Nonparametric return distribution approximation for reinforcement
  learning.
\newblock In {\em International Conference on Machine Learning}, 2010.

\bibitem{dabney2018implicit}
Will Dabney, Georg Ostrovski, David Silver, and R{\'e}mi Munos.
\newblock Implicit quantile networks for distributional reinforcement learning.
\newblock In {\em International conference on machine learning}, pages
  1096--1105. PMLR, 2018.

\bibitem{ma2020dsac}
Xiaoteng Ma, Li~Xia, Zhengyuan Zhou, Jun Yang, and Qianchuan Zhao.
\newblock {DSAC}: Distributional soft actor critic for risk-sensitive
  reinforcement learning.
\newblock {\em arXiv preprint arXiv:2004.14547}, 2020.

\bibitem{urpi2021risk}
N{\'u}ria~Armengol Urp{\'\i}, Sebastian Curi, and Andreas Krause.
\newblock Risk-averse offline reinforcement learning.
\newblock {\em International Conference on Learning Representations}, 2021.

\bibitem{ben2000robust}
Aharon Ben-Tal and Arkadi Nemirovski.
\newblock Robust solutions of linear programming problems contaminated with
  uncertain data.
\newblock {\em Mathematical programming}, 88(3):411--424, 2000.

\bibitem{wolff2012robust}
Eric~M Wolff, Ufuk Topcu, and Richard~M Murray.
\newblock Robust control of uncertain {Markov} decision processes with temporal
  logic specifications.
\newblock In {\em IEEE Conference on Decision and Control}, pages 3372--3379.
  IEEE, 2012.

\bibitem{wiesemann2013robust}
Wolfram Wiesemann, Daniel Kuhn, and Ber{\c{c}} Rustem.
\newblock Robust {Markov} decision processes.
\newblock {\em Mathematics of Operations Research}, 38(1):153--183, 2013.

\bibitem{tamar2014scaling}
Aviv Tamar, Shie Mannor, and Huan Xu.
\newblock Scaling up robust {MDPs} using function approximation.
\newblock In {\em International Conference on Machine Learning}, pages
  181--189. PMLR, 2014.

\bibitem{mankowitz2020robust}
Daniel~J Mankowitz, Nir Levine, Rae Jeong, Yuanyuan Shi, Jackie Kay, Abbas
  Abdolmaleki, Jost~Tobias Springenberg, Timothy Mann, Todd Hester, and Martin
  Riedmiller.
\newblock Robust reinforcement learning for continuous control with model
  misspecification.
\newblock {\em International Conference on Learning Representations}, 2020.

\bibitem{delage2010percentile}
Erick Delage and Shie Mannor.
\newblock Percentile optimization for {Markov} decision processes with
  parameter uncertainty.
\newblock {\em Operations research}, 58(1), 2010.

\bibitem{mannor2015lightning}
Shie Mannor, Ofir Mebel, and Huan Xu.
\newblock Lightning does not strike twice: Robust mdps with coupled
  uncertainty.
\newblock In {\em International Conference on Machine Learning}, 2012.

\bibitem{adulyasak2015solving}
Yossiri Adulyasak, Pradeep Varakantham, Asrar Ahmed, and Patrick Jaillet.
\newblock Solving uncertain {MDPs} with objectives that are separable over
  instantiations of model uncertainty.
\newblock In {\em Twenty-Ninth AAAI Conference on Artificial Intelligence},
  2015.

\bibitem{ahmed2013regret}
Asrar Ahmed, Pradeep Varakantham, Yossiri Adulyasak, and Patrick Jaillet.
\newblock Regret based robust solutions for uncertain {Markov} decision
  processes.
\newblock In {\em Advances in Neural Information Processing Systems}, 2013.

\bibitem{ahmed2017sampling}
Asrar Ahmed, Pradeep Varakantham, Meghna Lowalekar, Yossiri Adulyasak, and
  Patrick Jaillet.
\newblock Sampling based approaches for minimizing regret in uncertain {Markov}
  decision processes.
\newblock {\em Journal of Artificial Intelligence Research}, 59, 2017.

\bibitem{chen2012tractable}
Katherine Chen and Michael Bowling.
\newblock Tractable objectives for robust policy optimization.
\newblock In {\em Advances in Neural Information Processing Systems}, pages
  2069--2077, 2012.

\bibitem{cubuktepe2020scenario}
Murat Cubuktepe, Nils Jansen, Sebastian Junges, Joost-Pieter Katoen, and Ufuk
  Topcu.
\newblock Scenario-based verification of uncertain {MDPs}.
\newblock In {\em Tools and Algorithms for the Construction and Analysis of
  Systems}, pages 287--305. Springer International Publishing, 2020.

\bibitem{steimle2018multi}
Lauren~N Steimle, David~L Kaufman, and Brian~T Denton.
\newblock Multi-model {Markov} decision processes.
\newblock {\em Optimization Online}, 2018.

\bibitem{xu2009parametric}
Huan Xu and Shie Mannor.
\newblock Parametric regret in uncertain {Markov} decision processes.
\newblock In {\em Proceedings of the 48h IEEE Conference on Decision and
  Control}. IEEE, 2009.

\bibitem{jaksch2010near}
Thomas Jaksch, Ronald Ortner, and Peter Auer.
\newblock Near-optimal regret bounds for reinforcement learning.
\newblock {\em Journal of Machine Learning Research}, 11(Apr):1563--1600, 2010.

\bibitem{cohen2020near}
Alon Cohen, Haim Kaplan, Yishay Mansour, and Aviv Rosenberg.
\newblock Near-optimal regret bounds for stochastic shortest path.
\newblock {\em International Conference on Machine Learning}, 2020.

\bibitem{tarbouriech2020no}
Jean Tarbouriech, Evrard Garcelon, Michal Valko, Matteo Pirotta, and Alessandro
  Lazaric.
\newblock No-regret exploration in goal-oriented reinforcement learning.
\newblock {\em International Conference on Machine Learning}, 2020.

\bibitem{regan2009regret}
Kevin Regan and Craig Boutilier.
\newblock Regret-based reward elicitation for {Markov} decision processes.
\newblock In {\em Proceedings of the Twenty-Fifth Conference on Uncertainty in
  Artificial Intelligence}, 2009.

\bibitem{regan2010robust}
Kevin Regan and Craig Boutilier.
\newblock Robust policy computation in reward-uncertain {MDPs} using
  nondominated policies.
\newblock In {\em Twenty-Fourth AAAI Conference on Artificial Intelligence},
  2010.

\bibitem{regan2011robust}
Kevin Regan and Craig Boutilier.
\newblock Robust online optimization of reward-uncertain {MDPs}.
\newblock In {\em International Joint Conference on Artificial Intelligence},
  2011.

\bibitem{roy2017reinforcement}
Aurko Roy, Huan Xu, and Sebastian Pokutta.
\newblock Reinforcement learning under model mismatch.
\newblock {\em Advances in Neural Information Processing Systems}, 30, 2017.

\bibitem{wang2021online}
Yue Wang and Shaofeng Zou.
\newblock Online robust reinforcement learning with model uncertainty.
\newblock {\em Advances in Neural Information Processing Systems}, 34, 2021.

\bibitem{kuang2022learning}
Yufei Kuang, Miao Lu, Jie Wang, Qi~Zhou, Bin Li, and Houqiang Li.
\newblock Learning robust policy against disturbance in transition dynamics via
  state-conservative policy optimization.
\newblock In {\em Proceedings of the AAAI Conference on Artificial
  Intelligence}, volume~36, pages 7247--7254, 2022.

\bibitem{pinto17robust}
Lerrel Pinto, James Davidson, Rahul Sukthankar, and Abhinav Gupta.
\newblock Robust adversarial reinforcement learning.
\newblock In {\em Proceedings of the International Conference on Machine
  Learning}, volume~70, pages 2817--2826, 2017.

\bibitem{pinto2017robust}
Lerrel Pinto, James Davidson, Rahul Sukthankar, and Abhinav Gupta.
\newblock Robust adversarial reinforcement learning.
\newblock In {\em International Conference on Machine Learning}, pages
  2817--2826. PMLR, 2017.

\bibitem{tessler2019action}
Chen Tessler, Yonathan Efroni, and Shie Mannor.
\newblock Action robust reinforcement learning and applications in continuous
  control.
\newblock In {\em International Conference on Machine Learning}, pages
  6215--6224. PMLR, 2019.

\bibitem{pan2019risk}
Xinlei Pan, Daniel Seita, Yang Gao, and John Canny.
\newblock Risk averse robust adversarial reinforcement learning.
\newblock In {\em 2019 International Conference on Robotics and Automation
  (ICRA)}, pages 8522--8528. IEEE, 2019.

\bibitem{ghavamzadeh2015bayesian}
Mohammad Ghavamzadeh, Shie Mannor, Joelle Pineau, and Aviv Tamar.
\newblock Bayesian reinforcement learning: A survey.
\newblock {\em Foundations and Trends in Machine Learning}, 8(5-6):359--483,
  2015.

\bibitem{duff2002optimal}
Michael~O'Gordon Duff.
\newblock {\em Optimal Learning: Computational procedures for Bayes-adaptive
  Markov decision processes}.
\newblock University of Massachusetts Amherst, 2002.

\bibitem{kaelbling1998planning}
Leslie~Pack Kaelbling, Michael~L Littman, and Anthony~R Cassandra.
\newblock Planning and acting in partially observable stochastic domains.
\newblock {\em Artificial intelligence}, 101(1-2):99--134, 1998.

\bibitem{poupart2006analytic}
Pascal Poupart, Nikos Vlassis, Jesse Hoey, and Kevin Regan.
\newblock An analytic solution to discrete bayesian reinforcement learning.
\newblock In {\em Proceedings of the 23rd international conference on Machine
  learning}, pages 697--704, 2006.

\bibitem{silver2010monte}
David Silver and Joel Veness.
\newblock {Monte-Carlo} planning in large pomdps.
\newblock {\em Advances in neural information processing systems}, 23, 2010.

\bibitem{guez2012efficient}
Arthur Guez, David Silver, and Peter Dayan.
\newblock Efficient {Bayes}-adaptive reinforcement learning using sample-based
  search.
\newblock {\em Advances in Neural Information Processing Systems},
  25:1025--1033, 2012.

\bibitem{guez2014bayes}
Arthur Guez, Nicolas Heess, David Silver, and Peter Dayan.
\newblock Bayes-adaptive simulation-based search with value function
  approximation.
\newblock {\em Advances in Neural Information Processing Systems}, 27, 2014.

\bibitem{zintgraf2020varibad}
Luisa Zintgraf, Kyriacos Shiarlis, Maximilian Igl, Sebastian Schulze, Yarin
  Gal, Katja Hofmann, and Shimon Whiteson.
\newblock {VariBAD}: A very good method for {B}ayes-adaptive deep {RL} via
  meta-learning.
\newblock {\em International Conference on Learning Representations}, 2020.

\bibitem{osband2013more}
Ian Osband, Daniel Russo, and Benjamin Van~Roy.
\newblock {(More)} efficient reinforcement learning via posterior sampling.
\newblock {\em Advances in Neural Information Processing Systems}, 26, 2013.

\bibitem{russo2018tutorial}
Daniel~J Russo, Benjamin Van~Roy, Abbas Kazerouni, Ian Osband, Zheng Wen,
  et~al.
\newblock A tutorial on {T}hompson sampling.
\newblock {\em Foundations and Trends in Machine Learning}, 11(1):1--96, 2018.

\bibitem{kolter2009near}
J~Zico Kolter and Andrew~Y Ng.
\newblock Near-{Bayesian} exploration in polynomial time.
\newblock In {\em International Conference on Machine Learning}, pages
  513--520, 2009.

\bibitem{sorg2010variance}
Jonathan Sorg, Satinder Singh, and Richard~L Lewis.
\newblock Variance-based rewards for approximate {Bayesian} reinforcement
  learning.
\newblock {\em Uncertainty in Artificial Intelligence}, 2010.

\bibitem{lange2012batch}
Sascha Lange, Thomas Gabel, and Martin Riedmiller.
\newblock Batch reinforcement learning.
\newblock In {\em Reinforcement learning}, pages 45--73. Springer, 2012.

\bibitem{levine2020offline}
Sergey Levine, Aviral Kumar, George Tucker, and Justin Fu.
\newblock Offline reinforcement learning: Tutorial, review, and perspectives on
  open problems.
\newblock {\em arXiv preprint arXiv:2005.01643}, 2020.

\bibitem{nie2021learning}
Xinkun Nie, Emma Brunskill, and Stefan Wager.
\newblock Learning when-to-treat policies.
\newblock {\em Journal of the American Statistical Association},
  116(533):392--409, 2021.

\bibitem{shortreed2011informing}
Susan~M Shortreed, Eric Laber, Daniel~J Lizotte, T~Scott Stroup, Joelle Pineau,
  and Susan~A Murphy.
\newblock Informing sequential clinical decision-making through reinforcement
  learning: An empirical study.
\newblock {\em Machine learning}, 84(1):109--136, 2011.

\bibitem{kandasamy2017batch}
Kirthevasan Kandasamy, Yoram Bachrach, Ryota Tomioka, Daniel Tarlow, and David
  Carter.
\newblock Batch policy gradient methods for improving neural conversation
  models.
\newblock {\em International Conference on Learning Representations}, 2017.

\bibitem{jaques2019way}
Natasha Jaques, Asma Ghandeharioun, Judy~Hanwen Shen, Craig Ferguson, Agata
  Lapedriza, Noah Jones, Shixiang Gu, and Rosalind Picard.
\newblock Way off-policy batch deep reinforcement learning of implicit human
  preferences in dialog.
\newblock {\em arXiv preprint arXiv:1907.00456}, 2019.

\bibitem{mandlekar2020iris}
Ajay Mandlekar, Fabio Ramos, Byron Boots, Silvio Savarese, Li~Fei-Fei, Animesh
  Garg, and Dieter Fox.
\newblock {IRIS}: Implicit reinforcement without interaction at scale for
  learning control from offline robot manipulation data.
\newblock In {\em 2020 IEEE International Conference on Robotics and Automation
  (ICRA)}, pages 4414--4420. IEEE, 2020.

\bibitem{rafailov2021offline}
Rafael Rafailov, Tianhe Yu, Aravind Rajeswaran, and Chelsea Finn.
\newblock Offline reinforcement learning from images with latent space models.
\newblock In {\em Learning for Dynamics and Control}, pages 1154--1168. PMLR,
  2021.

\bibitem{xiao2022curse}
Chenjun Xiao, Ilbin Lee, Bo~Dai, Dale Schuurmans, and Csaba Szepesvari.
\newblock The curse of passive data collection in batch reinforcement learning.
\newblock In {\em International Conference on Artificial Intelligence and
  Statistics}, pages 8413--8438. PMLR, 2022.

\bibitem{matsushima2020deployment}
Tatsuya Matsushima, Hiroki Furuta, Yutaka Matsuo, Ofir Nachum, and Shixiang Gu.
\newblock Deployment-efficient reinforcement learning via model-based offline
  optimization.
\newblock {\em International Conference on Learning Representations}, 2021.

\bibitem{bai2019provably}
Yu~Bai, Tengyang Xie, Nan Jiang, and Yu-Xiang Wang.
\newblock Provably efficient {Q}-learning with low switching cost.
\newblock {\em Advances in Neural Information Processing Systems}, 32, 2019.

\bibitem{kalashnikov2018scalable}
Dmitry Kalashnikov, Alex Irpan, Peter Pastor, Julian Ibarz, Alexander Herzog,
  Eric Jang, Deirdre Quillen, Ethan Holly, Mrinal Kalakrishnan, Vincent
  Vanhoucke, et~al.
\newblock Scalable deep reinforcement learning for vision-based robotic
  manipulation.
\newblock In {\em Conference on Robot Learning}, pages 651--673. PMLR, 2018.

\bibitem{kostrikov2021offline}
Ilya Kostrikov, Rob Fergus, Jonathan Tompson, and Ofir Nachum.
\newblock Offline reinforcement learning with {Fisher} divergence critic
  regularization.
\newblock In {\em International Conference on Machine Learning}, pages
  5774--5783. PMLR, 2021.

\bibitem{xie2021policy}
Tengyang Xie, Nan Jiang, Huan Wang, Caiming Xiong, and Yu~Bai.
\newblock Policy finetuning: Bridging sample-efficient offline and online
  reinforcement learning.
\newblock {\em Advances in Neural Information Processing Systems},
  34:27395--27407, 2021.

\bibitem{kumar2019stabilizing}
Aviral Kumar, Justin Fu, Matthew Soh, George Tucker, and Sergey Levine.
\newblock Stabilizing off-policy {Q}-learning via bootstrapping error
  reduction.
\newblock {\em Advances in Neural Information Processing Systems}, 32, 2019.

\bibitem{clavera2018model}
Ignasi Clavera, Jonas Rothfuss, John Schulman, Yasuhiro Fujita, Tamim Asfour,
  and Pieter Abbeel.
\newblock Model-based reinforcement learning via meta-policy optimization.
\newblock In {\em Conference on Robot Learning}, pages 617--629. PMLR, 2018.

\bibitem{kidambi2020morel}
Rahul Kidambi, Aravind Rajeswaran, Praneeth Netrapalli, and Thorsten Joachims.
\newblock {MOReL}: Model-based offline reinforcement learning.
\newblock {\em Advances in Neural Information Processing Systems},
  33:21810--21823, 2020.

\bibitem{kurutach2018model}
Thanard Kurutach, Ignasi Clavera, Yan Duan, Aviv Tamar, and Pieter Abbeel.
\newblock Model-ensemble trust-region policy optimization.
\newblock {\em International Conference on Learning Representations}, 2018.

\bibitem{ernst2005tree}
Damien Ernst, Pierre Geurts, and Louis Wehenkel.
\newblock Tree-based batch mode reinforcement learning.
\newblock {\em Journal of Machine Learning Research}, 6, 2005.

\bibitem{lagoudakis2003least}
Michail~G Lagoudakis and Ronald Parr.
\newblock Least-squares policy iteration.
\newblock {\em The Journal of Machine Learning Research}, 4:1107--1149, 2003.

\bibitem{riedmiller2005neural}
Martin Riedmiller.
\newblock Neural fitted {Q} iteration--first experiences with a data efficient
  neural reinforcement learning method.
\newblock In {\em European Conference on Machine Learning}, pages 317--328.
  Springer, 2005.

\bibitem{fujimoto2019off}
Scott Fujimoto, David Meger, and Doina Precup.
\newblock Off-policy deep reinforcement learning without exploration.
\newblock In {\em International Conference on Machine Learning}, pages
  2052--2062. PMLR, 2019.

\bibitem{kostrikov2022offline}
Ilya Kostrikov, Ashvin Nair, and Sergey Levine.
\newblock Offline reinforcement learning with implicit {Q}-learning.
\newblock In {\em International Conference on Learning Representations}, 2022.

\bibitem{wu2019behavior}
Yifan Wu, George Tucker, and Ofir Nachum.
\newblock Behavior regularized offline reinforcement learning.
\newblock {\em arXiv preprint arXiv:1911.11361}, 2019.

\bibitem{siegel2020keep}
Noah Siegel, Jost~Tobias Springenberg, Felix Berkenkamp, Abbas Abdolmaleki,
  Michael Neunert, Thomas Lampe, Roland Hafner, Nicolas Heess, and Martin
  Riedmiller.
\newblock Keep doing what worked: Behavior modelling priors for offline
  reinforcement learning.
\newblock In {\em International Conference on Learning Representations}, 2020.

\bibitem{fujimoto2021minimalist}
Scott Fujimoto and Shixiang~Shane Gu.
\newblock A minimalist approach to offline reinforcement learning.
\newblock {\em Advances in Neural Information Processing Systems}, 34, 2021.

\bibitem{wu2022supported}
Jialong Wu, Haixu Wu, Zihan Qiu, Jianmin Wang, and Mingsheng Long.
\newblock Supported policy optimization for offline reinforcement learning.
\newblock {\em Advances in Neural Information Processing Systems}, 2022.

\bibitem{kumar2020conservative}
Aviral Kumar, Aurick Zhou, George Tucker, and Sergey Levine.
\newblock Conservative {Q}-learning for offline reinforcement learning.
\newblock {\em Advances in Neural Information Processing Systems},
  33:1179--1191, 2020.

\bibitem{cheng2022adversarially}
Ching-An Cheng, Tengyang Xie, Nan Jiang, and Alekh Agarwal.
\newblock Adversarially trained actor critic for offline reinforcement
  learning.
\newblock {\em International Conference on Machine Learning}, 2022.

\bibitem{xie2021bellman}
Tengyang Xie, Ching-An Cheng, Nan Jiang, Paul Mineiro, and Alekh Agarwal.
\newblock Bellman-consistent pessimism for offline reinforcement learning.
\newblock {\em Advances in Neural Information Processing Systems}, 34, 2021.

\bibitem{agarwal2020optimistic}
Rishabh Agarwal, Dale Schuurmans, and Mohammad Norouzi.
\newblock An optimistic perspective on offline reinforcement learning.
\newblock In {\em International Conference on Machine Learning}, pages
  104--114. PMLR, 2020.

\bibitem{an2021uncertainty}
Gaon An, Seungyong Moon, Jang-Hyun Kim, and Hyun~Oh Song.
\newblock Uncertainty-based offline reinforcement learning with diversified
  {Q}-ensemble.
\newblock {\em Advances in Neural Information Processing Systems},
  34:7436--7447, 2021.

\bibitem{fujimoto2018addressing}
Scott Fujimoto, Herke Hoof, and David Meger.
\newblock Addressing function approximation error in actor-critic methods.
\newblock In {\em International conference on Machine Learning}, pages
  1587--1596. PMLR, 2018.

\bibitem{brandfonbrener2021offline}
David Brandfonbrener, Will Whitney, Rajesh Ranganath, and Joan Bruna.
\newblock Offline {RL} without off-policy evaluation.
\newblock {\em Advances in Neural Information Processing Systems},
  34:4933--4946, 2021.

\bibitem{peng2019advantage}
Xue~Bin Peng, Aviral Kumar, Grace Zhang, and Sergey Levine.
\newblock Advantage-weighted regression: Simple and scalable off-policy
  reinforcement learning.
\newblock {\em arXiv preprint arXiv:1910.00177}, 2019.

\bibitem{liu2019off}
Yao Liu, Adith Swaminathan, Alekh Agarwal, and Emma Brunskill.
\newblock Off-policy policy gradient with state distribution correction.
\newblock {\em International Conference on Machine Learning RL4RealLife
  Workshop}, 2019.

\bibitem{nachum2019algaedice}
Ofir Nachum, Bo~Dai, Ilya Kostrikov, Yinlam Chow, Lihong Li, and Dale
  Schuurmans.
\newblock {AlgaeDICE}: Policy gradient from arbitrary experience.
\newblock {\em arXiv preprint arXiv:1912.02074}, 2019.

\bibitem{sutton1991dyna}
Richard~S Sutton.
\newblock Dyna, an integrated architecture for learning, planning, and
  reacting.
\newblock {\em ACM Sigart Bulletin}, 2(4):160--163, 1991.

\bibitem{argenson2021modelbased}
Arthur Argenson and Gabriel Dulac-Arnold.
\newblock Model-based offline planning.
\newblock In {\em International Conference on Learning Representations}, 2021.

\bibitem{yu2020mopo}
Tianhe Yu, Garrett Thomas, Lantao Yu, Stefano Ermon, James~Y Zou, Sergey
  Levine, Chelsea Finn, and Tengyu Ma.
\newblock {MOPO}: Model-based offline policy optimization.
\newblock {\em Advances in Neural Information Processing Systems},
  33:14129--14142, 2020.

\bibitem{yu2021combo}
Tianhe Yu, Aviral Kumar, Rafael Rafailov, Aravind Rajeswaran, Sergey Levine,
  and Chelsea Finn.
\newblock {COMBO}: Conservative offline model-based policy optimization.
\newblock {\em Advances in Neural Information Processing Systems}, 34, 2021.

\bibitem{ball2021augmented}
Philip~J Ball, Cong Lu, Jack Parker-Holder, and Stephen Roberts.
\newblock Augmented world models facilitate zero-shot dynamics generalization
  from a single offline environment.
\newblock In {\em International Conference on Machine Learning}, pages
  619--629. PMLR, 2021.

\bibitem{cang2021behavioral}
Catherine Cang, Aravind Rajeswaran, Pieter Abbeel, and Michael Laskin.
\newblock Behavioral priors and dynamics models: Improving performance and
  domain transfer in offline {RL}.
\newblock In {\em Deep RL Workshop NeurIPS 2021}, 2021.

\bibitem{swazinna2021overcoming}
Phillip Swazinna, Steffen Udluft, and Thomas Runkler.
\newblock Overcoming model bias for robust offline deep reinforcement learning.
\newblock {\em Engineering Applications of Artificial Intelligence},
  104:104366, 2021.

\bibitem{yang2021pareto}
Yijun Yang, Jing Jiang, Tianyi Zhou, Jie Ma, and Yuhui Shi.
\newblock Pareto policy pool for model-based offline reinforcement learning.
\newblock In {\em International Conference on Learning Representations}, 2022.

\bibitem{gawlikowski2021survey}
Jakob Gawlikowski, Cedrique Rovile~Njieutcheu Tassi, Mohsin Ali, Jongseok Lee,
  Matthias Humt, Jianxiang Feng, Anna Kruspe, Rudolph Triebel, Peter Jung,
  Ribana Roscher, et~al.
\newblock A survey of uncertainty in deep neural networks.
\newblock {\em arXiv preprint arXiv:2107.03342}, 2021.

\bibitem{lu2022revisiting}
Cong Lu, Philip Ball, Jack Parker-Holder, Michael Osborne, and Stephen~J
  Roberts.
\newblock Revisiting design choices in offline model based reinforcement
  learning.
\newblock In {\em International Conference on Learning Representations}, 2022.

\bibitem{ovadia2019can}
Yaniv Ovadia, Emily Fertig, Jie Ren, Zachary Nado, David Sculley, Sebastian
  Nowozin, Joshua Dillon, Balaji Lakshminarayanan, and Jasper Snoek.
\newblock Can you trust your model's uncertainty? {Evaluating} predictive
  uncertainty under dataset shift.
\newblock {\em Advances in Neural Information Processing Systems}, 32, 2019.

\bibitem{lee2020representation}
Byung-Jun Lee, Jongmin Lee, and Kee-Eung Kim.
\newblock Representation balancing offline model-based reinforcement learning.
\newblock In {\em International Conference on Learning Representations}, 2020.

\bibitem{rajeswaran2020game}
Aravind Rajeswaran, Igor Mordatch, and Vikash Kumar.
\newblock A game theoretic framework for model based reinforcement learning.
\newblock In {\em International Conference on Machine Learning}, pages
  7953--7963. PMLR, 2020.

\bibitem{hishinuma2021weighted}
Toru Hishinuma and Kei Senda.
\newblock Weighted model estimation for offline model-based reinforcement
  learning.
\newblock {\em Advances in Neural Information Processing Systems}, 34, 2021.

\bibitem{wang2021offline}
Jianhao Wang, Wenzhe Li, Haozhe Jiang, Guangxiang Zhu, Siyuan Li, and Chongjie
  Zhang.
\newblock Offline reinforcement learning with reverse model-based imagination.
\newblock {\em Advances in Neural Information Processing Systems}, 34, 2021.

\bibitem{schaal1996learning}
Stefan Schaal.
\newblock Learning from demonstration.
\newblock {\em Advances in Neural Information Processing Systems}, 9, 1996.

\bibitem{pomerleau1991efficient}
Dean~A Pomerleau.
\newblock Efficient training of artificial neural networks for autonomous
  navigation.
\newblock {\em Neural computation}, 3(1):88--97, 1991.

\bibitem{ziebart2008maximum}
Brian~D Ziebart, Andrew~L Maas, J~Andrew Bagnell, Anind~K Dey, et~al.
\newblock Maximum entropy inverse reinforcement learning.
\newblock In {\em AAAI Conference on Artificial Intelligence}, pages
  1433--1438, 2008.

\bibitem{ross2010efficient}
St{\'e}phane Ross and Drew Bagnell.
\newblock Efficient reductions for imitation learning.
\newblock In {\em International Conference on Artificial Intelligence and
  Statistics}, pages 661--668, 2010.

\bibitem{ho2016generative}
Jonathan Ho and Stefano Ermon.
\newblock Generative adversarial imitation learning.
\newblock {\em Advances in Neural Information Processing Systems}, 29, 2016.

\bibitem{ross2011reduction}
St{\'e}phane Ross, Geoffrey Gordon, and Drew Bagnell.
\newblock A reduction of imitation learning and structured prediction to
  no-regret online learning.
\newblock In {\em International Conference on Artificial Intelligence and
  Statistics}, pages 627--635, 2011.

\bibitem{kober2009learning}
Jens Kober and Jan Peters.
\newblock Learning motor primitives for robotics.
\newblock In {\em 2009 IEEE International Conference on Robotics and
  Automation}, pages 2112--2118. IEEE, 2009.

\bibitem{sun2017deeply}
Wen Sun, Arun Venkatraman, Geoffrey~J Gordon, Byron Boots, and J~Andrew
  Bagnell.
\newblock Deeply aggrevated: Differentiable imitation learning for sequential
  prediction.
\newblock In {\em International Conference on Machine Learning}, pages
  3309--3318. PMLR, 2017.

\bibitem{clouse1996integrating}
Jeffery~Allen Clouse.
\newblock {\em On integrating apprentice learning and reinforcement learning}.
\newblock University of Massachusetts Amherst, 1996.

\bibitem{chernova2009interactive}
Sonia Chernova and Manuela Veloso.
\newblock Interactive policy learning through confidence-based autonomy.
\newblock {\em Journal of Artificial Intelligence Research}, 34, 2009.

\bibitem{del2018not}
Francesco Del~Duchetto, Ayse Kucukyilmaz, Luca Iocchi, and Marc Hanheide.
\newblock Do not make the same mistakes again and again: Learning local
  recovery policies for navigation from human demonstrations.
\newblock {\em IEEE Robotics and Automation Letters}, 3(4), 2018.

\bibitem{petrik2012an}
Marek Petrik and Dharmashankar Subramanian.
\newblock An approximate solution method for large risk-averse {Markov}
  decision processes.
\newblock In {\em Proceedings of the Twenty-Eighth Conference on Uncertainty in
  Artificial Intelligence}, page 805–814, 2012.

\bibitem{mern2020bayesian}
John Mern, Anil Yildiz, Zachary Sunberg, Tapan Mukerji, and Mykel~J
  Kochenderfer.
\newblock Bayesian optimized {Monte Carlo} planning.
\newblock {\em AAAI Conference on Artificial Intelligence}, 2021.

\bibitem{doshi2016hidden}
Finale Doshi-Velez and George Konidaris.
\newblock Hidden parameter {Markov} decision processes: A semiparametric
  regression approach for discovering latent task parametrizations.
\newblock In {\em International Joint Conference on Artificial Intelligence},
  2016.

\bibitem{uehara2022pessimistic}
Masatoshi Uehara and Wen Sun.
\newblock Pessimistic model-based offline reinforcement learning under partial
  coverage.
\newblock {\em International Conference on Learning Representations}, 2022.

\bibitem{eriksson2022sentinel}
Hannes Eriksson, Debabrota Basu, Mina Alibeigi, and Christos Dimitrakakis.
\newblock Sentinel: taming uncertainty with ensemble based distributional
  reinforcement learning.
\newblock In {\em Uncertainty in Artificial Intelligence}, pages 631--640.
  PMLR, 2022.

\bibitem{ray2019benchmarking}
Alex Ray, Joshua Achiam, and Dario Amodei.
\newblock {Benchmarking Safe Exploration in Deep Reinforcement Learning}.
\newblock 2019.

\bibitem{vats2022synergistic}
Shivam Vats, Oliver Kroemer, and Maxim Likhachev.
\newblock Synergistic scheduling of learning and allocation of tasks in
  human-robot teams.
\newblock {\em International Conference on Robotics and Automation}, 2022.

\bibitem{dorfman2021offline}
Ron Dorfman, Idan Shenfeld, and Aviv Tamar.
\newblock Offline meta reinforcement learning--identifiability challenges and
  effective data collection strategies.
\newblock {\em Advances in Neural Information Processing Systems},
  34:4607--4618, 2021.

\bibitem{ghosh2022offline}
Dibya Ghosh, Anurag Ajay, Pulkit Agrawal, and Sergey Levine.
\newblock Offline {RL} policies should be trained to be adaptive.
\newblock In {\em International Conference on Machine Learning}, pages
  7513--7530. PMLR, 2022.

\bibitem{gurtler2023benchmarking}
Nico Gürtler, Sebastian Blaes, Pavel Kolev, Felix Widmaier, Manuel Wuthrich,
  Stefan Bauer, Bernard Schölkopf, and Georg Martius.
\newblock Benchmarking offline reinforcement learning on real-robot hardware.
\newblock In {\em International Conference on Learning Representations}, 2023.

\bibitem{zhao2020sim}
Wenshuai Zhao, Jorge~Pe{\~n}a Queralta, and Tomi Westerlund.
\newblock Sim-to-real transfer in deep reinforcement learning for robotics: {A}
  survey.
\newblock In {\em 2020 IEEE symposium series on computational intelligence
  (SSCI)}, pages 737--744. IEEE, 2020.

\end{thebibliography}
